\documentclass[11pt]{article}
\usepackage{iclr2026_conference,times}
\usepackage[margin=1in]{geometry}
\usepackage{amsmath,amssymb}
\usepackage{booktabs}
\usepackage{multirow}
\usepackage{makecell}
\usepackage{tabularx}
\usepackage{graphicx}
\usepackage{subcaption}
\usepackage{xcolor}
\usepackage{hyperref}
\usepackage[nameinlink,capitalise]{cleveref}
\usepackage{natbib}
\usepackage{url}
\usepackage{fancyvrb}
\usepackage{fvextra}
\usepackage{multicol}
\usepackage{enumitem}
\usepackage{xurl}
\usepackage[normalem]{ulem}

\usepackage[most]{tcolorbox}
\usepackage{listings}

\newcommand{\op}[1]{\item \path{#1}}
\definecolor{crimson}{HTML}{DC143C}

\newenvironment{oplist}[1][3]
{
\begin{tcolorbox}[
  colback=gray!3,
  colframe=gray!35,
  boxrule=0.4pt,
  arc=1mm,
  left=3pt,
  right=3pt,
  top=3pt,
  bottom=3pt,
  title=Allowed ATen Operators,
  fonttitle=\bfseries\small
]
\scriptsize
\setlength{\columnsep}{10pt}
\begin{multicols}{#1}
\begin{itemize}[
  leftmargin=*,
  itemsep=0pt,
  parsep=0pt,
  topsep=0pt,
  partopsep=0pt,
  label={}
]
}
{
\end{itemize}
\end{multicols}
\end{tcolorbox}
}
\iclrfinalcopy

\definecolor{mtblue}{HTML}{C2410C}

\newcommand{\makeMusaTitle}{
\begin{center}

\vspace*{-6.3em}

\begin{minipage}{\textwidth}
    \raggedright
    \includegraphics[height=0.65cm]{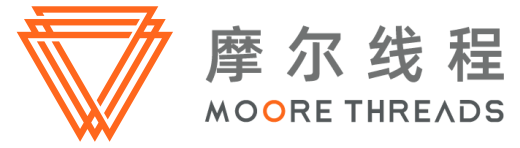}
\end{minipage}

\rule{\textwidth}{0.8pt}

{\Large\bfseries MusaCoder: Native GPU Kernel Generation\\[0.4em]
with Full-Stack Training on Moore Threads GPUs}

\vspace{1.0em}

{\large\bfseries
Kun Cheng$^{1}$,\quad
Songshuo Lu$^{1}$,\quad
Sicong Liao$^{1}$,\quad
Tankun Li$^{1}$,\quad
Yafei Zhang$^{1}$\\[0.35em]
Dong Yang,\quad
Qiheng Lv,\quad
Hua Wang,\quad
Zhi Chen$^{2}$,\quad
Yaohua Tang
}

\vspace{1.0em}

{\normalsize
Moore Threads AI\\[0.25em]
\texttt{tangyaohua28@gmail.com}
}

\vspace{0.6em}

\footnotetext[1]{Equal contribution.}
\footnotetext[2]{Corresponding author.}

\vspace{1.5em}

\end{center}
}

\title{
\includegraphics[height=1.2cm]{figures/logo.png}\\[1.0em]
\textbf{MusaCoder}: Native GPU Kernel Generation with Full-Stack Training on Moore Threads GPU}

\author{%
    Kun Chen\thanks{Equal contribution.} \quad
    Songshuo Lu\footnotemark[1] \quad
    Sicong Liao\footnotemark[1] \quad
    Tankun Li\footnotemark[1] \quad
    Yafei Zhang\footnotemark[1] \\
    \textbf{Dong Yang} \quad 
    \textbf{Qiheng Lv} \quad
    \textbf{Hua Wang} \quad
    \textbf{Zhi Chen}\thanks{Corresponding author.} \quad
    \textbf{Yaohua Tang} \\
    \\[0.2em]
    Moore Threads AI \\
    \texttt{tangyaohua28@gmail.com}
}
\date{}

\begin{document}
\makeMusaTitle

\begin{figure}[h]
    \centering
    \includegraphics[width=0.8\linewidth]{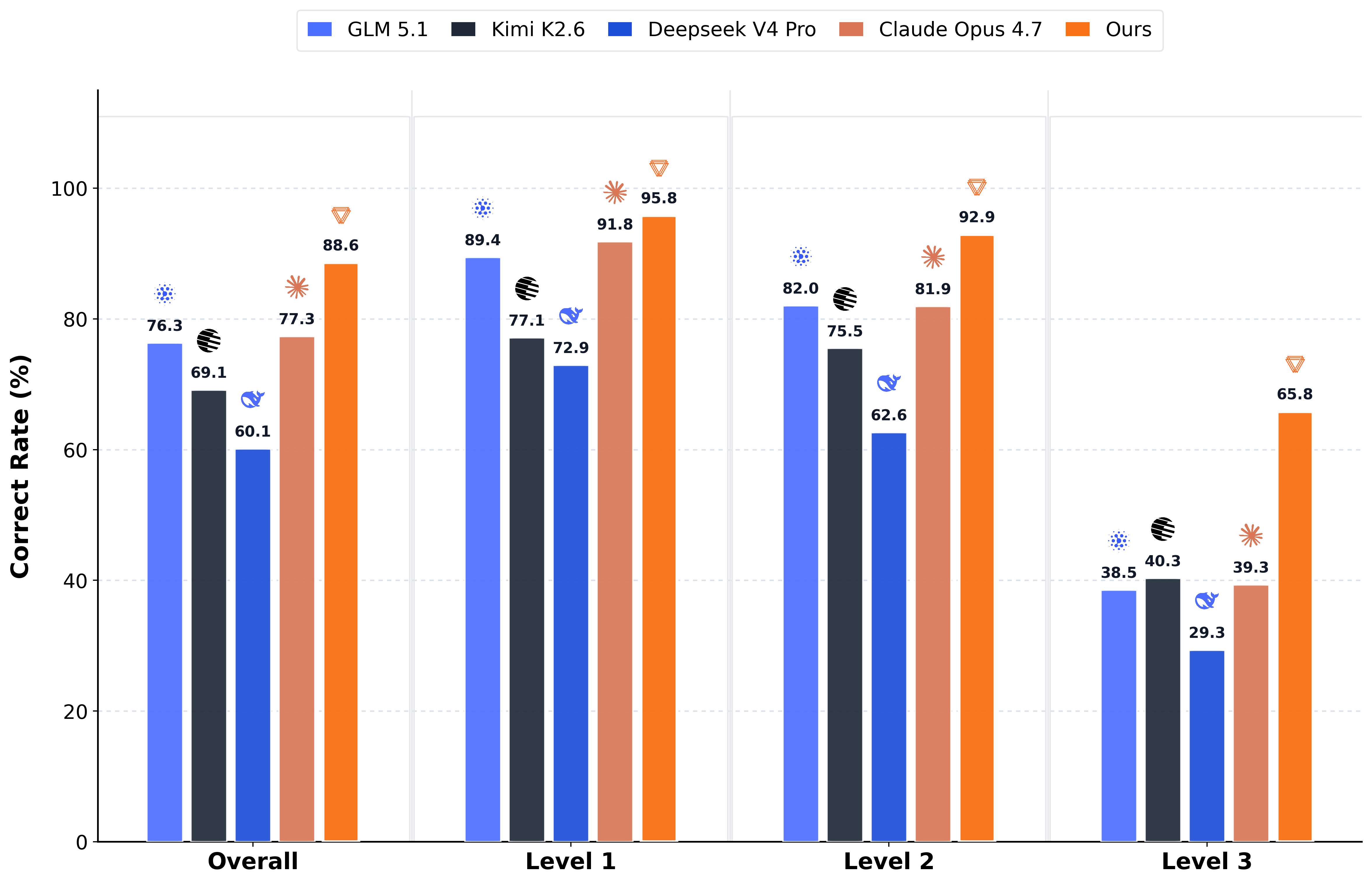}
    \caption{KernelBench performance comparison.}
    \label{fig:kernelbench_level_comparison}
\end{figure}

\begin{abstract}
Native GPU kernel generation turns high-level tensor programs into executable, efficient low-level code. Existing Large Language Models (LLMs) struggle with this task, while execution-based reinforcement learning suffers from sparse rewards, reward hacking, and training instability. We present \textbf{MusaCoder}, a full-stack training framework for native GPU kernel generation on CUDA and MUSA backends. MusaCoder combines progressive kernel-oriented data synthesis, diversity-preserving rejection fine-tuning, and execution-feedback Reinforcement Learning (RL) through \textbf{MooreEval}, a distributed verifier and reward environment. To stabilize RL, MusaCoder introduces \textbf{PrimeEcho} for first-turn-anchored multi-turn rewards, \textbf{Buffered Dynamic Retry} for recovering signals from all-failed hard samples, and \textbf{MirrorPop} for off-policy sequence filtering. Experiments on KernelBench and a MUSA-ported variant show that MusaCoder outperforms strong open-source and proprietary baselines in both correctness and empirical speedup, with the 9B model matching or exceeding frontier closed-source models and the 27B model establishing a new state of the art. These results demonstrate not only the effectiveness of full-stack execution-feedback training for native kernel generation, but also the capability of Moore Threads GPUs to support the complete LLM post-training stack, providing a practical foundation for large-model training and optimization on emerging accelerators.  
\end{abstract}

\section{Introduction}

Modern neural networks increasingly rely on highly optimized GPU kernels to fully utilize accelerator hardware. While vendor-specific libraries (e.g., NVIDIA's cuBLAS and cuDNN) and static templates (e.g., CUTLASS) deliver near-optimal performance for canonical operators, they struggle to keep pace with the rapid proliferation of novel and long-tail operator fusions in frontier AI models. Consequently, native GPU kernel generation—translating high-level tensor computations directly into optimized low-level device code—has become increasingly important. Recently, Large Language Models (LLMs) have shown promising potential for automating this process, offering a scalable alternative to labor-intensive manual kernel engineering.

However, translating high-level PyTorch semantics into executable GPU kernels with LLMs remains fundamentally challenging. Unlike general-purpose code generation, synthesizing native kernels \textit{from scratch} suffers from extremely low initial success rates. 
Without explicit domain grounding, off-the-shelf LLMs frequently struggle with GPU execution semantics, mathematical derivations, and multi-dimensional index mapping.
Moreover, GPU kernels must satisfy strict correctness constraints, requiring the generated code to be compilable, numerically stable across edge cases, and compliant with hardware execution semantics. As a result, even minor logical or mathematical errors can lead to widespread compilation failures, illegal memory accesses, or incorrect numerical outputs.



Recent years have witnessed the widespread adoption of reinforcement learning (RL) with execution-based feedback in code generation. 
However, extending these paradigms to GPU kernel generation 
introduces several major challenges. 
First, \textit{reward sparsity}: the high failure rate of generated kernels often produces ``all-failed'' rollout groups, yielding limited useful learning signal. Second, \textit{reward hacking}: models tend to exploit high-level API fallbacks (e.g., invoking \texttt{torch.matmul} or \texttt{aten::*} routines) instead of generating fully native kernels. Third, \textit{off-policy drift}: the asynchronous and multi-turn nature of execution-based rollouts amplifies gradient instability during training. These challenges become even more pronounced in emerging non-CUDA accelerator ecosystems, where large-scale training corpora and reliable verification infrastructures remain limited.

\begin{figure}[t]
    \centering
    \includegraphics[width=\linewidth]{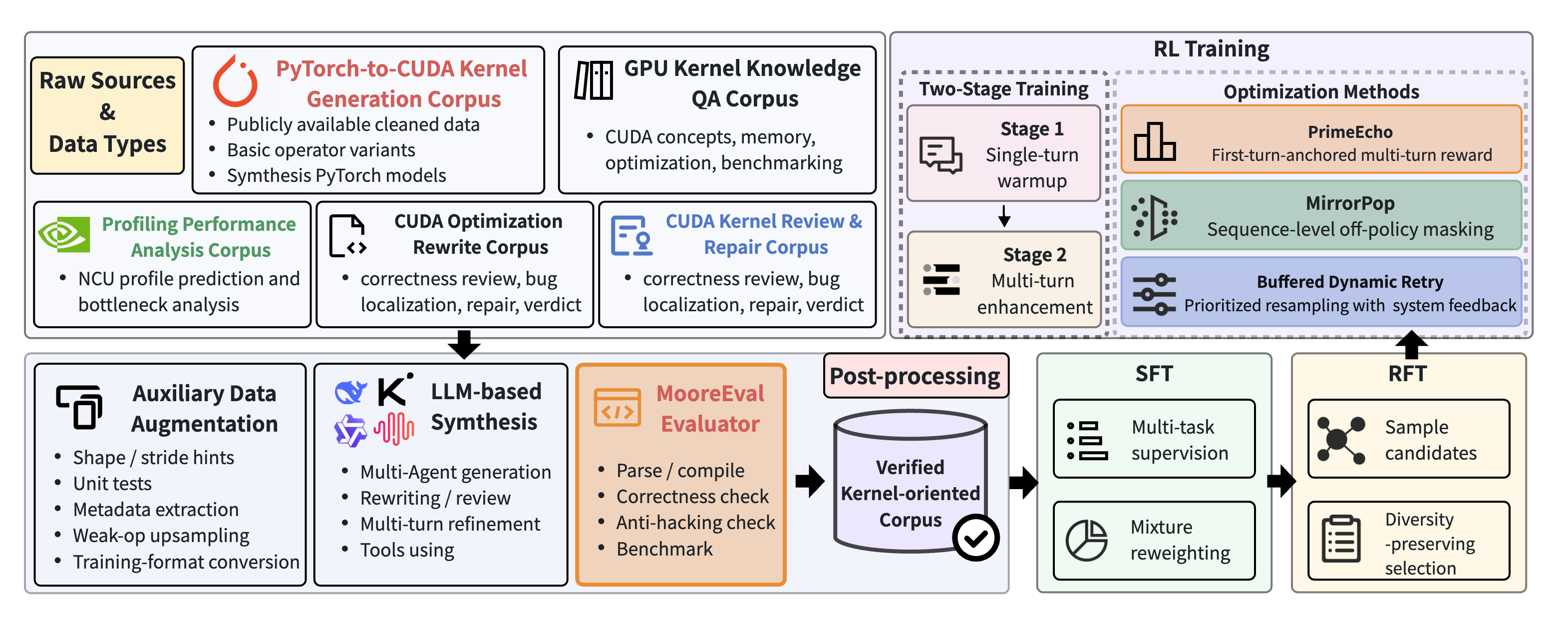}
    \caption{\textbf{Overview of the MusaCoder training pipeline.} The pipeline first constructs a kernel-oriented corpus from diverse raw sources, including PyTorch-to-CUDA/MUSA generation data, GPU kernel knowledge QA, profiling analysis, optimization rewrite, and kernel review/repair data. The data are further augmented through shape/stride hints, unit tests, metadata extraction, weak-operator upsampling, LLM-based synthesis, and MooreEval-based verification. After post-processing, the verified corpus is used for multi-task SFT and diversity-preserving RFT. Finally, MusaCoder is optimized with a two-stage RL procedure, consisting of single-turn warmup and multi-turn enhancement, together with three stabilization techniques: PrimeEcho, MirrorPop, and Buffered Dynamic Retry.
} 
    \label{fig:main_pipeline}
\end{figure}

To address these challenges, we present \textbf{MusaCoder}, an end-to-end LLM-driven framework for native GPU kernel synthesis developed on Moore Threads GPU clusters (Fig.~\ref{fig:main_pipeline}).
To deal with the extremely low initial success rate of kernel generation and establish a strong policy initialization, we design a comprehensive three-stage data synthesis pipeline. Rather than relying on naive code translation, the pipeline first expands task diversity and injects foundational CUDA/MUSA knowledge to improve the model's understanding of GPU programming semantics. It then introduces explicit tensor shape annotations and a structured six-step reasoning format to guide the mapping from high-level tensor operations to thread- and memory-level execution semantics. Finally, the pipeline incorporates profiling analysis and multi-turn interactive trajectories to reduce syntactic, logical, and execution-level errors that would otherwise
undermine the stability and effectiveness of the overall synthesis pipeline.


Leveraging this progressive data pipeline, we first perform \textbf{Multi-task Supervised Fine-Tuning (SFT)} to teach the model canonical kernel generation patterns and error diagnosis behaviors. To further align the model with execution correctness prior to RL, we subsequently apply \textbf{Diversity-Preserving Rejection Sampling Fine-Tuning (RFT)}. Conventional RFT typically retains only the single fastest implementation, which can lead to rapid entropy collapse and reduced exploration diversity. In contrast, our approach employs an execution sandbox to filter deterministic failures while deliberately preserving a heterogeneous set of verified correct implementations. This stage substantially improves absolute correctness while maintaining the behavioral diversity necessary for effective downstream reinforcement learning.


Building upon this diverse initialization, we formulate kernel synthesis as a verifiable reinforcement learning problem. At the core of our RL stage is \textbf{MooreEval}, a scalable distributed execution sandbox for automated kernel evaluation. MooreEval enforces a strict \textit{correctness-first} verification hierarchy and incorporates anti-hacking mechanisms to detect and penalize forbidden PyTorch fallbacks. Guided by execution feedback from MooreEval, we employ a two-stage RL training strategy consisting of a single-turn execution warmup followed by iterative multi-turn feedback rollouts.

To improve the robustness and stability of execution-based RL,
we introduce three complementary stabilization mechanisms:

\begin{itemize}
\item \textbf{PrimeEcho}: A first-turn-anchored trajectory reward formulation that incorporates multi-turn corrective feedback to improve exploration while mitigating multi-turn reward hacking, without degrading zero-shot generation quality.

\item \textbf{Buffered Dynamic Retry (BDR)}: A dynamic resampling strategy that converts terminal all-failed groups into feedback-conditioned repair tasks, recovering useful learning signals from difficult long-tail samples and alleviating reward sparsity.

\item \textbf{MirrorPop}: A sequence-level off-policy masking strategy that more accurately estimates distributional drift magnitude, improving the detection of severe off-policy samples and enabling more stable RL training.

\end{itemize}


Empirical evaluations on the KernelBench suite and its MUSA-ported variant show that MusaCoder significantly outperforms leading proprietary and open-source models, including Claude Opus 4.7 and DeepSeek-V4-Pro (Fig.~\ref{fig:kernelbench_level_comparison}). In particular, the 27B variant of MusaCoder achieves state-of-the-art kernel correctness and runtime speedups. 
These results demonstrate the effectiveness of our end-to-end alignment pipeline in enabling hardware-aware GPU kernel generation, paving the way toward fully automated kernel synthesis across platforms.

\section{Related Work}

LLM-driven GPU kernel generation has become an important research direction in code synthesis. 
High-performance kernels are critical for maximizing accelerator utilization in modern deep learning systems. Unlike general-purpose program synthesis, GPU kernel generation imposes strict execution requirements: generated kernels must compile successfully, produce numerically correct outputs, avoid prohibited high-level framework fallbacks, and deliver measurable runtime improvements on real hardware. Consequently, recent research has increasingly focused not only on code generation itself, but also on execution-based evaluation, reward verification, and training models to perform iterative error correction and repair.

\subsection{Vendor Libraries, Compilers, and DSLs}

Existing approaches to high-performance GPU computing largely fall into two paradigms.
The first relies on highly optimized vendor libraries such as NVIDIA's \textbf{cuBLAS} and \textbf{cuDNN}, as well as template-based frameworks like \textbf{CUTLASS}. 
While these libraries deliver excellent performance for canonical operations such as GEMM and convolution, extending them to support rapidly evolving model architectures, novel operators, and fused computation patterns often requires substantial manual engineering effort~\citep{jaber2026autokernelautonomousgpukernel, qimengkernel2025}. Moreover, modern deep learning models introduce new computation patterns (e.g., SwiGLU and Grouped-Query Attention) at a pace that frequently exceeds the update cycles of vendor-maintained libraries and template frameworks.

To address this inflexibility, the systems community has heavily invested in the second paradigm: \textbf{bare-metal code generation (or native kernel synthesis)}. Deep learning compilers and Domain-Specific Languages (DSLs) were proposed to generate low-level code directly from computational graphs~\citep{chen2018tvm, zheng2020ansor, ligerkernel2024, liu2023nnsmith, autotriton2025, guo2026drtritonlargescalesyntheticdata}. Methods like TVM~\citep{chen2018tvm} and Ansor~\citep{zheng2020ansor} map optimization spaces into Intermediate Representations (IRs), using search algorithms to find optimal implementations. Similarly, the Triton ecosystem~\citep{tillet2019triton} provides a higher abstraction for GPU programming, with recent works exploring LLM integration within this DSL space~\citep{ligerkernel2024, autotriton2025}. These pioneering systems demonstrated that bypassing standard libraries is essential for fusing operations and maximizing hardware utilization.

MusaCoder inherits this bare-metal synthesis philosophy. However, instead of generating code that calls cuDNN/CUTLASS or searching for schedules in explicit IRs, MusaCoder is trained to write \textbf{native CUDA/MUSA kernels from scratch}. By learning the mapping from PyTorch reference programs to underlying hardware instructions, MusaCoder helps developers support long-tail and novel operators without relying on standard libraries, directly mastering low-level kernel design and optimization.

\subsection{LLM-Driven Kernel Generation and Iterative Refinement}

Recent post-training methods for code and kernel generation have expanded from supervised fine-tuning to verifier-guided reinforcement learning, multi-turn self-correction, and agent search~\citep{selfrefine2023, reflexion2023, kevin2025, cudal12025, cudaagent2026, dice2026, cudaforge2025, tutoringcuda2025, dong2026kernelblastercontinualcrosstaskcuda, yang2025qwen3}. Works like \textbf{Kevin}~\citep{kevin2025} and \textbf{CUDA-L1}~\citep{cudal12025} show that mastering kernel generation requires learning from compiler, runtime, and performance feedback via RL. Meanwhile, frozen-model agent systems (e.g., \textbf{CUDA Agent}~\citep{cudaagent2026}, \textbf{CudaForge}~\citep{cudaforge2025}, \textbf{QiMeng-Kernel}~\citep{qimengkernel2025}, and \textbf{AutoKernel}~\citep{jaber2026autokernelautonomousgpukernel}) use LLMs at inference time to interact with profilers and tools. More broadly, software engineering frameworks like OpenHands and SWE-agent~\citep{openhands2024, sweagent2024}, along with self-correction studies~\citep{cycle2024, psrlcode2025, cure2025}, show the effectiveness of using test-time computation to improve code quality.

While agent-based methods achieve strong performance by relying on massive inference-time computation, MusaCoder internalizes these capabilities directly into the model weights through an SFT $\rightarrow$ single-turn RL $\rightarrow$ multi-turn RL pipeline. By embedding compiler and runtime feedback into the training loop, MusaCoder empowers much smaller models (e.g., 9B parameters) to surpass the accuracy of state-of-the-art closed-source models. This approach fundamentally eliminates the reliance on extensive test-time search, drastically reducing deployment and inference costs while retaining the model's ability to autonomously fix errors and optimize code.

\subsection{Executable Evaluation and Reward Verification}

Benchmarking and verifier design are essential for execution-driven generation. Frameworks like \textbf{KernelBench}~\citep{ouyang2025kernelbench} and \textbf{KernelBenchX}~\citep{kernelbenchx2026}, along with other executable evaluations~\citep{kevin2025, cudal12025, cudaagent2026, dice2026, tutoringcuda2025}, require outputs to pass actual compilation, random testing, and synchronized profiling, rather than just measuring text similarity. 

Furthermore, GPU kernel generation is prone to reward hacking, such as calling forbidden PyTorch operators, using asynchronous execution to bypass timers, or modifying benchmark settings. Therefore, recent systems emphasize strict anti-cheating mechanisms~\citep{kernelbenchx2026, cudal12025, cudaagent2026, cudaforge2025, qimengkernel2025}. MusaCoder adopts this approach. Our verifier categorizes outputs into compilation failures, runtime exceptions, correctness failures, cheating/fallback behaviors, and valid performance metrics. This structured feedback prevents evaluation loopholes and provides precise debugging information for multi-turn RL.

\subsection{Emerging Accelerator Ecosystems and MusaCoder's Positioning}

While CUDA remains the dominant ecosystem, software stacks for emerging alternative accelerators—such as Huawei's CANN~\citep{huawei-cann-docs}, Cambricon's BANG~\citep{cambricon-bangc-docs}, and Moore Threads' MUSA SDK~\citep{moorethreads-musa-docs}—are developing rapidly. A major systems challenge in these heterogeneous ecosystems is the high manual engineering cost of migrating and optimizing novel operators across different platforms. A model that can generate native kernels from scratch can significantly reduce this cross-platform development cost and mitigate vendor lock-in.
MusaCoder addresses this problem by training CUDA and MUSA kernel generators within a unified full-stack pipeline on Moore Threads GPUs, showing that emerging accelerators can support closed-loop LLM post-training for executable code generation.


\begin{figure}[ht]
    \centering
    \includegraphics[width=\linewidth, keepaspectratio]{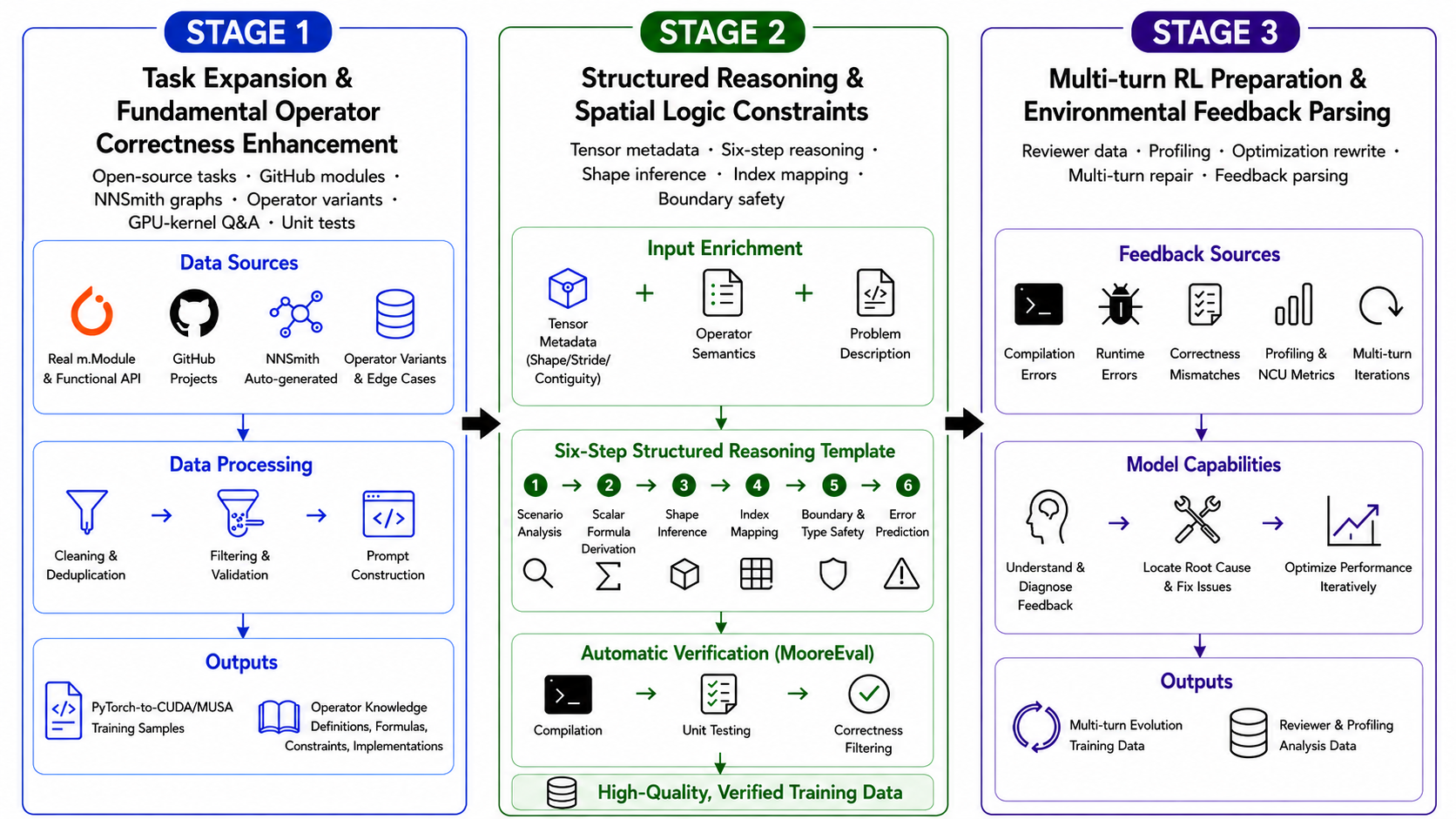}
    \caption{\textbf{Three-stage evolution of the SFT data construction pipeline.} Stage 1 expands the PyTorch-to-CUDA/MUSA workload distribution and aligns baseline operator correctness through real PyTorch modules, GitHub projects, NNSmith-generated graphs, operator variants, cleaning, filtering, and prompt construction. Stage 2 injects tensor metadata and enforces a six-step structured reasoning template to strengthen shape inference, index mapping, boundary handling, and semantic correctness, followed by MooreEval-based automatic verification. Stage 3 introduces execution-feedback understanding through compilation errors, runtime failures, correctness mismatches, profiling signals, and multi-turn iterations, producing data for feedback-driven repair, reviewer training, and performance optimization.} 
    \label{fig:sft_data_evolution}
\end{figure}

\section{Data Synthesis Pipeline}
\label{sec:data_pipeline}

Existing PyTorch-to-CUDA datasets are useful for KernelBench-style warmup, but they are not sufficient for full-stack post-training. They provide limited coverage of long-tail operators, often lack reusable verification assets, and may rely on vendor-library implementations rather than native kernels. MusaCoder therefore constructs SFT data as a staged capability-building process (Fig.~\ref{fig:sft_data_evolution}).

\begin{itemize}
    \item \textbf{Stage 1: Task Expansion and Fundamental Operator Correctness Enhancement.} We enlarge PyTorch-to-CUDA/MUSA task coverage with open-source tasks, GitHub modules, NNSmith-generated graphs, targeted basic-operator variants, GPU-kernel Q\&A data, and automatically generated unit tests.
    \item \textbf{Stage 2: Structured Reasoning and Spatial Logic Constraints.} We add explicit tensor metadata and a six-step reasoning template to reduce common failures in shape inference, scalar formulas, indexing, boundary handling, and fallback avoidance.
    \item \textbf{Stage 3: Multi-turn RL Preparation and Environmental Feedback Parsing.} We synthesize reviewer, profiling, optimization-rewrite, and multi-turn repair data so that the model can interpret compiler/runtime errors, correctness mismatches, and performance feedback before RL.
\end{itemize}

Through the aforementioned three-stage pipeline, MusaCoder's SFT data evolves from simple PyTorch-to-CUDA/MUSA translation pairs into a rich training corpus that integrates operator knowledge, structured reasoning, automated verification, and execution-feedback parsing. This pipeline not only improves the model's execution correctness in generating native CUDA/MUSA kernels, but also strengthens its ability to interpret compiler errors, runtime feedback, and performance bottlenecks. As a result, it provides a strong and stable initialization for subsequent RFT and RL training while preserving sufficient behavioral diversity for effective exploration.

\subsection{Task Expansion and Fundamental Operator Correctness Enhancement}

\subsubsection{PyTorch-to-CUDA/MUSA Workload Expansion}

This stage first expands the scale and distributional coverage of PyTorch-to-CUDA/MUSA generation tasks. Rather than relying on a narrow set of manually written operators, we construct reference workloads from three complementary sources: existing open-source PyTorch modules and kernel generation tasks~\citep{cudaagent2026, gope2024highly}, real-world PyTorch modules and functional API calls from GitHub repositories, and valid PyTorch computation graphs automatically synthesized by NNSmith~\citep{liu2023nnsmith}. Open-source data provides representative kernel-generation scenarios, GitHub data introduces realistic module structures and operator compositions, and NNSmith systematically supplements workload regions that are insufficiently covered by manually collected or real-world code.

For the GitHub data, we perform rigorous cleaning, deduplication, and filtering. We eliminate samples that depend on external states, contain unstable control flows, or are difficult to verify, retaining only reference programs that can be converted into executable PyTorch-to-CUDA/MUSA generation tasks.

We employ NNSmith to synthesize executable computation graphs under strict operator, shape, and dtype constraints. The operator pool covers activation functions, basic mathematical operations, logical operations, shape manipulations, pooling, reductions, normalization layers, convolution variants, interpolation, padding, type casting, and matrix operations. We further extend this pool to include 1D/3D operators, transposed convolutions, PixelShuffle, Adaptive Pooling, Dropout, Bmm, Addmm, and normalization variants, ultimately covering approximately 162 empirically tested operators. During generation, we specify operator subsets via \texttt{mgen.include}, control graph size via \texttt{mgen.max\_nodes}, and direct the generation of 1D, 2D, or 3D workloads via \texttt{mgen.rank\_choices}. This facilitates the construction of diverse graph archetypes, such as CNN-like, 3D convolutional, 1D sequential, reduction-heavy, and shape-sensitive computation graphs. All generated computation graphs are validated via the TorchScript backend and filtered based on executability, input-output stability, graph scale, and task adaptability before being converted into kernel generation prompts.

\subsubsection{Basic Operator Variant Augmentation}

Workload expansion alone does not resolve the model's weaknesses on fundamental operator semantics. Early experiments show that the model's accuracy is close to zero on several basic operator families, especially convolution variants. Under diverse parameter configurations, including varying strides, paddings, dilations, groups, kernel sizes, input ranks, 1D/2D/3D convolutions, and transposed convolutions, the model frequently fails in semantic interpretation, shape inference, boundary handling, and index mapping. Because these operators serve as essential building blocks for complex models and fused kernels, unstable correctness at this level would flood the subsequent RL stage with invalid samples and substantially weaken executable learning signals.

To address this issue, we construct large-scale variant data for fundamental operators. The dataset spans diverse dtypes, shapes, broadcast patterns, reduction dimensions, contiguous and non-contiguous layouts, and extreme input scenarios. Guided by failure distributions observed in preliminary validation, we upsample vulnerable operator families, including \texttt{nn.Conv2d}, transposed convolutions, 3D convolutions, reductions, normalizations, softmax, broadcasting, and complex indexing. This targeted augmentation is designed not merely to increase data volume, but to expose the model to the parameter regimes and boundary cases where basic operator generation most often breaks down.

The construction pipeline consists of ``Data Pre-processing $\rightarrow$ Two-Stage LLM Generation $\rightarrow$ Compilation and Correctness Filtering.'' In the pre-processing stage, we aggregate real-world PyTorch code snippets and low-level operator APIs. We employ Small Language Models (SLMs) for semantic cleansing to eliminate state-control APIs (e.g., \texttt{torch.set\_device()}), retaining exclusively pure computational operators as generation seeds. In the first stage of LLM generation, rather than directly outputting CUDA code, we prompt the LLM to design diverse implementation pathways for a given operator, including naive baselines, single-feature optimizations, algorithmic or input-specialized implementations, and composite optimized variants. Each pathway must explicitly define the thread mapping, memory hierarchy transitions, synchronization primitives, and resource bottleneck trade-offs, thereby forming a structured mechanism description. In the second stage, the LLM generates concrete CUDA kernels based on these mechanisms and target scenarios. Compared with generating code directly from operator descriptions, this ``plan-then-implement'' paradigm reduces monotonic template outputs and improves implementation diversity. Finally, we couple candidate implementations with automatically generated unit tests for compilation, execution, and correctness verification, filtering out samples that fail to compile, trigger runtime errors, exhibit numerical inconsistencies, or display suspicious speculative behaviors.

Overall, this stage establishes the data foundation for PyTorch-to-CUDA/MUSA tasks through two complementary interventions. Workload expansion broadens the distribution of reference programs and operator compositions, while basic operator variant augmentation directly targets the model's most severe fundamental correctness failures, especially convolution, reduction, normalization, broadcasting, and complex indexing. Together, they provide a more stable correctness baseline for the subsequent structured reasoning, multi-turn feedback, and RL phases.

\subsubsection{GPU Kernel Knowledge Q\&A Data}
\label{sec:prior_knowledge_qa}

Beyond direct PyTorch-to-CUDA/MUSA generation data, we construct a GPU Kernel Knowledge Question-and-Answer (Q\&A) dataset to supplement the model's foundational knowledge encompassing mathematical principles, CUDA programming models, GPU architectures, PyTorch tensor semantics, memory hierarchies, performance optimization strategies, numerical stability, and benchmarking methodologies. We observe that many errors in kernel generation do not merely stem from the code implementation itself, but rather originate from an inadequate comprehension of underlying concepts. Examples include erroneous thread indexing, neglected boundary checks, misjudged tensor layouts, mishandled reduction precision, or a lack of understanding regarding the impact of asynchronous CUDA execution on latency measurement. Therefore, this dataset aims to equip the model with more robust knowledge priors for kernel generation.

The questions in this dataset are primarily sourced from two pipelines. The first pipeline extracts data from open-source CUDA tutorials, GPU programming documentation~\citep{nvidia-cuda-guide}, kernel optimization examples, and common CUDA bug/FAQ repositories via rule matching, keyword extraction, and manual cleansing, predominantly covering general concepts and common optimization principles. The second pipeline synthesizes questions based on KernelBench task patterns and the model's failure cases from the previous kernel generation stage, serving to precisely bridge the model's exposed vulnerabilities. Specifically, we perform targeted augmentation for operator types that exhibit low accuracy or suboptimal performance during direct kernel generation, such as convolution, reduction, normalization, softmax, broadcasting, and complex shape indexing.

Distinct from direct kernel generation data, the knowledge Q\&A data does not require the model to output complete CUDA implementations. Instead, it trains the model to conduct deep causal reasoning: elucidating why a particular implementation is efficient, how underlying physical constraints restrict code topologies, and identifying the root causes of specific runtime errors. The questions encompass mathematical definitions, basic CUDA implementations and memory access optimizations, high-performance tuning, Tensor Core utilization and hardware adaptation, common bug debugging and bottleneck analysis, advanced applications, and future architectural evolutions. This data serves as a knowledge enhancement signal during SFT, forming a complement to the code generation data: the generation data teaches the model \textit{how} to write kernels, whereas the knowledge Q\&A data helps the model comprehend the underlying rationale for correctness and performance behind specific implementation choices.

\subsubsection{Automated Unit Test Generation and Verification}
\label{sec:unit_test_generation}

To guarantee the executability and correctness of the synthesized data, we construct an automated unit test generation and verification pipeline for PyTorch-to-CUDA/MUSA tasks. For a given PyTorch reference program, the code alone is insufficient to determine the correctness of the generated kernel; it necessitates test cases encompassing diverse input shapes, data types, value ranges, and edge cases. Therefore, we leverage Large Language Models (LLMs) to generate candidate test cases based on PyTorch code semantics and interface signatures, and execute them within a native PyTorch environment to obtain reference outputs (ground truth) for subsequent sample filtering and MooreEval evaluation. 

Specifically, test case generation primarily covers four dimensions of variation. First is \textbf{shape coverage}, including regular dimensions, unaligned dimensions, highly imbalanced dimensions, and shape combinations that trigger broadcasting. Second is \textbf{dtype coverage}, encompassing \texttt{FP32}, \texttt{FP16}, \texttt{BF16}, and common integer inputs. Third is \textbf{value range coverage}, which includes standard random distributions, mixed positive and negative values, zero-inclusive inputs, and extreme maximum/minimum values that may trigger overflow, underflow, or numerically unstable paths. Finally, \textbf{edge case coverage} involves scenarios such as empty tensors, all-zero/all-one inputs, non-contiguous tensors, and special dimensional configurations.

The generated candidate test cases are initially executed in the native PyTorch environment. We filter out test samples that exhibit syntax errors, invalid parameter combinations, runtime exceptions, out-of-memory (OOM) errors, or unstable outputs, retaining only those cases that execute stably and yield deterministic reference outputs. The retained test suite is subsequently utilized in two scenarios: first, during the SFT data construction phase for compiling and correctness-filtering candidate CUDA/MUSA implementations; second, within the subsequent MooreEval framework for random-input correctness checks and performance profiling of model-generated results. Through this pipeline, we mitigate the influx of erroneous implementations into the training data while providing a unified, reusable evaluation foundation for the ensuing RFT/RL phases.

In summary, this stage establishes the foundational data substrate across three dimensions: tasks, knowledge, and verification. The PyTorch-to-CUDA/MUSA data broadens the task distributions exposed to the model; the knowledge Q\&A data supplements the underlying CUDA and PyTorch semantic priors; and the automated unit tests provide a unified standard for data filtering and subsequent execution-based evaluations. The objective of this stage is not to directly pursue optimal performance, but rather to prioritize enhancing the executability, semantic correctness, and fundamental robustness of the generated results.

\subsection{Structured Reasoning and Spatial Logic Constraints}

\subsubsection{Standardized Data Structure for Six-Step Reasoning}
\label{sec:six_stage_format}

During direct CUDA kernel generation, models frequently bypass intermediate reasoning steps, leaping straight from PyTorch code to low-level implementations. This often results in logical disjoints across operator semantic comprehension, shape inference, index mapping, boundary handling, and numerical stability. To mitigate these logical disjoints, we introduce a six-step reasoning format into the SFT data, decomposing the PyTorch-to-CUDA/MUSA translation process into an ordered sequence of intermediate reasoning steps. The objective of this format is not to induce redundant explanations, but to compel the model to execute a structured derivation—from high-level operator semantics to thread-level memory accesses—prior to code generation.

The six-step reasoning structure is defined as follows:
\begin{description}
    \item[\textbf{Step 1: Holistic Analysis and Operator Deconstruction}] 
    Analyze the primary computational paradigm of the PyTorch code (e.g., element-wise operations, reductions, matrix multiplications, or spatial transformations), and explicitly list the involved inputs, outputs, and critical parameters such as \texttt{weight}, \texttt{bias}, \texttt{stride}, and \texttt{padding}. This phase establishes the mapping between PyTorch APIs and low-level kernel inputs, mitigating issues related to omitted parameters or misinterpreted operator semantics.

    \item[\textbf{Step 2: Scalar Mathematical Formula Derivation}] 
    Deconstruct tensor-level operations in PyTorch into the scalar computations required for a single output element or a single thread. For example, for normalization operators, the calculation formulas for mean, variance, normalization, and affine transformations must be explicitly defined. This phase facilitates the paradigm shift from tensor-level expressions to thread-level computational logic.

    \item[\textbf{Step 3: Shape and Dimensionality Calculus}] 
    Deduce the shape transitions across input, intermediate, and output tensors, clarifying the dimensional impact of operations such as broadcasting, reshape, transpose, slice, pooling, convolution, and reduction. For reduction operations, the reduction domain must also be explicitly determined. This phase provides the geometric basis for subsequent determinations of total thread counts, grid/block configurations, and output memory layouts.

    \item[\textbf{Step 4: Indexing Relations and Memory Mapping}] 
    Establish the bi-directional mapping between 1D global thread indices and multi-dimensional logical coordinates, and compute the corresponding linear memory offsets based on the tensor layout. For non-contiguous tensors, explicit address calculations incorporating strides are mandated. This phase is primarily designed to minimize erroneous indexing, out-of-bounds memory accesses, and incorrect dimensional unrolling.

    \item[\textbf{Step 5: Boundary Handling and Type Safety}] 
    Analyze edge cases requiring explicit handling within the kernel, such as out-of-bounds protection for trailing threads, padding regions, division-by-zero risks, low-precision accumulation errors, integer/floating-point type casting, and anomalous shapes. This phase guides the model to inject necessary defensive logic, thereby enhancing code robustness across diverse input conditions.

    \item[\textbf{Step 6: Error Pre-analysis and Negative Constraints}] 
    Summarize potential pitfalls in the current task based on historical failure modes—such as hardcoding dimensions, ignoring strides, omitting biases, mishandling broadcasting, inappropriately assuming contiguous memory layouts, or invoking disallowed PyTorch fallbacks. Serving as a pre-generation self-diagnostic step, this phase prevents the model from reiterating high-frequency errors.
\end{description}

Through this standardized format, we decompose end-to-end code generation into a six-dimensional intermediate reasoning process encompassing semantics, mathematics, shape, indexing, boundaries, and error pre-analysis. During the SFT phase, this format serves as high-quality reasoning traces incorporated into the training corpus, enabling the model to internalize a more stable and deterministic kernel generation workflow. In subsequent inference or multi-turn rectification scenarios, the model can leverage this structured intermediate representation for self-diagnosis and precise bug localization.

\subsubsection{Explicit Tensor Shape Annotation}
\label{sec:shape_augmentation}

We observe that models frequently stumble in PyTorch-to-CUDA/MUSA generation during shape-sensitive operations, such as \texttt{view}/\texttt{reshape}, \texttt{transpose}, broadcasting, slicing, pooling, convolution, and reduction. Once an error occurs in deducing the output shape or the layout of an intermediate tensor, the subsequent determinations of grid sizes, thread indices, memory offsets, and output write-back locations inherently cascade into failure. To mitigate such shape reasoning errors, we introduce an explicit shape annotation strategy during the construction of training samples: subsequent to the original PyTorch reference code but prior to the model's generation of the CUDA/MUSA kernel, we append a structured \texttt{[Tensor Shape Information]} block. This block explicitly enumerates the shapes, strides, and contiguity flags of critical inputs, intermediate nodes, and output tensors. Consequently, the model is directly exposed to explicit tensor layout constraints prior to kernel generation, rather than relying exclusively on implicit reasoning derived from the PyTorch code.

Specifically, we automatically extract tensor metadata from the PyTorch computation graph via predefined graph analysis scripts. PyTorch's dynamic graph mechanism, \texttt{torch.fx} graph capture, and shape propagation provide the foundational utilities for this precise metadata extraction~\citep{paszke2019pytorch, reed2022torchfx}. For tasks with static input dimensions, we employ a static analysis pipeline based on \texttt{torch.fx}: we first trace the computation graph via \texttt{torch.fx.symbolic\_trace}, and subsequently execute \texttt{ShapeProp} with dummy inputs to extract the concrete shape, stride, and \texttt{is\_contiguous} attributes from the \texttt{tensor\_meta} of each critical node. For tasks featuring dynamic input dimensions, we adopt a symbolic pipeline based on \texttt{torch.export}. We represent dynamic dimensions via \texttt{torch.export.Dim} and extract symbolic shape and stride expressions from the exported \texttt{ExportedProgram}. For instance, given a slice operation on an input of shape $[B, L]$, the system can record the output shape as $[B, L-1]$ while preserving the stride semantics of that sliced view.

In the final prompt, this annotation is inserted immediately following the PyTorch reference in a format akin to the following:
\begin{tcblisting}{
    colback=gray!5,
    colframe=gray!40,
    title=Shape-enhanced prompt block,
    fonttitle=\bfseries,
    breakable,
    listing only,
    listing engine=listings,
    listing options={
        language={},
        basicstyle=\ttfamily\small,
        breaklines=true,
        breakatwhitespace=true,
        columns=fullflexible
    }
}
[Tensor Shape Information]
- input x: shape=[B, L], stride=[L, 1], contiguous=True
- narrow/slice node: shape=[B, L-1], stride=[L, 1], contiguous=False
- cumsum node: shape=[B, L-1], stride=[L-1, 1], contiguous=True
- output: shape=[B, L], stride=[L, 1], contiguous=True
[Kernel Generation Hint]
Use the listed shapes and strides when deriving thread mapping and memory offsets.
Do not assume an intermediate tensor is contiguous unless marked contiguous=True.
\end{tcblisting}

This specific placement integrates the shape annotation intrinsically into the generation task itself, rather than treating it as a disjoint offline analysis log. For non-contiguous views, the prompt explicitly provides the strides and instructs the model to employ stride-based address calculations, averting the default assumption of contiguous memory layouts. For operations such as reshape, transpose, broadcast, and reduction, the annotation retains the input-output shape transitions of critical nodes, guiding the model in determining total thread counts, output coordinate unrolling schemes, and write-back indices.

This strategy is primarily utilized during SFT data construction, though it can also serve as explicit prompt augmentation during inference. During training, it provides the model with explicit tensor layout supervision, training it to align shapes, strides, and output layouts prior to writing the kernel. During inference, it reduces the model's misjudgments regarding complex shapes and non-contiguous memory layouts. Because we have thus far only conducted manual inspections on a limited subset of shape mismatch cases and their rectified outcomes, without yet formalizing systematic ablation studies, we do not present this as a standalone empirical claim in this paper. Instead, we provide a representative case study in the Appendix~\ref{app:shape} to demonstrate how explicit shape annotation facilitates the correction of indexing logic.

\subsection{Multi-turn RL Preparation and Environmental Feedback Parsing}

\subsubsection{Multi-turn Interactive Trajectory Synthesis and Context Pre-adaptation}
\label{sec:multi_turn_data}

During single-turn CUDA kernel generation, models frequently struggle to simultaneously satisfy compilation, numerical correctness, and performance constraints. Compared to static PyTorch-to-CUDA/MUSA samples, the environmental feedback returned by the evaluation sandbox—such as textual compilation/runtime errors and quantitative performance metrics—provides explicit rectification signals. To lay the empirical foundation for subsequent multi-turn Reinforcement Learning (RL), we synthesize multi-turn iterative interaction trajectories during this stage. This not only trains the model to parse environmental feedback and localize functional or micro-architectural bugs, but crucially allows it to pre-adapt to the multi-turn conversational context and dependency structures, thereby executing targeted self-correction.

Specifically, we adopt a ``$5+3$'' multi-turn interactive rollout mechanism. For each task, the model initially generates a candidate kernel, which is submitted to the MooreEval sandbox for verification. If the candidate implementation fails compilation or correctness checks, the subsequent prompt injects the error logs, numerical discrepancies, or failure rationales from the preceding turn, guiding the model to execute bug rectification. Conversely, if the candidate implementation is functionally correct, subsequent turns pivot towards performance optimization, prompting the model to refine memory access patterns, thread partitioning, synchronization strategies, or kernel fusion based on the current implementation. By default, a maximum of 5 iterative turns is permitted; to enhance the coverage of hard tasks, an additional allowance of up to 3 rectification attempts is granted.

Regarding trajectory retention, we mandate that the final-turn output must pass correctness verification, and we prioritize retaining trajectories that exhibit significant performance gains relative to their historical correct versions. Specifically, if a trajectory contains multiple correct implementations, we designate the first correct version as the performance baseline and require the final version to achieve a speedup of at least $1.2\times$. This criterion prevents the training set from being inundated with protracted rectification trajectories that lack optimization dividends. Concurrently, failure samples induced by environmental anomalies, timeouts, or non-model-related infrastructure issues are meticulously purged during the filtering phase to prevent the injection of noisy feedback.

During training, the historical turns within the multi-turn trajectories function primarily as contextual inputs, supplying error diagnostics and evolutionary topologies; the specific loss masking strategies will be detailed comprehensively in the SFT training section. Fundamentally, this data compels the model to transcend the mere imitation of static code snapshots, enabling it to master dynamic parsing and refactoring driven by environmental feedback. This paradigm establishes an optimal policy initialization and context adaptation for the subsequent closed-loop, multi-turn RL training phase.

\subsubsection{CUDA Kernel Reviewer}
\label{sec:kernel_reviewer_data}

To elevate the model's diagnostic proficiency regarding the correctness and performance bottlenecks of CUDA kernels, we construct specialized CUDA Kernel Reviewer data. Taking a PyTorch reference code and a candidate CUDA kernel as inputs, this data tasks the model with verifying whether the candidate implementation aligns with the PyTorch semantics. If an error is detected, the model must provide a root-cause analysis alongside the rectified code; if the candidate kernel is functionally correct, the model must further profile its potential performance bottlenecks. In contrast to direct kernel generation data, the Reviewer data does not primarily train the model to author kernels \textit{de novo}. Instead, it compels the model to comprehend the underlying causality—why a kernel succeeds or fails, and how it can be systematically rectified or optimized.

Each Reviewer sample comprises two input components: the PyTorch reference code, which delineates the target computational semantics, and the candidate CUDA kernel, which serves as the implementation under scrutiny. The model's output adheres to a standardized format. Initially, it conducts a correctness analysis, verifying whether the mathematical computations, shape inference, index mapping, boundary conditions, dtype handling, and synchronization logic are strictly congruent with the PyTorch reference. If the kernel exhibits flaws, the model must pinpoint the specific anomalies, elucidate the root causes and their downstream impacts, and generate the complete rectified code. Conversely, if the kernel is functionally correct, the model must analyze its performance characteristics and optimization headroom from micro-architectural perspectives, including memory access patterns, thread-level parallelism, warp divergence, occupancy, and arithmetic intensity. Finally, the model is mandated to output a standardized categorical conclusion: \texttt{VERDICT: CORRECT} or \texttt{VERDICT: INCORRECT}.

This data systematically cultivates four core capabilities. First, \textbf{functional equivalence verification}, which assesses whether the CUDA kernel accurately manifests the mathematical semantics and tensor shape logic of the PyTorch reference. Second, \textbf{common CUDA bug identification}, encompassing erroneous thread indexing, out-of-bounds memory accesses, mismatched grid/block configurations, omitted synchronizations, invalid non-contiguous tensor addressing, and improper handling of dtypes or accumulation precision. Third, \textbf{bug rectification capability}, which entails synthesizing a corrected kernel aligned with the PyTorch reference upon successful bug localization. Fourth, \textbf{performance profiling capability}, which involves identifying latent bottlenecks within functionally correct kernels—such as uncoalesced memory accesses, excessive global memory transactions, sub-optimal shared memory utilization, low warp efficiency, or branch divergence.

Within the training pipeline, the Reviewer data acts as auxiliary SFT data, forming a synergistic complement to the direct PyTorch-to-CUDA/MUSA generation data. While the generation data trains the model on \textit{how to author kernels}, the Reviewer data trains it on \textit{how to audit kernels}. This auditing proficiency is critically foundational for the subsequent multi-turn rectification and execution feedback training: when the model receives environmental signals such as compilation errors, correctness mismatches, or performance feedback, it must inherently comprehend the underlying root causes to synthesize a logically sound rectified version. Consequently, this data also serves as the infrastructural basis for cultivating subsequent critique/judge models or synthesizing multi-turn refinement trajectories.

\subsubsection{Performance Feedback Comprehension and Targeted Optimization Refactoring}
\label{sec:profiling_and_optimization_data}

In PyTorch-to-CUDA/MUSA tasks, once the generated outputs pass correctness verification, the subsequent objective shifts from ``authoring correct kernels'' to ``authoring efficient kernels.'' However, relying solely on scalar execution time feedback often leaves the model ill-equipped to diagnose the root causes of performance bottlenecks—such as uncoalesced memory accesses, excessive global memory transactions, sub-optimal shared memory utilization, low occupancy, excessive register pressure, warp divergence, or underutilized Tensor Cores. To elevate the model's proficiency in comprehending performance feedback and executing optimization refactoring, we construct two categories of complementary data: one for learning to parse profiling feedback, and the other for learning targeted refactoring from naive kernels to optimized kernels.

The first category comprises profiling-based performance analysis data. We utilize tools such as PyTorch Profiler, Nsight Systems, and Nsight Compute to collect kernel execution telemetry~\citep{paszke2019pytorch, nvidia-nsight-systems, nvidia-nsight-compute}, subsequently extracting the metrics most pertinent to optimization decisions from the raw reports. Because complete NCU reports typically contain voluminous redundant fields that would inject contextual noise into training samples, we filter the data to retain exclusively the Top-$K$ critical metrics. For memory-bound kernels, emphasis is placed on metrics such as L2/DRAM accesses, memory throughput, and memory efficiency; for compute-bound kernels, the focus shifts to SM utilization, occupancy, instruction mix, Tensor Core activity, and primary stall reasons. Anchored on these structured metrics, we formulate two distinct tasks: (1) given a CUDA kernel, the model must predict its latent bottlenecks and performance characteristics; (2) given a CUDA kernel alongside its empirical profiling results, the model must analyze the bottlenecks and propose concrete optimization directives. Appendix~\ref{app:ncu_data} provides the corresponding data template, report-extraction procedure, and a condensed example.

The second category encompasses targeted optimization refactoring data. Taking functionally correct yet poorly performing CUDA kernels as inputs, these samples require the model to generate highly optimized implementations while strictly preserving the original PyTorch semantics. The candidate naive kernels typically exhibit prevalent anti-patterns, such as excessive \texttt{atomicAdd} operations, uncoalesced memory accesses, redundant global memory passes, excessive kernel launches, sub-optimal thread block configurations, superfluous \texttt{cudaDeviceSynchronize()} invocations, and the failure to exploit shared memory or warp-level primitives. The corresponding optimization objectives include memory coalescing, shared memory reductions, warp shuffles, kernel fusion, loop unrolling, mitigating branch divergence, and tuning block/grid configurations.

During training, these two data categories serve distinct yet complementary functions. The profiling data trains the model to deduce performance bottlenecks from hardware metrics or runtime behaviors, whereas the optimization refactoring data further trains the model to operationalize these diagnostic insights into concrete code modifications. For instance, in operators such as reduction, softmax, normalization, or channel-wise scaling, a naive implementation might rely on multiple global memory passes or extensive atomic operations. Conversely, the optimized variants typically necessitate block-level reductions, shared memory reuse, warp-level reductions, or kernel fusion to minimize memory and synchronization overheads. Through this data, the model learns not only ``where the bottlenecks lie'' but also ``how to architect the optimizations,'' thereby furnishing a superior performance optimization initialization for the subsequent multi-turn feedback and RL phases.

\section{Method}
\subsection{Overview}

The training pipeline of MusaCoder is orchestrated across three progressive stages: supervised warmup, task alignment, and execution feedback-driven reinforcement learning (RL).

Initially, we leverage the multi-source data synthesized in the preceding section to conduct multi-task Supervised Fine-Tuning (SFT). This phase equips the model with proficiency in PyTorch-to-CUDA/MUSA task formats, prevalent native kernel implementation patterns, CUDA extension boilerplates, bug reviewing, feedback comprehension, and performance profiling. The primary objective of this stage is not to directly search for optimal kernels, but rather to establish robust code generation priors for the subsequent verifiable training phases.

Following the supervised warmup, we conduct Rejection Sampling Fine-Tuning (RFT) to align the model closer to the final PyTorch-to-CUDA/MUSA generation tasks. Specifically, starting from the SFT checkpoint, we sample multiple candidate implementations for each PyTorch workload and utilize MooreEval to filter for positive samples that are parsable, compilable, numerically correct, and satisfy task constraints. Unlike standard RFT, which typically retains only the single best solution, we adopt a \textit{diversity-preserving filtering} strategy: we cluster the correct implementations generated under the same prompt and randomly sample supervisory targets from this positive pool during training. This approach enhances the correctness of the RFT samples while preventing the model from prematurely collapsing into a narrow set of fixed implementation templates, thereby preserving the requisite exploration space for the subsequent RL phase.

Unlike open-ended code generation, kernel generation tasks can derive programmatic feedback through actual compilation, execution, correctness verification, and performance measurement. \textbf{MooreEval} not only verifies whether candidate codes compile successfully, align their outputs with PyTorch references, and achieve empirical performance gains, but also enforces a rigorous anti-hacking protocol. Through static rules and runtime profiling, it detects forbidden PyTorch/\texttt{aten::*} compute fallbacks, preventing the model from bypassing custom kernel authoring by merely invoking off-the-shelf PyTorch operators within \texttt{ModelNew.forward()}. Only candidate solutions that execute the core computations via genuine native kernels—while strictly satisfying both correctness and legitimacy constraints—are granted positive rewards.

Building upon the structured verification telemetry returned by MooreEval, we formulate kernel generation as a verifiable reinforcement learning problem, executing single-turn RL and multi-turn feedback RL sequentially. Single-turn RL directly optimizes the model's capacity to author native kernels on the first attempt without environmental feedback. Conversely, multi-turn feedback RL further trains the model to agentically leverage actual execution feedback for iterative bug rectification and performance optimization. To robustly orchestrate this closed-loop training paradigm, we propose three critical mechanisms: \textbf{PrimeEcho} maintains the optimization pressure on the first-turn generation quality while utilizing multi-turn rectification signals, thereby balancing the final success rate with inference efficiency; \textbf{Buffered Dynamic Retry (BDR)} transforms \textit{all-failed groups} into learnable rectification tasks augmented with execution feedback, mitigating reward sparsity on hard samples; 
\textbf{MirrorPop} introduces a novel sequence-level off-policy metric that more accurately estimates policy drift magnitude, enabling reliable masking of severely off-policy samples for stable RL optimization.
Collectively, these three mechanisms guarantee \textbf{effective exploration, objective consistency, and update stability} during RL: PrimeEcho prevents the multi-turn objective from deviating from first-turn generation capabilities, BDR recovers training signals from long-tail failure samples, and MirrorPop curtails the instability induced by off-policy updates.

\subsection{Supervised Warmup and Task Alignment}

\subsubsection{Multi-task Supervised Fine-Tuning}
\label{sec:multi_task_sft}

After completing data synthesis, we first conduct Supervised Fine-Tuning (SFT) on the base LLMs to initialize the subsequent alignment tasks. Through SFT, the model learns code formats, reasoning processes, and response structures from high-quality samples. 
Without this phase, the model would generate a massive volume of unparsable, uncompilable, interface-mismatched, or semantically flawed candidate kernels during the early stages of RL, leading to extreme reward sparsity.

Instead of training exclusively on standalone PyTorch-to-CUDA/MUSA translation tasks, we construct a multi-task SFT corpus spanning several complementary data categories. Kernel generation data provides direct code synthesis capabilities; Reviewer data improves semantic error detection and bug correction; Profiling/NCU data helps the model to interpret performance feedback and identify execution bottlenecks; Knowledge QA data strengthens understanding of GPU programming concepts and PyTorch tensor semantics; and optimization rewrite data teaches the transformation of inefficient implementations into optimized kernels. Through this multi-task supervision, the model learns not only how to generate kernels, but also how to analyze, debug, and optimize them.

\subsubsection{Data Mixture and Prior Diagnostics}
\label{sec:sft_data_mixture}

To align the SFT data mixture more closely with the final PyTorch-to-CUDA/MUSA generation objective, we construct a small-scale prior diagnostic set before training to evaluate the base model's initial proficiencies across different operator families. Specifically, we execute a single forward pass on a batch of PyTorch reference models using \texttt{torch.profiler.profile}~\citep{paszke2019pytorch}, extract the \texttt{aten::*} operators invoked during actual execution, and perform statistical aggregation and resampling based on operator categories to construct a small evaluation set with a balanced operator distribution. Subsequently, we deploy the base model to generate CUDA/MUSA kernels on this diagnostic set and utilize MooreEval to measure the compilation rates, correctness rates, and performance metrics across different operator categories.

These diagnostic results guide the data sampling strategy during the SFT phase. For operator families where the base model exhibits weakness—such as convolutions, reductions, normalizations, softmax, broadcasting, and complex indexing—we upsample the corresponding kernel generation, reviewer, optimization rewrite, and knowledge QA data. Conversely, for relatively stable tasks such as simple element-wise operations or activations, we appropriately downsample their proportions. This prevents uniform training across all tasks, allowing SFT to focus on bridging the model's capability gaps for the final evaluation tasks.

Furthermore, to prevent the model from overfitting to the specific output style of kernel code, we retain approximately $30\%$ general instruction-following and coding data within the SFT mixture. This portion of data primarily serves to maintain the base model's generalized coding capabilities and instruction-following proficiency, mitigating issues such as format rigidness or degraded generalization after prolonged training.

\subsubsection{Loss Masking for Multi-turn Samples}
\label{sec:sft_loss_masking}

For multi-turn execution feedback data, we align the training format as closely as possible with actual inference scenarios: historical turns serve exclusively as contextual inputs, and the supervisory signal is derived solely from the model's final-turn output. Specifically, intermediate code attempts, environmental feedback, compilation errors, runtime exceptions, and rectification processes from preceding turns are retained in the context but are masked from the loss computation. Concurrently, the reasoning content from historical turns is removed, retaining only the essential code, error feedback, and interactive states. Only the final-turn model response retains the necessary reasoning content and the ultimately submittable code, which are then utilized to compute the cross-entropy loss.

This loss masking strategy prevents the model from imitating failed implementations, incomplete code snippets, redundant debugging telemetry, or intermediate outputs that deviate from the final submission format. Meanwhile, the historical turns still provide crucial task evolution and error feedback information, training the model to directly generate high-quality final results given the task description and verifier feedback, rather than merely reproducing the entire trial-and-error trajectory.

\subsubsection{Diversity-Preserving Rejection Fine-Tuning}
\label{sec:diversity_preserving_rft}

Following multi-task SFT, we conduct Rejection Sampling Fine-Tuning (RFT) to further align the model with the definitive PyTorch-to-CUDA/MUSA generation task. Distinct from the SFT phase, RFT exclusively utilizes direct PyTorch-to-CUDA/MUSA generation samples. For each PyTorch workload, we sample multiple candidate CUDA/MUSA implementations using the SFT model. These candidates are then submitted to our evaluation sandbox, MooreEval (detailed comprehensively in Section~\ref{sec:moore_eval}), for parsing, compilation, correctness verification, anti-hacking detection, and performance benchmarking. Candidate solutions exhibiting parsing failures, compilation errors, numerical inaccuracies, runtime instability, or suspicious hacking behaviors are systematically discarded. Conversely, CUDA/MUSA kernels that successfully pass verification are retained as novel supervisory targets to further fine-tune the model.

In preliminary experiments, we attempted a greedy RFT approach, retaining only the single best verified response per prompt. While this strategy effectively enhanced the model's alignment with the primary task format, it severely constricted the output distribution. We empirically observed that following this ``one-positive'' RFT, the model's generation entropy plummeted rapidly, collapsing to approximately one-tenth of the SFT checkpoint's entropy. Such entropy collapse is highly detrimental to the subsequent RL phase: the model becomes excessively prone to imitating a narrow subset of high-frequency implementation templates. This degradation in sampling diversity inherently stifles the model's capacity to explore diverse kernel implementations, thread partitioning schemes, and optimization strategies during RL.

To mitigate this pathology, we employ a \textbf{diversity-preserving rejection filtering} strategy. Specifically, we initially retain all positive samples that pass correctness verification for a given prompt, rather than exclusively selecting the fastest or singular optimal implementation. Subsequently, we cluster these correct implementations based on a rich set of micro-architectural and structural features, including code similarity, Abstract Syntax Tree (AST) topology, kernel count, grid/block configurations, and the utilization of shared memory, atomic operations, kernel fusion, or divergent reduction strategies. Finally, we select representative samples from each cluster, constraining the positive pool for each prompt to a bounded capacity of 4 to 8 trajectories. This methodology systematically filters out erroneous implementations while preserving a heterogeneous distribution of valid implementations for the same computational workload.

During RFT training, instead of supervising each prompt with a single fixed target, we randomly sample solutions from its corresponding positive pool at each epoch. This reformulates the RFT objective from imitating a single canonical implementation to modeling a diverse distribution of valid solutions. 
By improving correctness while mitigating rapid entropy collapse, the approach preserves sufficient implementation diversity for downstream RL training.

\subsection{MooreEval: Verifier and Reward Environment}
\label{sec:moore_eval}

In the PyTorch-to-CUDA/MUSA kernel generation task, model outputs cannot be evaluated merely via textual similarity or static rules; they must undergo authentic compilation, execution, correctness verification, and performance benchmarking. More crucially, during RL training, the evaluation system ceases to be a mere offline benchmarking utility; it transfigures into the \textit{reward environment} within the closed-loop training pipeline. A large number of candidate code snippets in each batch must be compiled and executed reliably and efficiently, with their telemetry subsequently fed back into the policy optimization process. Following executable-feedback benchmarks for kernel and code generation~\citep{ouyang2025kernelbench}, we designed \textbf{MooreEval}, serving as the unified executable evaluation environment for both RFT data filtering and RL training.

The architectural objectives of MooreEval encompass four pillars: \textbf{high throughput}, \textbf{robustness}, \textbf{reward interpretability}, and \textbf{anti-hacking resilience}. Given a PyTorch reference and the model-generated \texttt{ModelNew} implementation, MooreEval parses and compiles the candidate code, subjects it to correctness testing with randomized inputs, executes anti-hacking detection, and performs synchronized speed profiling. It subsequently transcribes the outcomes into structured verification telemetry and scalar rewards. Distinct from single-node synchronous evaluation scripts, MooreEval employs a distributed heterogeneous pipeline that decouples CPU-side compilation from GPU-side execution. Compile Workers handle the pre-compilation of candidate codes, Exec Workers execute correctness, anti-cheat, and speed tests within isolated sandboxes, and the Reward Adapter bridges the final telemetry into the RL training framework. This decoupled architecture allows CPU compilation resources and GPU execution resources to scale independently, preventing GPUs from idling during compilation and thereby optimally sustaining large-batch reward computations during RL.

Given that model-generated CUDA/MUSA codes are inherently untrusted, MooreEval incorporates rigorous failure isolation and fault tolerance mechanisms. Candidate codes may induce Python syntax errors, NVCC compilation failures, runtime illegal memory accesses, GPU kernel timeouts, or catastrophic Python process crashes. To prevent an anomalous sample from derailing the entire training loop, MooreEval orchestrates task execution via task queues, inflight states, visibility timeouts, bounded retries, and sandbox subprocess isolation. If a worker crashes or a task remains unacknowledged beyond the timeout threshold, the system automatically requeues the task. Conversely, if the error is deterministically caused by the candidate code itself—such as syntax errors, illegal memory accesses, or numerical mismatches—MooreEval immediately returns the corresponding error taxonomy. Additional system implementation details are provided in Appendix~\ref{app:moore_eval_system}.

\subsubsection{Structured Verification Protocol}
\label{sec:structured_verification}

MooreEval employs a rigorous tiered verification protocol for each candidate implementation. First, the system extracts executable code from the model output and verifies whether it defines a \texttt{ModelNew} class satisfying the interface requirements. Samples that are unparsable, lack critical interfaces, or exhibit invalid code structures are immediately deemed invalid. Subsequently, the candidate implementation is compiled in an isolated environment. If a Python syntax error, NVCC compilation error, linking error, or compilation timeout occurs, MooreEval returns the corresponding compilation failure category, preempting the allocation of downstream GPU execution resources.

Following successful compilation, MooreEval executes the candidate implementation on randomized inputs and aligns it against the PyTorch reference. The correctness check encompasses output shapes, dtypes, and numerical errors, with numerical verification anchored by \texttt{torch.allclose} and task-specific tolerance settings. MooreEval adopts a correctness-first verification sequence: only candidate solutions that pass correctness verification proceed to the anti-hacking detection and performance testing phases. This guarantees that performance rewards are strictly predicated on functional correctness, preventing erroneous implementations from acquiring illusory speedups by skipping computations or exploiting anomalous timings.

\textbf{Anti-hacking detection} is a core verification mechanism within MooreEval. To ensure that the model generates fully native CUDA/MUSA kernels rather than relying on high-level PyTorch fallbacks, MooreEval combines static analysis with runtime profiling to detect \texttt{aten::*} operations invoked during \texttt{ModelNew.forward()}. The system maintains an allowlist (detailed in Appendix~\ref{app:op_whitelist}) that permits only non-computational utilities such as shape queries, view/reshape operations, memory copies, and tensor creation. If prohibited high-level PyTorch/\texttt{aten::*} operations are detected—including matrix multiplications, convolutions, reductions, or other high-level PyTorch operators—the candidate implementation is flagged as \texttt{cheating:disallowed\_aten} and receives zero reward regardless of numerical correctness. In addition, MooreEval performs supplementary checks for anomalously large speedups to prevent spurious rewards caused by invalid implementations or timing manipulations.

Finally, MooreEval conducts performance benchmarking on candidate solutions that pass both the correctness and legitimacy checks. Speed measurement utilizes CUDA events~\citep{nvidia-cuda-guide}, executing repeatedly post-warmup while explicitly synchronizing the CUDA device to minimize variance introduced by asynchronous execution, initialization overheads, and stochastic fluctuations. Following this pipeline, each candidate implementation is reduced to a structured verification tuple:
\begin{equation}
V(c, x) =
(
\mathrm{compiled},
\mathrm{correct},
\mathrm{legal},
\mathrm{speedup},
\mathrm{category},
\mathrm{detail}
),
\end{equation}
where $c$ represents the candidate code extracted from the model output, and $x$ represents the task prompt. ``$\mathrm{compiled}$'' indicates whether the candidate code compiled successfully, ``$\mathrm{correct}$'' denotes whether the output aligns with the PyTorch reference, ``$\mathrm{legal}$'' indicates passage of the anti-hacking detection, ``$\mathrm{speedup}$'' represents the speedup ratio relative to the PyTorch baseline, and ``$\mathrm{category}$'' alongside ``$\mathrm{detail}$'' record failure telemetry such as compilation errors, runtime exceptions, numerical mismatches, prohibited operators, or infrastructural anomalies.

In subsequent training phases, MooreEval generates two distinct categories of signals based on $V(c, x)$. The first is a scalar score $s(c)$, utilized for RFT filtering and RL reward computation; this score is deterministically governed by the candidate implementation's compilation status, correctness, legitimacy, and speedup. The second is the textual feedback $f(c)$, utilized for multi-turn feedback rollouts; this is synthesized from the error telemetry within ``$\mathrm{category}$'' and ``$\mathrm{detail}$'', yielding actionable information such as compilation failure summaries, shape/dtype mismatches, numerical error statistics, runtime exceptions, or forbidden \texttt{aten::*} invocation warnings.

Consequently, MooreEval does not merely return a binary pass/fail verdict; rather, it simultaneously provisions structured verification states, scalar rewards, and interpretable textual feedback. This tiered architecture establishes an explicit hierarchy of training priorities: it primarily guarantees the executability of the candidate implementation, subsequently ensures numerical correctness and adherence to native kernel constraints, and ultimately awards supplementary rewards predicated on empirical performance gains. Concurrently, it supplies interpretable bug rectification signals essential for multi-turn RL.

\subsubsection{Reward Design}
\label{sec:moore_eval_reward}

Based on the structured verification telemetry returned by MooreEval, we map each candidate implementation to a scalar reward for RL training. The reward employs a \textit{correctness-gated} design: performance metrics contribute to the reward computation only if the candidate implementation compiles successfully, passes correctness verification, and does not trigger forbidden PyTorch/\texttt{aten::*} fallbacks. This ensures that the training optimization strictly prioritizes executability, semantic correctness, and native kernel legitimacy, with speedup treated as a secondary objective. In other words, the model cannot acquire illusory positive rewards via erroneous outputs, skipped computations, PyTorch fallbacks, or anomalous timing manipulations.

Let $q \in [0,1]$ denote the correctness rate of the candidate implementation across randomized tests, and let $\nu$ denote the runtime speedup relative to the PyTorch baseline. For a candidate code $c$, the reward function $s(c)$ is defined as:
\begin{equation}
s(c)=
\begin{cases}
-1, & \text{if code extraction fails, compilation fails, or runtime fails}, \\
-1, & \text{if a disallowed PyTorch/\texttt{aten::*} fallback is detected}, \\
-1, & \text{if } q = 0, \\
-0.5 + 0.5q, & \text{if } 0 < q < 1, \\
1 + \lambda \cdot \min(\max(\nu - 1, 0), \nu_{\max}), & \text{if } q = 1 \text{ and the implementation is legal}.
\end{cases}
\end{equation}

Here, $\lambda$ controls the weight of the performance reward, and $\nu_{\max}$ clips the impact of anomalously high speedups. This formulation provides a bounded shaping signal for partially correct implementations, enabling the model to transition smoothly from completely unusable to partially correct states. However, strictly positive rewards and speedup bonuses are exclusively reserved for fully correct and legal native kernels. Candidate implementations that trigger forbidden operations or exhibit anomalous speedups are treated as hard failures by MooreEval, ensuring the training objective remains strictly aligned with the constraints of native kernel generation.

Unlike Reinforcement Learning from Human Feedback (RLHF)~\citep{christiano2017deep, stiennon2020learning, ouyang2022training, bai2022training}, which relies on human preferences, the reward here is derived entirely from programmatic verification encompassing compilation, execution, correctness, legitimacy, and performance metrics. Consequently, MooreEval provides consistent, reproducible, and interpretable environmental feedback for large-scale RL training.

\subsubsection{Feedback Generation for Multi-turn Training}
\label{sec:moore_eval_feedback}

Beyond outputting scalar rewards, MooreEval translates structured error taxonomies into natural language feedback for multi-turn feedback training. This feedback is not manually authored; rather, it is synthesized automatically from the evaluation telemetry, ensuring a consistent feedback format across diverse tasks and failure modes.

For compilation failures, the feedback includes a condensed error summary, such as missing headers, type mismatches, invalid function signatures, or the specific locations of NVCC errors. For runtime failures, the feedback specifies issues like CUDA launch errors, illegal memory accesses, out-of-memory (OOM) exceptions, or execution timeouts. For correctness failures, the feedback details output shape mismatches, dtype mismatches, or numerical discrepancies, providing summaries of the maximum error, mean error, or failing inputs whenever possible. For samples triggering anti-hacking detection, the feedback explicitly identifies the forbidden \texttt{aten::*} operations and reminds the model that it must execute the core computation using custom native kernels. For functionally correct but underperforming samples, the feedback returns the runtime, speedup, and potential performance bottlenecks to guide optimization in subsequent turns. Detailed prompt templates are provided in Appendix~\ref{app:multidialogue}.

During multi-turn training, after the model generates candidate code $c_t$ at turn $t$, MooreEval returns the structured verification result $V_t$, the scalar score $s_t$, and the textual feedback $f_t$. If the candidate solution passes verification, the rollout can terminate early. Otherwise, the feedback $f_t$ is appended to the context for the subsequent turn, guiding the model to generate a rectified version $c_{t+1}$. Therefore, MooreEval functions not merely as a reward function, but as a critical data engine for multi-turn bug rectification and performance optimization. Through this feedback mechanism, the model learns to localize issues based on actual execution outcomes and systematically refine its code implementation in subsequent turns.

\subsection{Reinforcement Learning}
\label{sec:rl_training}

For each task prompt $x$, the policy model $\pi_\theta$ generates a candidate implementation $c$; MooreEval then extracts the candidate code from the output, executes compilation and verification, and returns the structured telemetry $V(c,x)$ alongside a scalar reward $s(c)$. Consequently, the RL phase does not rely on human preference annotations; instead, it directly leverages programmatic execution feedback to optimize the model. This paradigm elevates the probability of generating compilable, numerically correct, legal, and high-performance kernels while strictly satisfying the native kernel constraints.

Our RL training pipeline is bifurcated into two stages. The first stage is a \textbf{single-turn RL warmup}, wherein candidate solutions are generated and evaluated in a single pass for each task. The policy is updated using the rewards returned by MooreEval, with the objective of elevating the model's first-attempt generation capability under zero-feedback conditions. The second stage is \textbf{multi-turn feedback RL}. When a candidate solution fails, the structured error feedback synthesized by MooreEval is appended to the context, guiding the model to execute bug rectification or performance optimization in subsequent turns. Because the ultimate deployment and evaluation remain primarily focused on the quality of the first-turn generation, we architect PrimeEcho (first-turn-anchored multi-turn reward). This mechanism enables multi-turn feedback to provision supplementary exploration signals while preventing the model from becoming overly reliant on subsequent rectifications.

\subsubsection{Single-turn RL Warmup}
\label{sec:single_turn_rl}

During the single-turn RL phase, generation and evaluation are executed exclusively in a single turn per task. Given a prompt $x$, the current policy $\pi_\theta$ samples candidate implementations $c_i$, and MooreEval computes the corresponding rewards $r_i=s(c_i)$. For a given prompt, we sample a group of $G$ candidate solutions and construct the response-level advantage using intra-group normalized rewards:

\begin{equation}
\hat A_i =
\frac{
r_i - \operatorname{mean}(\{r_j\}_{j=1}^{G})
}{
\operatorname{std}(\{r_j\}_{j=1}^{G})+\epsilon
}
\end{equation}

Subsequently, we update the policy using Group Relative Policy Optimization (GRPO)~\citep{shao2024deepseekmath, guo2025deepseek}. Given the prompt $x$, let $o_i=(o_{i,1},\ldots,o_{i,L_i})$ denote the $i$-th response, where $o_{i,t}$ is its $t$-th token and $o_{i,<t}$ is the previously generated prefix. 
During training, the importance ratio of the target policy relative to the rollout policy is defined as:

\begin{equation}
\rho_{i,t}(\theta)
=
\frac{
\pi_\theta(o_{i,t}\mid x,o_{i,<t})
}{
\pi_{\mathrm{rollout}}(o_{i,t}\mid x,o_{i,<t})
}
\end{equation}

The corresponding optimization objective is defined as:
\begin{equation}
\mathcal{L}_{\mathrm{GRPO}}(\theta)
=
-
\frac{1}{\sum_{i=1}^{G} L_i}
\sum_{i=1}^{G}
\sum_{t=1}^{L_i}
\min \Big(
\rho_{i,t}(\theta)\hat A_i,\,
\mathrm{clip}\big(\rho_{i,t}(\theta), 1-\epsilon_{\mathrm{low}}, 1+\epsilon_{\mathrm{high}}\big)\hat A_i
\Big)
+
\beta \, D_{\mathrm{KL}}(\pi_\theta \| \pi_{\mathrm{ref}}).
\end{equation}
Here, $L_i$ denotes the length of the $i$-th response, $\hat A_i$ is the response-level advantage, $\pi_{\mathrm{ref}}$ is the reference policy, and $\beta$ controls the strength of KL regularization. 
The parameters $\epsilon_{\mathrm{low}}$ and $\epsilon_{\mathrm{high}}$ define the lower and upper clipping margins of the importance ratio, yielding an asymmetric clipping range $[1-\epsilon_{\mathrm{low}}, 1+\epsilon_{\mathrm{high}}]$.

This stage directly optimizes the model's first-turn generation capability, encouraging it to produce compilable, numerically correct, and fully native CUDA/MUSA kernels that satisfy the anti-hacking constraints. 
Unlike the subsequent multi-turn feedback stage, this warmup phase does not condition generation on iterative compiler or execution feedback, thereby forcing the model to improve its zero-shot kernel synthesis ability.

\subsubsection{Multi-turn feedback RL}
\label{sec:multi_turn_rl}

While single-turn GRPO optimizes for the reward after the initial generation, many failures in CUDA/MUSA kernel synthesis are rectifiable. For instance, candidate code may fail due to missing header files, mismatched function signatures, shape inference errors, improper dtype handling, numerical overflows, or the inadvertent invocation of forbidden \texttt{aten::*} fallbacks. Since MooreEval provides not only scalar rewards but also granular error taxonomies and execution logs, we can reinject this feedback into the model's context to guide iterative refinement.

\begin{figure}[htbp]
\centering

\begin{minipage}[t]{0.48\textwidth}
    \centering
    \begin{minipage}[c][0.28\textheight][c]{\linewidth}
        \centering
        \includegraphics[
            width=\linewidth,
            height=0.28\textheight,
            keepaspectratio
        ]{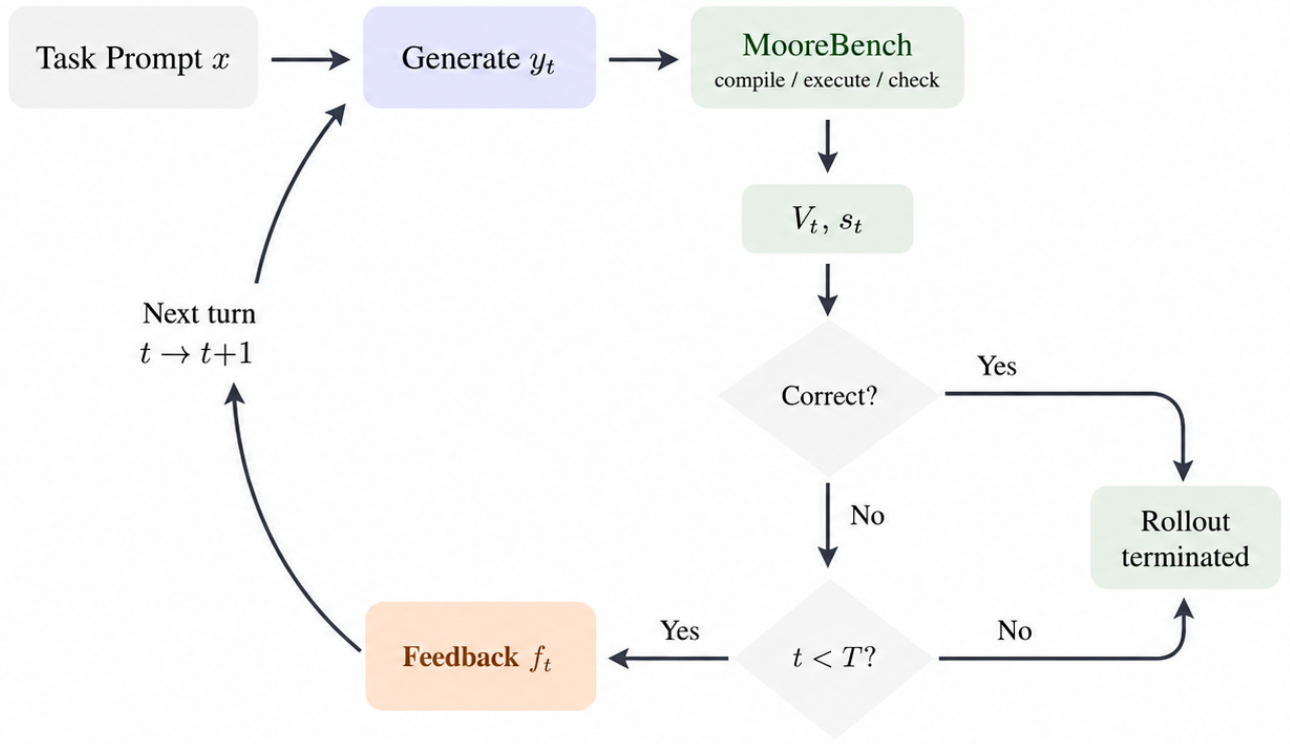}
    \end{minipage}
    \caption{Multi-turn RL rollout process.}
    \label{fig:multi_turn_rollout}
\end{minipage}
\hfill
\begin{minipage}[t]{0.48\textwidth}
    \centering
    \begin{minipage}[c][0.28\textheight][c]{\linewidth}
        \centering
        \includegraphics[
            width=\linewidth,
            height=0.28\textheight,
            keepaspectratio
        ]{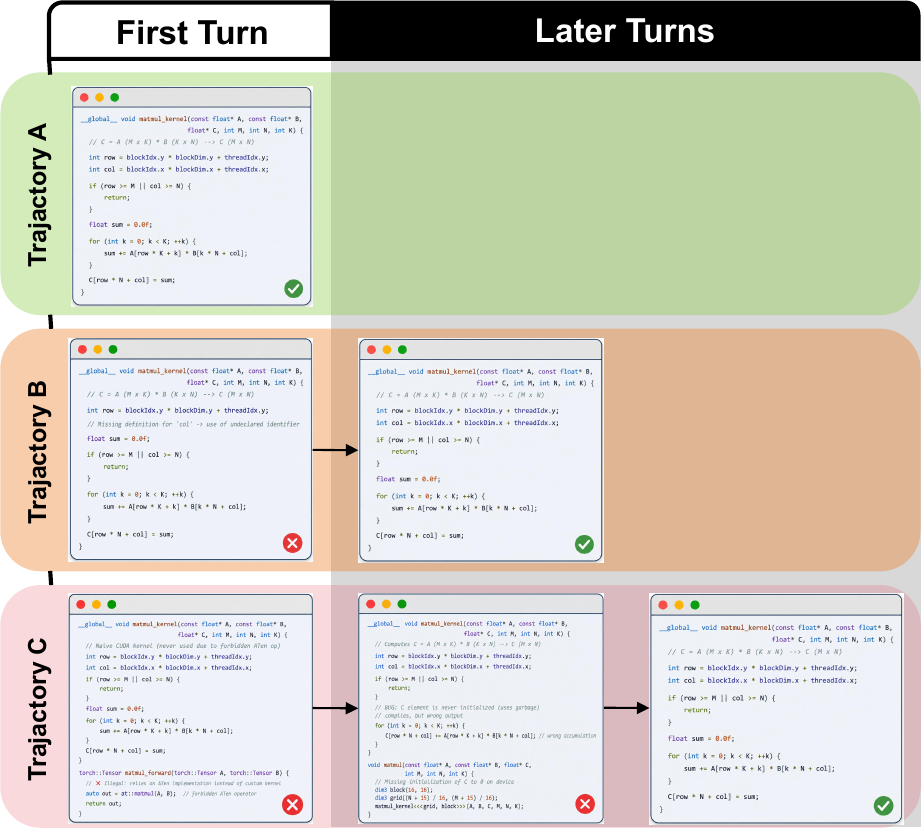}
    \end{minipage}
    \caption{Multi-turn reward design.}
    \label{fig:reward_design}
\end{minipage}

\end{figure}

As illustrated in Figure~\ref{fig:multi_turn_rollout}, given a task prompt $x$, the $i$-th rollout trajectory initiates at turn $k=1$ with the model outputting $y_{i,1}$, from which we extract the candidate code:
\begin{equation}
c_{i,1} = \mathrm{Extract}(y_{i,1}).
\end{equation}
MooreEval subjects $c_{i,1}$ to compilation, correctness verification, anti-hacking detection, and performance benchmarking, yielding the structured verification result:
\begin{equation}
V_{i,1} = V(c_{i,1}, x).
\end{equation}
From this result, we derive the scalar score $s_{i,1} = s(c_{i,1})$ and synthesize the feedback text $f_{i,1} = f(c_{i,1})$ by aggregating error categories and log summaries. If $c_{i,1}$ passes verification, the rollout terminates prematurely. Otherwise, the feedback $f_{i,1}$ is appended to the context, prompting the model to generate a rectified output $y_{i,2}$. This process iterates until a candidate solution passes verification or the maximum iteration limit $K$ is reached.

The $i$-th multi-turn trajectory is formally represented as:
\begin{equation}
\tau_i = (x, y_{i,1}, c_{i,1}, V_{i,1}, s_{i,1}, f_{i,1}, y_{i,2}, c_{i,2}, V_{i,2}, s_{i,2}, f_{i,2}, \ldots, y_{i,K}, c_{i,K}, V_{i,K}, s_{i,K}).
\end{equation}
Here, $y_{i,k}$ denotes the model output at turn $k$, $c_{i,k}$ the extracted candidate, $V_{i,k}$ the MooreEval verification telemetry, $s_{i,k}$ the scalar reward, and $f_{i,k}$ the aggregated textual feedback. The rollout ceases upon successful verification or when the iteration limit is exhausted.

The objective of multi-turn rollout is to cultivate the model's capacity for agentic, iterative coding rather than merely optimizing for final-turn success. For complex kernel tasks, a model may struggle to simultaneously satisfy compilation, correctness, legitimacy, and performance constraints in the first pass. The execution feedback from MooreEval enables the model to systematically rectify compilation errors, shape/dtype mismatches, numerical deviations, and suboptimal implementations.

However, the quality of the first-turn generation remains paramount. While multi-turn inference improves final success rates, each interaction turn incurs additional compilation, execution, and verification latency. If the model can generate a correct and efficient native kernel in the first attempt, it drastically reduces inference overhead. Therefore, our goal is to enhance the model's rectification capabilities without fostering over-reliance on environmental feedback. Directly optimizing for the final-turn or best-turn reward would inherently bias the model toward multi-turn success, incentivizing weaker first-turn generation followed by heavy reliance on subsequent MooreEval feedback for rectification.

To mitigate this objective drift, we introduce \textbf{PrimeEcho}, a first-turn-anchored multi-turn reward mechanism. It recontextualizes multi-turn feedback not as the sole target for final success, but as a supplementary exploration signal for first-turn generation. By leveraging the rectification traces revealed in subsequent turns while maintaining primary optimization pressure on first-turn generation, PrimeEcho achieves a balanced equilibrium between agentic multi-turn rectification proficiency and the low-latency efficiency of first-attempt kernel synthesis.

\subsubsection{PrimeEcho: First-turn-anchored Multi-turn Reward}
\label{sec:first_turn_anchored_reward}

The core philosophy of PrimeEcho is to explicitly decouple the initial generation quality from subsequent rectification gains within the trajectory-level reward. Given a group of $G$ multi-turn trajectories $\{\tau_i\}_{i=1}^{G}$ sampled for a single prompt, let $c_{i,k}$ denote the candidate code at turn $k$ for the $i$-th trajectory, and let $s_{i,k}=s(c_{i,k})$ denote its corresponding MooreEval score. We define the trajectory-level reward as:

\begin{equation}
R_{\tau_i}
=
\alpha s_{i,1}
+
(1-\alpha)\max_{1 \le k \le K} s_{i,k}
+
b_{\mathrm{early}}(\tau_i)
\end{equation}

where $s_{i,1}$ represents the score of the first-turn generation, $\max_{1 \le k \le K} s_{i,k}$ represents the peak score achieved across the entire multi-turn interaction, and $\alpha \in [0,1]$ dictates the emphasis placed on the initial performance. A larger $\alpha$ skews the training closer to single-turn RL, whereas a smaller $\alpha$ permits the subsequent rectification outcomes to provide a stronger auxiliary signal.

The early-success bonus $b_{\mathrm{early}}(\tau_i)$ exclusively rewards initial successes achieved within the first two turns, thereby incentivizing the model to synthesize a correct solution as expeditiously as possible. Specifically, let $\mathrm{success}(c_{i,k})$ denote an indicator function representing whether the candidate code $c_{i,k}$ passes MooreEval verification. The bonus is formulated as:

\begin{equation}
b_{\mathrm{early}}(\tau_i)
=
\beta_1 \mathbf{1}[\mathrm{success}(c_{i,1})]
+
\beta_2 \mathbf{1}[\neg \mathrm{success}(c_{i,1}) \land \mathrm{success}(c_{i,2})]
\end{equation}

where $\beta_1 > \beta_2 \ge 0$. This term guarantees that among trajectories that ultimately succeed, those achieving success earlier accumulate higher rewards.

Formally, PrimeEcho interpolates between a strict first-turn reward and a best-turn reward:

\begin{equation}
\alpha=1
\Rightarrow
R_{\tau_i}=s_{i,1}+b_{\mathrm{early}}(\tau_i),
\qquad
\alpha=0
\Rightarrow
R_{\tau_i}=\max_{1\le k\le K}s_{i,k}+b_{\mathrm{early}}(\tau_i)
\end{equation}

Thus, $\alpha$ explicitly governs the trade-off between first-turn generation quality and multi-turn rectification signals. By default, we adopt a relatively large $\alpha$, ensuring that multi-turn feedback serves predominantly as a supplementary exploration signal during training, rather than eclipsing the primary objective of first-turn generation.

Finally, for multiple trajectories generated under the same prompt, we utilize $R_{\tau_i}$ to compute the intra-group normalized advantage:

\begin{equation}
\hat A_i =
\frac{
R_{\tau_i} - \operatorname{mean}(\{R_{\tau_j}\}_{j=1}^{G})
}{
\operatorname{std}(\{R_{\tau_j}\}_{j=1}^{G}) + \epsilon
}
\end{equation}

This formulation is structurally identical to the response-level advantage in single-turn GRPO, substituting the single-turn reward with the PrimeEcho trajectory reward. As illustrated in Figure~\ref{fig:reward_design}, we predominantly apply this advantage to the tokens comprising the first-turn output, effectively subordinating the multi-turn feedback to serve the enhancement of first-turn generation.

\subsection{RL Stabilization Techniques}
\label{sec:rl_stabilization}

While MooreEval enables RL training to directly optimize compilation success, numerical correctness, and runtime performance, execution-based RL still faces two major stability challenges in practice. 
First, complex kernel-generation tasks frequently produce an \textit{all-failed group}, where all candidate implementations sampled for the same prompt fail to obtain a positive reward. 
In group-relative optimization, such groups provide little discriminative learning signal, since the intra-group advantages cannot reliably distinguish useful directions for policy improvement. 
Second, off-policy drift can arise between the rollout policy used to generate samples and the policy being updated during training. 
If left uncontrolled, severely off-policy samples can increase gradient variance and induce unstable or biased policy updates.

To address these challenges, we introduce two complementary stabilization components. 
\textbf{Buffered Dynamic Retry (BDR)} strengthens learning signals on hard samples by converting all-failed groups into feedback-conditioned retry tasks, while \textbf{MirrorPop} improves training stability by identifying and masking high-risk off-policy samples at the sequence level.

\subsubsection{Buffered Dynamic Retry}
\label{sec:buffered_dynamic_retry}

During RL training with GRPO, $G$ candidate implementations are typically sampled for a given prompt $x$, and the advantage is computed based on intra-group rewards. However, for challenging kernel generation tasks, an \textit{all-failed group} frequently emerges, wherein all candidate solutions fail:
\begin{equation}
r_i = -1 \quad \text{for any } 1 \le i \le G
\end{equation}
Within such an \textit{all-failed group}, the absence of correct positive samples deprives the intra-group rewards of meaningful discriminability, failing to provide a valid optimization direction. Consequently, the model repeatedly samples failed implementations on long-tail, difficult tasks. Furthermore, our empirical observations indicate that the most prevalent failure modes during both the training phase and on the validation set are verification errors (i.e., kernel precision inaccuracies) and runtime exceptions (i.e., execution bugs). Therefore, a transitional training mechanism bridging single-turn and multi-turn RL is important—one designed to exclusively target these formidable challenges with minimal training overhead. To this end, we propose \textbf{Buffered Dynamic Retry (BDR)}, which dynamically transfigures these hard samples into feedback-conditioned rectification tasks, thereby recovering actionable training signals.

An auxiliary motivation for this mechanism is to investigate whether empowering the model to rectify its own errors can simultaneously elevate its \textit{first-turn generation} capability (i.e., zero-shot code generation). In practical implementation, the core of the multi-turn training described later serves primarily to provide exploration signals; during policy updates, only the first-turn tokens are subjected to optimization. This is done to prevent the degradation of \textit{first-turn generation} quality, though it inherently sacrifices the direct enhancement of the model's bug-rectification capabilities. Conversely, BDR considers all tokens during updates. It explicitly trains the model to reflect upon and rectify errors, internalizing the root causes of failures. This allows us to explore whether combining BDR with subsequent multi-turn training can augment the model's intrinsic CUDA domain knowledge without compromising first-turn generation efficacy.

Specifically, given a training dataset $\mathcal{D}_{\mathrm{RL}}$, upon the conclusion of each rollout epoch, we scan the MooreEval telemetry within the current batch to isolate all-failed prompts. For each hard prompt, the system selects a representative erroneous candidate $c^{-}$ and extracts the structured error telemetry returned by MooreEval—such as correctness errors, runtime exceptions, compilation failures, or forbidden \texttt{aten::*} fallbacks. Subsequently, we concatenate the original task description, the failed code, and the error feedback into a \textit{feedback-conditioned prompt}:
\begin{equation}
x' = \mathrm{Compose}(x, c^{-}, f^{-})
\end{equation}
where $f^{-}$ denotes the error summary or rectification hint automatically synthesized by MooreEval. In doing so, an originally intractable zero-shot generation task is structurally transformed into a feedback-driven code rectification task, significantly easing the model's exploration toward a correct trajectory.

To prevent these feedback-augmented samples from distorting the primary task distribution, we refrain from immediately replacing the original training data. Instead, we enqueue $x'$ into a dynamic buffer pool $\mathcal{B}$. The buffer utilizes a First-In-First-Out (FIFO) eviction policy to constrain its capacity:
\begin{equation}
\mathcal{B} \leftarrow \mathrm{FIFOUpdate}(\mathcal{B}, x')
\end{equation}
We maintain the buffer at a fixed, predefined capacity. When a new sample is eligible for insertion and the pool is full, the oldest sample is evicted. This mechanism ensures the continuous retention of contemporary hard samples reflecting the model's current failure boundaries, while purging outdated error trajectories. This is crucial because the policy model evolves continuously during training; an early failure sample may no longer pose a challenge in later stages. The FIFO design mitigates the interference of \textit{stale feedback} on current policy updates, guaranteeing that the buffer consistently approximates the frontier of the model's current capabilities.

In subsequent training iterations, we sample feedback-conditioned prompts from the buffer pool with a small probability $p_{\mathrm{buf}}$, mixing them with prompts from the original dataset $\mathcal{D}_{\mathrm{RL}}$:
\begin{equation}
x_{\mathrm{train}} \sim
\begin{cases}
\mathcal{B}, & \text{with probability } p_{\mathrm{buf}} \\
\mathcal{D}_{\mathrm{RL}}, & \text{with probability } 1-p_{\mathrm{buf}}
\end{cases}
\end{equation}

By default, the overwhelming majority of samples continue to originate from the original single-turn tasks, with only a marginal fraction drawn from the dynamic buffer. The rationale is that BDR intrinsically converts a subset of first-turn generation tasks into feedback-driven rectification tasks. If deployed extensively or during the early stages of training, it could introduce task interference, fostering an over-reliance on feedback and subsequently eroding the first-turn generation capabilities. Thus, we primarily activate this mechanism during the later stages of RL, deploying it as a stabilization protocol to conquer long-tail hard samples rather than as a substitute for the primary single-turn training objective.

Crucially, we rely exclusively on the native execution feedback returned by MooreEval to guide retries, deliberately avoiding explicit rewriting prompts generated by external, more capable models. While external models can synthesize more articulate natural language explanations, such feedback frequently diverges from the data distribution established during the SFT phase. This divergence risks introducing out-of-distribution (OOD) contexts, precipitating entropy fluctuations and policy instability during training. By contrast, native sandbox feedback originates directly from the actual execution environment; it is structurally stable, reproducible, and seamlessly aligns with the authentic feedback the model will encounter in the subsequent multi-turn RL phase.

\subsubsection{MirrorPop Off-policy Sequence Masking}
\label{sec:mirrorpop}

In addition to reward sparsity, off-policy mismatch is another major source of instability in RL training. 
During training, the rollout policy used to generate samples may diverge from the current policy being optimized. 
Such drift can arise from train--inference backend discrepancies, policy changes introduced by reusing the same rollout batch across multiple mini-batch updates, or discrete distributional shifts caused by routing perturbations in MoE models. 
When the mismatch becomes severe, the resulting importance ratios can produce high-variance gradient estimates and unstable policy updates. 
Therefore, it is important to identify and filter high-risk samples that deviate substantially from the rollout policy.

A straightforward way to control off-policy updates is to apply token-level clipping or hard filtering to the importance ratio $\rho_{i,t}$, as in Truncated Importance Sampling (TIS)~\citep{yao2025offpolicy} and IcePop~\citep{icepop2025}. 
However, similar to token-level PPO clipping~\citep{schulman2017ppo}, such methods operate on individual tokens and therefore do not directly capture whether an entire response has drifted substantially from the rollout distribution~\citep{li2026trust}. 
This limitation is particularly relevant for long-form kernel generation, where moderate token-level deviations can accumulate into a severe sequence-level mismatch.

To address this issue, sequence-level masking introduces a binary mask $M_i \in \{0,1\}$ for each response. 
When $M_i=0$, the entire response is excluded from the policy-gradient update; when $M_i=1$, the response is retained and contributes normally to optimization. 
The GRPO objective augmented with sequence-level masking is then formulated as:

\begin{equation}
\mathcal{L}_{\mathrm{GRPO\text{-}mask}}(\theta)
=
-\frac{1}{\sum_{i=1}^{G} M_i L_i}
\sum_{i=1}^{G} 
\sum_{t=1}^{L_i}
\min\!\Bigl(
\rho_{i,t}(\theta)\hat A_i,\,
\mathrm{clip}\bigl(\rho_{i,t}(\theta),1-\epsilon_{\mathrm{low}},1+\epsilon_{\mathrm{high}}\bigr)\hat A_i
\Bigr) M_i
+
\beta D_{\mathrm{KL}}(\pi_\theta \| \pi_{\mathrm{ref}})
\end{equation}

The key design question is how to determine the response-level mask $M_i$. 
A common sequence-level off-policy masking strategy defines a response-level importance ratio using the length-normalized geometric mean of token-level ratios~\citep{liu2025speed, liu2025deepseek}:

\begin{equation}
\bar{\rho}^{\mathrm{geo}}_i
=
\left(
\prod_{t=1}^{L_i}
\rho_{i,t}
\right)^{1/L_i}
=
\exp\left(
\frac{1}{L_i}
\sum_{t=1}^{L_i}
\log \rho_{i,t}
\right)
\end{equation}

The sequence is then retained or discarded according to whether 
$\bar{\rho}^{\mathrm{geo}}_i$ falls within a predefined trust interval. 
However, this geometric-mean statistic is equivalent to exponentiating the signed mean log-ratio, and therefore suffers from directional cancellation. 
For example, consider two tokens with importance ratios $\rho_1=10$ and $\rho_2=0.1$. 
Both tokens exhibit large deviations from the rollout policy, but their signed log-ratio deviations cancel:
$
\log \rho_1 + \log \rho_2
=
\log 10 + \log 0.1
=
0.
$
As a result, the corresponding geometric mean becomes close to $1$, causing the conventional sequence-level mask to regard the response as nearly on-policy. 
This judgment is misleading: an over-weighted token and an under-weighted token offset each other algebraically in log space, but such cancellation does not imply that the sequence is safe for training.

Figure~\ref{fig:cancellation_case} illustrates this effect with a real example. 
Red and green tokens denote positive and negative log-ratio deviations, respectively. 
When $\log \rho_t$ is directly accumulated or averaged across the sequence, deviations with opposite signs can cancel, obscuring the true magnitude of sequence-level policy mismatch. 
In severe cases, a response containing strong local deviations may still appear nearly on-policy under the signed statistic. 
For off-policy filtering, however, the relevant quantity is the magnitude of deviation from the rollout policy rather than its signed direction. 
Thus, signed aggregation can underestimate the risk of high-mismatch trajectories.

To address this cancellation problem, we propose \textbf{MirrorPop}. 
Instead of distinguishing whether a token is assigned a higher or lower probability by the current policy than by the rollout policy, MirrorPop measures the direction-invariant magnitude of the deviation. 
For each token, we define the \textit{mirror deviation} as:

\begin{equation}
m_{i,t}
=
\max(\rho_{i,t}, \rho_{i,t}^{-1})
=
\exp(|\log \rho_{i,t}|),
\qquad m_{i,t}\ge 1
\end{equation}

This formulation is symmetric with respect to reciprocal ratios: both $\rho_{i,t}=10$ and $\rho_{i,t}=0.1$ are mapped to the same deviation magnitude, since 
$\max(10, 1/10)=\max(0.1, 1/0.1)=10$. 
Aggregating this direction-invariant deviation over the response, we define the length-normalized MirrorPop statistic as:

\begin{equation}
\bar m^{\mathrm{geo}}_i
=
\left(
\prod_{t=1}^{L_i}
\max(\rho_{i,t},\rho_{i,t}^{-1})
\right)^{1/L_i}
=
\exp\!\left(
\frac{1}{L_i}\sum_{t=1}^{L_i}|\log \rho_{i,t}|
\right)
\end{equation}

The corresponding sequence-level mask is thus defined as:

\begin{equation}
M_i^{\mathrm{mirrorpop}}
=
\mathbf{1}
\left[
\frac{1}{L_i}
\sum_{t=1}^{L_i}
|\log \rho_{i,t}| \le \delta
\right]
\end{equation}
where $\delta$ denotes the threshold in log-ratio space.

The key distinction between MirrorPop and conventional sequence-level masking lies in how token-level ratios are aggregated. 
Conventional masking relies on the signed average 
$\frac{1}{L_i}\sum_t \log \rho_{i,t}$, whereas MirrorPop uses the absolute-deviation average 
$\frac{1}{L_i}\sum_t |\log \rho_{i,t}|$. 
As a result, positive and negative deviations no longer cancel; instead, both contribute to the sequence-level mismatch score. 
Revisiting the example $\rho=[10,0.1]$, the conventional signed statistic gives 
$\frac{1}{2}(\log 10+\log 0.1)=0$, corresponding to a geometric mean ratio of $1$. 
In contrast, MirrorPop gives 
$\frac{1}{2}(|\log 10|+|\log 0.1|)=\log 10$, equivalently assigning a ratio-space mismatch magnitude of $10$. 
Thus, MirrorPop correctly identifies the response as a high-risk off-policy sample.

In RL training with GRPO, we use $M_i=M_i^{\mathrm{mirrorpop}}$ as the response-level masking criterion. 
MirrorPop complements token-level ratio clipping by operating at a different granularity: token-level clipping limits the update contribution of individual tokens, whereas MirrorPop determines whether the entire response has drifted too far from the rollout policy to be safely used for optimization. 
In this sense, MirrorPop measures ``how far'' a response deviates from the rollout policy, rather than ``in which direction'' it deviates. 
This magnitude-based sequence filtering mitigates the underestimation caused by signed log-ratio cancellation, enabling more reliable removal of severely off-policy samples and improving the stability of RL updates.

\section{Experiments}

\subsection{Training Details}
\label{sec:training_details}

During the supervised training phase, we fine-tune two base checkpoints: MusaCoder-9B is initialized from Qwen3.5-9B\footnote{https://qwen.ai/blog?id=qwen3.5}, and MusaCoder-27B is initialized from Qwen3.6-27B\footnote{https://qwen.ai/blog?id=qwen3.6-27b}. Unless otherwise specified, both model scales follow the same training pipeline. We use the AdamW\citep{loshchilov2017decoupled} optimizer with a learning rate of $1 \times 10^{-5}$, a warmup ratio of $3\%$, weight decay of $0.01$, and \texttt{bf16} precision. The maximum sequence length is set to $40$K, and the global batch size is $256$, training for $1$ epoch. For multi-turn SFT samples, we apply the loss masking strategy detailed in Section~\ref{sec:sft_loss_masking}: historical turns serve exclusively as contextual input, and the loss is computed solely on the model's final-turn response. The SFT implementation is built on DeepSpeed, incorporating ZeRO/offload techniques to support long-context supervised fine-tuning\citep{rajbhandari2020zero}.

The reinforcement learning phase employs a two-stage GRPO pipeline. The first stage, single-turn GRPO, aims to enhance the model's capacity to directly generate correct, legal, and efficient kernels under zero-feedback conditions. The second stage proceeds from the single-turn checkpoint into multi-turn RL, introducing online feedback from MooreEval as a rectification signal for subsequent turns. A multi-turn rollout spans a maximum of $3$ model response turns; the trajectory terminates early if a candidate implementation passes verification at any turn.

Across both RL stages, the rollout group size is set to $8$, with a training batch size of $64$. Rollouts are executed using SGLang in async mode. The training sampling parameters are temperature $0.9$ and top-$p=0.95$, while validation sampling uses temperature $0.7$ and top-$p=0.7$. Multi-turn RL employs the PrimeEcho reward with $\alpha=0.75$ by default. Notably, multi-turn RL calculates the policy gradient loss exclusively on the first-turn model response; subsequent turns participate only in trajectory evaluation and reward computation.

The RL training infrastructure is built on Megatron + SGLang\citep{shoeybi2019megatron, zheng2024sglang}, where Megatron manages the distributed training of the actor and reference models, and SGLang handles high-throughput asynchronous rollouts. Tensor parallelism is introduced to distribute model parameters and mitigate activation memory overhead, especially for the 27B model. The RL learning rate is set to $1 \times 10^{-6}$, with a warmup ratio of $0.1$, weight decay of $0.1$, and gradient clipping at $0.5$. The maximum prompt length is $8$K, and the maximum response length is $32$K.

All experiments are conducted on 64 Moore Threads MTT S5000 machines, each equipped with $8 \times 80$GB accelerator cards\footnote{\url{https://www.mthreads.com/product/S5000}}. This cluster robustly sustains the end-to-end training loop, encompassing long-context SFT, asynchronous rollouts, MooreEval online verification, and GRPO policy updates. Specifically, executing both supervised fine-tuning and execution-feedback reinforcement learning for 9B and 27B models on this domestic hardware environment demonstrates that the platform can support not only standard LLM fine-tuning but also complex RL training workloads involving large-scale code generation, compile-and-execute feedback, and online reward computation.

\subsection{Evaluation Setup}
\label{sec:eval_setup}

We evaluate MusaCoder's PyTorch-to-CUDA/MUSA kernel generation capabilities on the KernelBench benchmark, employing the unified MooreEval protocol to validate all models. Unlike the original evaluation scripts, we strictly require candidate implementations to successfully parse and compile, pass shape, dtype, and numerical correctness checks against the PyTorch reference on randomized inputs, and refrain from invoking forbidden PyTorch/\texttt{aten::*} computational fallbacks within \texttt{ModelNew.forward()}. Only candidate implementations that pass both correctness and legality checks proceed to performance testing. Performance measurements are conducted using repeated runs after warmup and timed via synchronized CUDA events to reduce variance from asynchronous execution and initialization overhead.

For each task, we sample $8$ candidate implementations at temperature $0.7$ and report two categories of metrics. \textbf{Pass Rate} measures correctness. Pass@8 indicates whether at least one of the $8$ sampled candidates passes MooreEval verification, while Avg.@8 reports the average fraction of verified candidates among the $8$ samples. \textbf{Faster Rate} measures performance improvement among valid native kernel implementations. A candidate is counted as faster only if it first passes correctness and legality checks and then achieves a runtime speedup larger than $1.1\times$ over the corresponding baseline. We report Faster Rate against both PyTorch eager execution and \texttt{torch.compile}. The $1.1\times$ threshold filters out marginal runtime differences that may arise from measurement noise.

We compare MusaCoder with both frontier code models and its underlying base models. The external baselines include Claude Opus 4.7, GLM-5.1, Kimi K2.6, and DeepSeek-V4-Pro/ProMax. We also report Qwen3.5-9B and Qwen3.6-27B to quantify the improvement brought by MusaCoder's SFT/RFT and RL pipeline over the initial base checkpoints. All models are evaluated under identical prompt templates, sampling configurations, validation scripts, and hardware environments. Unless otherwise specified, all main results are reported across Overall, Level 1, Level 2, and Level 3 categories.

\begin{table}[t]
\centering
\small
\caption{
KernelBench evaluation results under the strict MooreEval protocol. We report correctness-oriented Pass Rate metrics and performance-oriented Faster Rate metrics across Overall, Level 1, Level 2, and Level 3. Pass@8 denotes whether at least one of 8 sampled candidates passes verification, while Avg.@8 measures the average correctness rate among the 8 samples. Faster Rate is reported against PyTorch eager execution~\citep{paszke2019pytorch} and \texttt{torch.compile}~\citep{pytorch2}; a candidate is counted as faster only if it passes correctness and legality checks and achieves speedup larger than $1.1\times$ over the corresponding baseline.
}
\label{tab:kernelbench_main_results}

\resizebox{\textwidth}{!}{
\begin{tabular}{lcccccccc}
\toprule
\multirow{2}{*}{Model}
& \multicolumn{2}{c}{Overall}
& \multicolumn{2}{c}{Level 1}
& \multicolumn{2}{c}{Level 2}
& \multicolumn{2}{c}{Level 3} \\
\cmidrule(lr){2-3}
\cmidrule(lr){4-5}
\cmidrule(lr){6-7}
\cmidrule(lr){8-9}
& Pass@8 & Avg.@8
& Pass@8 & Avg.@8
& Pass@8 & Avg.@8
& Pass@8 & Avg.@8 \\
\midrule
Kimi K2.6                 & 84.0 & 69.10 & 94 & 77.13 & 89 & 75.50 & 54 & 40.25 \\
GLM-5.1              & 85.6 & 76.25 & 98 & 89.38 & 89 & 82.00 & 54 & 38.50 \\
DeepSeek-V4-Pro      & 83.2 & 54.90 & 97 & 68.88 & 86 & 56.38 & 50 & 24.00 \\
DeepSeek-V4-ProMax   & 84.8 & 60.05 & 97 & 72.88 & 92 & 62.63 & 46 & 29.25 \\
Claude Opus 4.7      & 87.2 & 77.30 & 99 & 91.75 & 92 & 81.88 & 54 & 39.25 \\
\midrule
Qwen3.5-9B            & 23.6 & 7.05 & 35 & 11.63 & 22 & 5.75 & 4 & 0.50 \\
Qwen3.6-27B           & 67.2 & 35.60 & 83 & 44.38 & 68 & 36.13 & 34 & 17.00 \\
\midrule
MusaCoder-9B-SFT      & 69.6 & 61.60 & 88 & 79.88 & 71 & 63.50 & 30 & 21.25 \\
MusaCoder-9B-RL       & 83.6 & 77.20 & 94 & 89.25 & 89 & 84.50 & 52 & 38.50 \\
MusaCoder-27B-SFT     & 84.8 & 79.40 & 95 & 89.00 & 92 & 87.75 & 50 & 43.50 \\
\textbf{MusaCoder-27B-RL} & \textbf{93.2} & \textbf{88.60} & \textbf{99} & \textbf{95.75} & \textbf{98} & \textbf{92.88} & \textbf{72} & \textbf{65.75} \\
\bottomrule
\end{tabular}
}

\vspace{0.8em}

\resizebox{\textwidth}{!}{
\begin{tabular}{lcccccccc}
\toprule
\multirow{2}{*}{Model}
& \multicolumn{2}{c}{Overall}
& \multicolumn{2}{c}{Level 1}
& \multicolumn{2}{c}{Level 2}
& \multicolumn{2}{c}{Level 3} \\
\cmidrule(lr){2-3}
\cmidrule(lr){4-5}
\cmidrule(lr){6-7}
\cmidrule(lr){8-9}
& \shortstack{vs.\\Eager} & \shortstack{vs.\\Compile}
& \shortstack{vs.\\Eager} & \shortstack{vs.\\Compile}
& \shortstack{vs.\\Eager} & \shortstack{vs.\\Compile}
& \shortstack{vs.\\Eager} & \shortstack{vs.\\Compile} \\
\midrule
Kimi K2.6                 & 3.3 & 1.4 & 7.5 & 3.0 & 0.6 & 0.5 & 0.2 & 0.0 \\
GLM-5.1              & 7.4 & 3.9 & 14.3 & 6.0 & 3.8 & 3.6 & 1.0 & 0.0 \\
DeepSeek-V4-Pro      & 5.2 & 3.0 & 10.3 & 4.8 & 2.8 & 2.6 & 0.0 & 0.0 \\
DeepSeek-V4-ProMax   & 5.7 & 3.0 & 11.6 & 5.0 & 2.3 & 2.3 & 0.5 & 0.3 \\
Claude Opus 4.7      & 11.8 & 7.5 & 21.0 & 10.5 & 6.9 & 6.8 & 3.3 & 2.8 \\
\midrule
Qwen3.5-9B            & 0.9 & 0.5 & 2.0 & 1.1 & 0.3 & 0.1 & 0.0 & 0.0 \\
Qwen3.6-27B           & 3.4 & 1.6 & 8.3 & 3.6 & 0.3 & 0.3 & 0.0 & 0.0 \\
\midrule
MusaCoder-9B-SFT      & 1.6  & 1.2  & 3.0  & 2.1  & 1.0  & 0.8  & 0.0  & 0.0 \\
MusaCoder-9B-RL       & 5.4  & 3.2  & 9.8  & 5.0  & 3.3  & 2.9  & 0.7  & 0.4 \\
MusaCoder-27B-SFT     & 6.3  & 4.1  & 12.1 & 7.6  & 3.0  & 2.2  & 1.3  & 1.0 \\
\textbf{MusaCoder-27B-RL} & \textbf{15.0} & \textbf{9.2} & \textbf{25.7} & \textbf{12.8} & \textbf{10.2} & \textbf{8.7} & \textbf{3.2} & \textbf{3.0} \\
\bottomrule
\end{tabular}
}
\end{table}

\subsection{Main Results on KernelBench}
\label{sec:main_results}

Table~\ref{tab:kernelbench_main_results} reports the main KernelBench results under the strict MooreEval protocol. In addition to frontier code models, we also include the two base checkpoints, Qwen3.5-9B and Qwen3.6-27B, to quantify the improvement brought by MusaCoder's training pipeline. All models are evaluated with the same prompts, sampling configuration, verifier, and hardware environment.

The base models show limited native kernel generation ability under strict executable verification. Qwen3.5-9B obtains only $23.6\%$ Overall Pass@8 and $7.05\%$ Avg.@8, while Qwen3.6-27B improves to $67.2\%$ Pass@8 and $35.60\%$ Avg.@8 but still falls behind strong general-purpose code models. After supervised alignment, MusaCoder-9B-SFT increases Overall Pass@8 to $69.6\%$ and Avg.@8 to $61.60\%$, while MusaCoder-27B-SFT reaches $84.8\%$ and $79.40\%$. These results indicate that the proposed SFT/RFT data pipeline substantially improves both task-format following and single-sample correctness stability.

Execution-feedback RL further improves both model scales. MusaCoder-9B-RL raises Overall Pass@8 from $69.6\%$ to $83.6\%$ and Avg.@8 from $61.60\%$ to $77.20\%$, approaching Claude Opus 4.7 in correctness under the same strict verifier. For the larger model, MusaCoder-27B-RL achieves the best overall correctness, reaching $93.2\%$ Pass@8 and $88.60\%$ Avg.@8. Compared with Claude Opus 4.7, this corresponds to absolute gains of $6.0$ points in Pass@8 and $11.30$ points in Avg.@8. The advantage is more pronounced on Level 3, where MusaCoder-27B-RL improves Pass@8 from Claude's $54\%$ to $72\%$ and Avg.@8 from $39.25\%$ to $65.75\%$, suggesting stronger robustness on complex shape reasoning, indexing, and multi-operator workloads.

Faster Rate is more stringent than correctness because a candidate is counted only when it is correct, legal, and achieves more than $1.1\times$ speedup over the corresponding PyTorch baseline. Under this criterion, the base models rarely produce accelerated kernels: Qwen3.5-9B achieves only $0.9\%$ Faster Rate vs. Eager and $0.5\%$ vs. Compile, while Qwen3.6-27B reaches $3.4\%$ and $1.6\%$. MusaCoder-9B-RL already slightly surpasses Claude Opus 4.7 on performance, achieving $12.6\%$ vs. Eager and $7.9\%$ vs. Compile, compared with Claude's $11.8\%$ and $7.5\%$.

MusaCoder-27B-RL shows the strongest performance gains, achieving $15.0\%$ Faster Rate vs. Eager and $9.2\%$ vs. Compile. These are absolute improvements of $3.2$ and $1.7$ points over Claude Opus 4.7. Overall, these results show that MusaCoder's execution-feedback training improves not only functional correctness but also the probability of generating native kernels with measurable runtime benefits.


\subsection{Ablation and RL Training Analysis}
\label{sec:ablation_rl_analysis}

To systematically deconstruct the contributions of individual training components within MusaCoder, we conduct rigorous ablation studies and training dynamics analyses across the supervised phase, the reinforcement learning phase, and the RL stabilization mechanisms. Diverging from the main experiments—which merely report the terminal KernelBench benchmarks—this section explicitly decouples first-turn generation, multi-turn repair, reward dynamics, and training stability to illuminate the precise origins of the model's performance gains.

\subsubsection{Evaluation Metrics for Multi-turn RL}
\label{sec:multiturn_metrics}

To clearly delineate the model's zero-shot generation capability from its multi-turn feedback-driven rectification proficiency, we report both \textit{first-turn metrics} and \textit{best-turn metrics} concurrently during multi-turn evaluation. For the $i$-th validation sample, let $c_{i,t}$ denote the candidate code generated at turn $t$, and let its corresponding MooreEval score be $s_{i,t}$. We define the binary success indicator function as:

\begin{equation}
\mathrm{success}(c_{i,t}) = \mathbf{1}[s_{i,t} \ge 1]
\end{equation}

First-turn accuracy measures the model's capability to correctly synthesize a kernel in a single attempt, strictly under zero-feedback conditions:

\begin{equation}
\mathrm{Acc}_{\mathrm{first}}
=
\frac{1}{N}
\sum_{i=1}^{N}
\mathrm{success}(c_{i,1})
\end{equation}

This metric directly proxies the model's performance in standard single-turn deployment scenarios, thus acting as one of the most critical constraint metrics during multi-turn RL optimization.

Best-turn accuracy measures whether the model can ultimately rectify a failure within a maximum horizon of $T$ feedback turns:

\begin{equation}
\mathrm{Acc}_{\mathrm{best}}
=
\frac{1}{N}
\sum_{i=1}^{N}
\max_{1 \le t \le T}
\mathrm{success}(c_{i,t})
\end{equation}

If a specific sample fails on the first attempt but is successfully rectified in subsequent turns, it does not positively impact the first-turn accuracy but directly elevates the best-turn accuracy. Consequently, the best-turn accuracy primarily reflects the model's agentic proficiency in leveraging MooreEval feedback for continuous rectification.

Beyond binary correctness rates, we also report the first-turn score and best-turn score:

\begin{equation}
S_{\mathrm{first}}
=
\frac{1}{N}
\sum_{i=1}^{N}
s_{i,1},
\qquad
S_{\mathrm{best}}
=
\frac{1}{N}
\sum_{i=1}^{N}
\max_{1 \le t \le T} s_{i,t}
\end{equation}

The scalar score preserves granular telemetry—such as partial correctness shaping, structural infraction penalties, and empirical speedup bonuses—rendering it a higher-resolution metric than binary accuracy. Finally, we aggregate the \textit{turn-to-success} (i.e., the specific turn number at which a sample first passes verification) to empirically observe whether the policy exhibits a distributional bias toward immediate zero-shot success or an over-reliance on deferred feedback-driven rectifications.

\subsubsection{Component Ablation}
\label{sec:component_ablation}

Table~\ref{tab:ablation} reports the ablation results for the main training components. Overall, each component contributes to the final performance, but their effects are reflected in different aspects: RFT improves supervised task alignment, single-turn warmup provides a stronger initialization for multi-turn RL, PrimeEcho controls the multi-turn reward objective, BDR recovers learning signals from hard samples, and MirrorPop stabilizes off-policy updates.

\begin{table}[htbp]
\centering
\small
\caption{
Ablation study on the main training components. We report Overall Pass Rate and Faster Rate to measure the effect of each component on correctness and performance.
}
\label{tab:ablation}
\begin{tabular}{lcccc}
\toprule
\multirow{2}{*}{Setting}
& \multicolumn{2}{c}{Overall Pass Rate}
& \multicolumn{2}{c}{Overall Faster Rate} \\
\cmidrule(lr){2-3}
\cmidrule(lr){4-5}
& Pass@8 & Avg.@8
& vs. Eager & vs. Compile \\
\midrule
MusaCoder-SFT                 & 84.8 & 79.40 & 6.3 & 4.1 \\
\quad w/o RFT                 & 82.6 & 75.10 & 5.8 & 3.8 \\
\midrule
MusaCoder-RL                  & 93.2 & 88.60 & 15.0 & 9.2 \\
\quad w/o Single-turn Warmup  & 90.8 & 84.25 & 14.2 & 8.6 \\
\quad w/o PrimeEcho           & 88.4 & 83.50 & 13.9 & 8.4 \\
\quad w/o Buffered Dynamic Retry & 88.6 & 83.20 & 13.8 & 8.2 \\
\quad w/o MirrorPop           & 86.0 & 80.75 & 13.1 & 7.8 \\
\bottomrule
\end{tabular}
\end{table}

First, RFT improves the supervised checkpoint before RL. Removing RFT reduces MusaCoder-SFT's Overall Pass@8 from $84.8\%$ to $82.6\%$ and Avg.@8 from $79.40\%$ to $75.10\%$, corresponding to drops of $2.2$ and $4.30$ percentage points. The performance metrics also decline: Faster Rate vs. Eager decreases from $6.3\%$ to $5.8\%$, and vs. Compile decreases from $4.1\%$ to $3.8\%$. This indicates that MooreEval-filtered RFT improves not only correctness but also the stability of single-sample generation. Since our RFT uses diversity-preserving filtering rather than selecting only the fastest candidate, it strengthens task alignment while maintaining implementation diversity for downstream RL exploration.

Second, single-turn warmup is important for initializing the multi-turn RL stage. Without single-turn warmup, MusaCoder-RL drops from $93.2\%$ to $90.8\%$ in Overall Pass@8 and from $88.60\%$ to $84.25\%$ in Avg.@8. The Faster Rate also decreases substantially, from $15.0\%$ to $14.2\%$ against PyTorch eager execution and from $9.2\%$ to $8.6\%$ against \texttt{torch.compile}. These results suggest that directly entering multi-turn RL from the supervised checkpoint weakens the training signal, because early feedback turns are more likely to be spent on basic formatting, compilation, or interface errors. Single-turn warmup first aligns the policy with executable verification, giving the subsequent multi-turn stage a stronger starting point.

PrimeEcho also brings clear gains. Removing PrimeEcho reduces Overall Pass@8 from $93.2\%$ to $88.4\%$ and Avg.@8 from $88.60\%$ to $83.50\%$. Faster Rate decreases from $15.0\%/9.2\%$ to $13.9\%/8.4\%$ against PyTorch eager execution and \texttt{torch.compile}, respectively. This confirms the importance of first-turn-anchored multi-turn reward: if the trajectory reward over-emphasizes final-turn or best-turn success, the model may rely more heavily on later feedback instead of improving first-turn generation quality. PrimeEcho preserves the benefit of multi-turn exploration while maintaining pressure on the first response.

Buffered Dynamic Retry also contributes noticeably to the final model. Removing BDR reduces Overall Pass@8 from $93.2\%$ to $88.6\%$ and Avg.@8 from $88.60\%$ to $83.20\%$, corresponding to drops of $4.6$ and $5.40$ percentage points. The Faster Rate decreases from $15.0\%/9.2\%$ to $13.9\%/8.4\%$. Although BDR is designed to target all-failed hard samples rather than every training example, its ablation still leads to a sizable degradation, suggesting that recovering learning signals from long-tail difficult prompts is important for both correctness and performance.

Finally, MirrorPop has the largest impact among the stabilization components. Without MirrorPop, Overall Pass@8 drops from $93.2\%$ to $86.0\%$, while Avg.@8 decreases from $88.60\%$ to $80.75\%$, corresponding to drops of $7.2$ and $7.85$ percentage points. Faster Rate also falls from $15.0\%/9.2\%$ to $13.1\%/7.8\%$. This demonstrates that off-policy sequence masking is not merely a background stability trick; it directly affects final correctness and speedup. By measuring the absolute magnitude of log-ratio deviations, MirrorPop avoids the cancellation problem of signed log-ratios and filters out high-mismatch sequences that would otherwise introduce unstable policy updates.

\subsubsection{Effect of Single-turn RL Warmup}
\label{sec:analysis_single_turn_rl}

Single-turn RL constitutes the critical transitional phase bridging supervised learning and execution-feedback optimization in MusaCoder. Both the main results and the ablation studies unequivocally demonstrate that this phase not only elevates terminal metrics but profoundly ameliorates the initialization quality for subsequent multi-turn training. Taking the 27B model as a primary example, the Overall Pass@8 and Avg.@8 during the SFT phase stand at $84.8\%$ and $79.40\%$, respectively. Following RL, these metrics surge to $93.2\%$ and $88.60\%$. In terms of performance, the Faster Rate vs. Eager escalates from $6.3\%$ to $15.0\%$, and vs. Compile from $4.1\%$ to $9.2\%$. This substantiates that execution-driven rewards do not merely amplify the probability of authoring functionally correct kernels; they dramatically inflate the proportion of kernels that achieve genuine hardware acceleration.

Unlike SFT/RFT—which intrinsically relies on offline static samples—every candidate implementation synthesized during single-turn RL is subjected to live, iterative processing by MooreEval. Each response is compiled, executed, validated for mathematical correctness, and rigidly inspected for forbidden PyTorch/\texttt{aten::*} fallbacks. Only after satisfying strict functional and architectural compliance checks is the candidate benchmarked for relative speedup against the PyTorch baseline. Consequently, the optimization gradient in this phase perfectly mirrors the terminal evaluation objective: high rewards are exclusively minted by native CUDA/MUSA kernels that are compilable, numerically sound, architecturally legal, and empirically performant.

Analyzing the empirical training trajectory reveals three primary dimensions of improvement induced by single-turn RL. First, it systematically eradicates low-level formatting anomalies—such as code extraction failures, missing \texttt{ModelNew} structural declarations, and return type mismatches—enforcing an output distribution that stably adheres to the evaluation interface. Second, MooreEval's draconian penalization of forbidden fallbacks actively suppresses the model's inclination to invoke high-level PyTorch/\texttt{aten::*} operators as a shortcut to bypassing the core task, aggressively steering the policy toward explicit custom kernel synthesis. Finally, among samples that are already functionally correct, the continuous performance reward further stratifies implementation quality. It computationally incentivizes the model to favor optimal thread mapping, contiguous memory access patterns, and efficient reduction heuristics, thereby directly elevating the acceleration ratio among verified kernels.

The ablation results further codify the single-turn warmup as an indispensable foundational prerequisite for subsequent multi-turn RL. Ablating the single-turn warmup regresses MusaCoder-RL's Overall Pass@8 from $93.2\%$ down to $90.8\%$, and its Avg.@8 from $88.60\%$ down to $84.25\%$; the Faster Rate vs. Compile concurrently drops from $9.2\%$ to $8.6\%$. This underscores a severe pathology: thrusting a supervised model directly into multi-turn training forces the system to squander valuable feedback epochs on rectifying primitive formatting violations and trivial compilation aborts. By contrast, interposing the single-turn RL phase ensures that the subsequent multi-turn regime can exclusively focalize on high-value rectification modes, such as numerical convergence debugging, boundary condition handling, complex tensor indexing, and algorithmic performance profiling.

\subsubsection{Training Dynamics of Multi-turn RL}
\label{sec:multiturn_training_dynamics_analysis}

\begin{figure}[htbp]
    \centering
    \begin{subfigure}{0.48\linewidth}
        \centering
        \includegraphics[width=\linewidth]{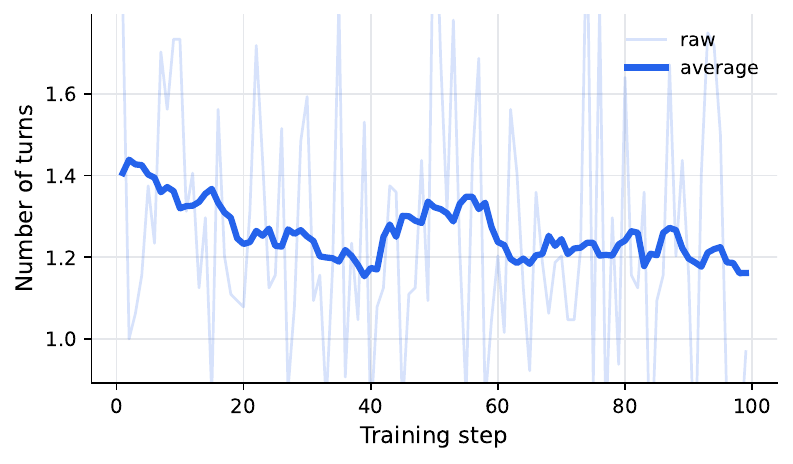}
        \caption{Number of turns}
        \label{fig:multiturn_num_turns}
    \end{subfigure}
    \hfill
    \begin{subfigure}{0.48\linewidth}
        \centering
        \includegraphics[width=\linewidth]{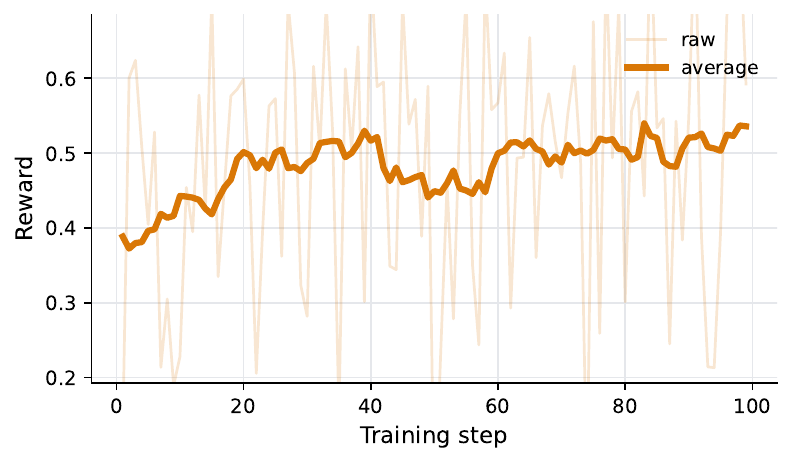}
        \caption{Reward}
        \label{fig:multiturn_reward}
    \end{subfigure}
    \caption{Training dynamics of multi-turn reinforcement learning. Raw curves are shown in light color and smoothed curves are used for readability.}
    \label{fig:multiturn_training_dynamics}
\end{figure}

Figure~\ref{fig:multiturn_training_dynamics} illustrates the training dynamics throughout the multi-turn RL phase. As training progresses, the average number of interaction turns per sample steadily declines, while the aggregated multi-turn reward exhibits a continuous upward trajectory. This indicates that the policy model not only converges toward higher terminal scores across multi-turn interactions but concurrently curtails its over-reliance on deferred feedback epochs, exhibiting a pronounced inductive bias toward synthesizing viable candidate solutions in earlier turns. Given the inherently sparse nature of rewards in kernel generation tasks and the profound variance in task difficulty, the raw telemetry exhibits anticipated structural volatility; nonetheless, the smoothed trajectories delineate a definitive and stable optimization trend.

\begin{figure}[htbp]
\centering
\begin{minipage}{0.32\textwidth}
    \includegraphics[width=\linewidth]{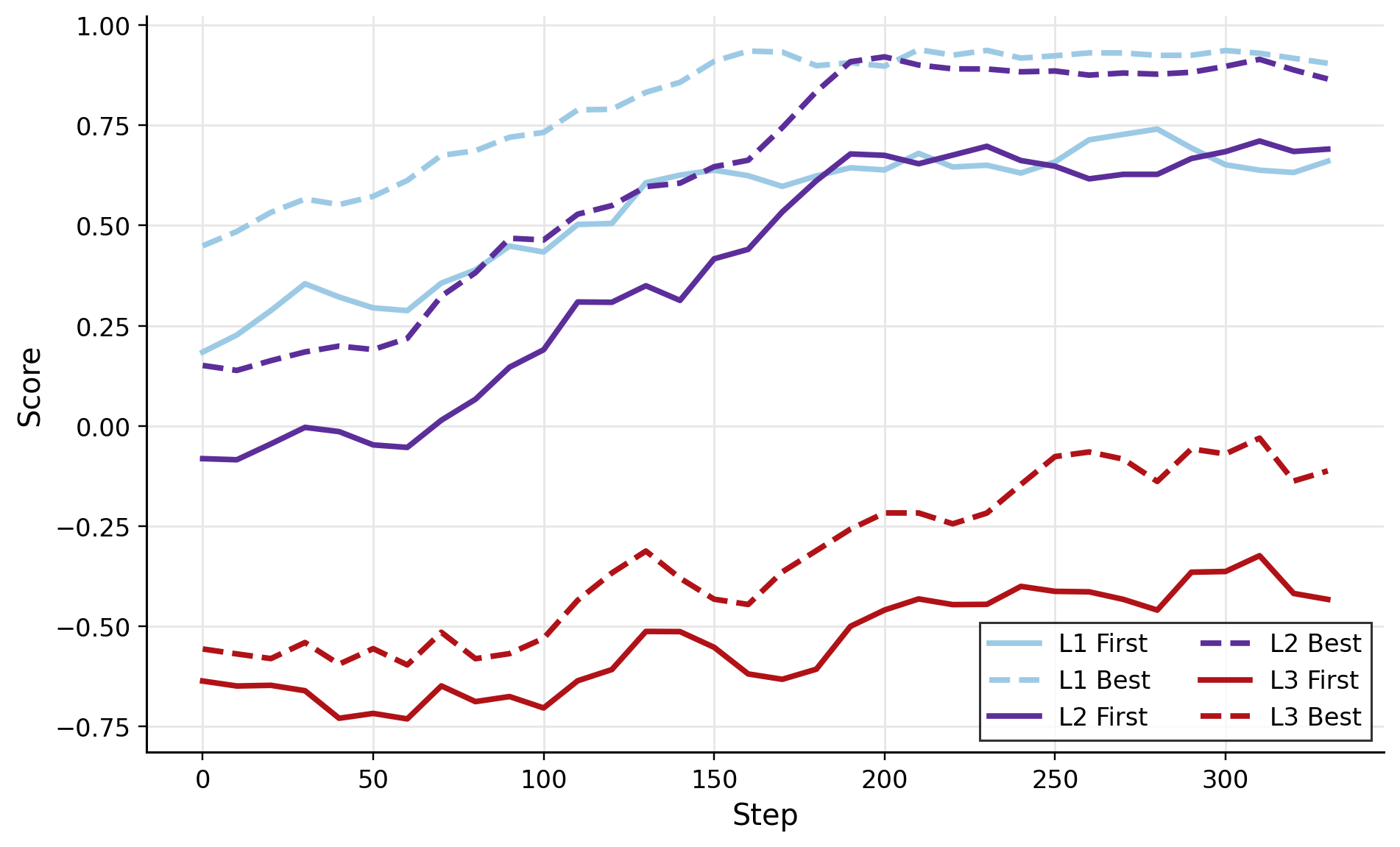}
    \centering Score
\end{minipage}%
\hfill
\begin{minipage}{0.32\textwidth}
    \includegraphics[width=\linewidth]{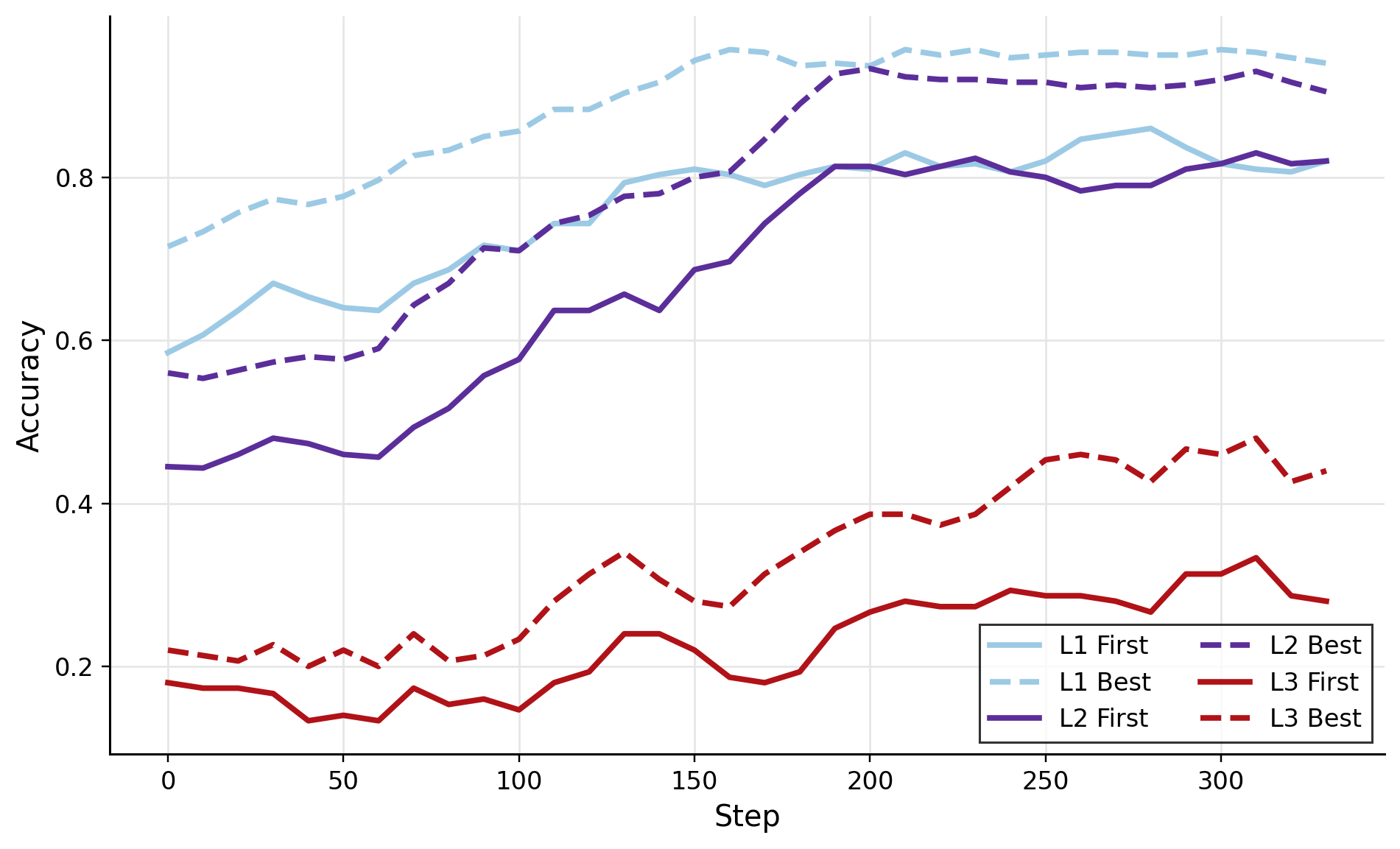}
    \centering Accuracy
\end{minipage}%
\hfill
\begin{minipage}{0.32\textwidth}
    \includegraphics[width=\linewidth]{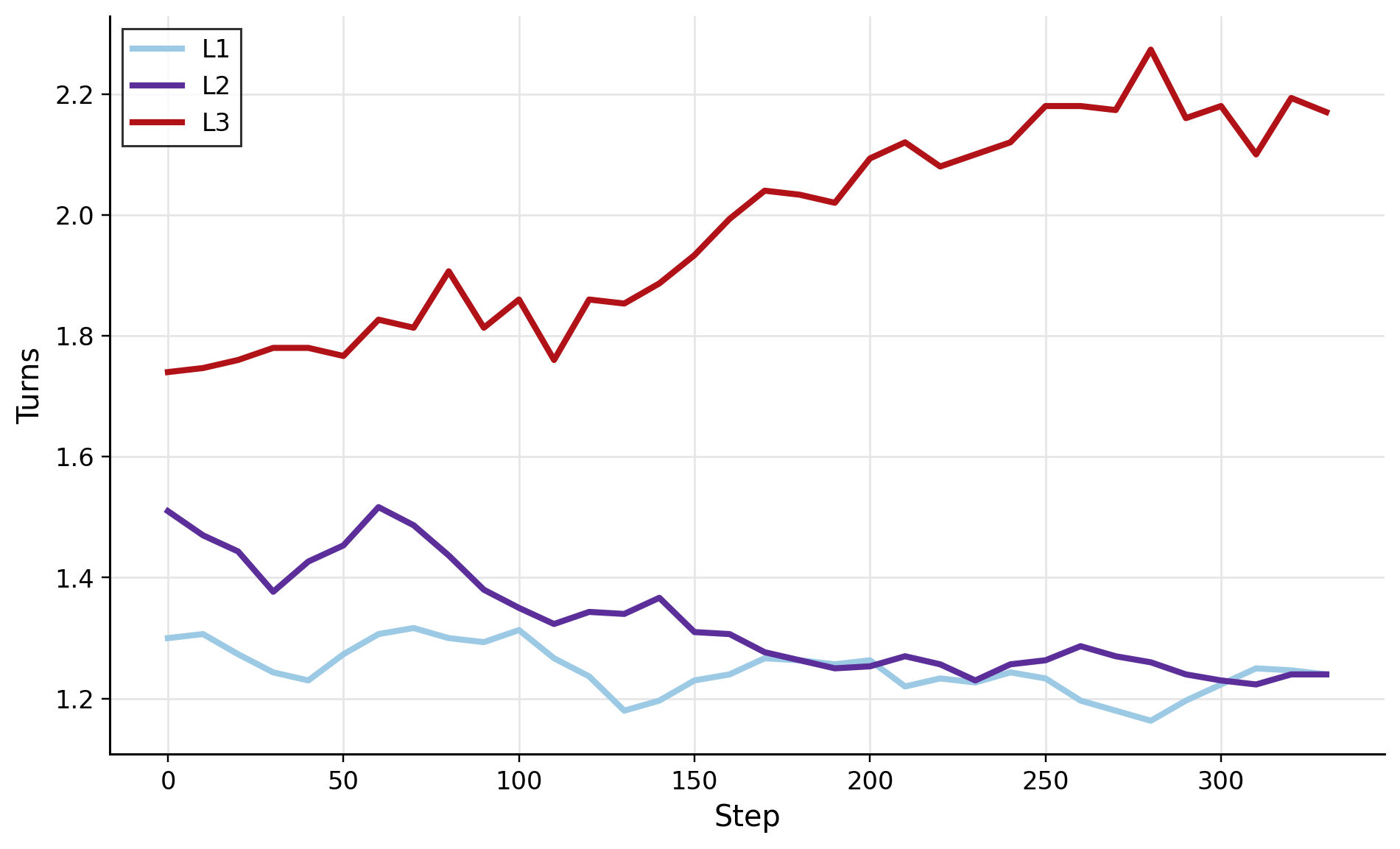}
    \centering Turns
\end{minipage}
\caption{Multi-turn evaluation metrics across KernelBench levels.}
\label{fig:kernelbench_all}
\end{figure}

Figure~\ref{fig:kernelbench_all} further decomposes the multi-turn evaluation results across distinct difficulty tiers. The Score metric proxies the holistic quality of the model's output, encapsulating partial correctness, architectural legality, and empirical performance parameters. The Accuracy metric is concurrently reported for both first-turn and best-turn outcomes, explicitly isolating zero-shot generation proficiency from multi-turn rectification capability. The Avg. Eval Turns metric quantifies the mean number of evaluation iterations required for the model to converge upon a functionally correct solution.

Analyzing the disparate difficulty levels reveals that Level 1 tasks predominantly achieve high correctness rates within the very first turn, accompanied by a low average evaluation turn count. This suggests that the model has robustly internalized the heuristics for foundational operators and trivial composite patterns. Conversely, Level 2 and Level 3 tasks exhibit noticeably lower first-turn accuracies; however, their best-turn accuracies experience a marked surge following multi-turn feedback. This confirms that MooreEval feedback fundamentally empowers the model to debug and rectify complex shape, dtype, indexing, and numerical convergence errors. Concurrently, the elevated average evaluation turn count for Level 3 tasks underscores that intricate, model-level kernels inherently demand protracted feedback horizons to achieve full rectification.

These empirical dynamics vindicate the architectural rationale underpinning PrimeEcho. While multi-turn feedback intrinsically elevates the terminal rectification success rate, optimizing the policy exclusively for best-turn success risks cultivating an over-reliance on subsequent feedback, inevitably eroding first-turn zero-shot generation quality. PrimeEcho mitigates this by rigidly anchoring the trajectory reward to the initial turn's score, relegating multi-turn feedback to an auxiliary exploratory signal designed to augment—rather than eclipse—the primary objective of first-turn generation. The ablation studies unequivocally validate this: omitting PrimeEcho precipitates a regression in Overall Avg.@8 from $88.60\%$ to $83.50\%$, and a drop in Faster Rate vs. Compile from $9.2\%$ to $8.4\%$. This demonstrates that the first-turn-anchored reward is absolutely paramount for sustaining both strict generation quality and empirical performance dividends.

\subsubsection{Effect of Buffered Dynamic Retry}
\label{sec:ablation_bdr}

Buffered Dynamic Retry (BDR) is engineered primarily to alleviate the pathology of \textit{all-failed groups} endemic to long-tail hard samples. In the main ablation study, removing BDR precipitates a regression in MusaCoder-RL's Overall Pass@8 from $93.2\%$ to $88.6\%$, and its Avg.@8 from $88.60\%$ to $83.20\%$. Commensurately, Faster Rates vs. Eager and vs. Compile decline from $15.0\%$ and $9.2\%$ to $13.8\%$ and $8.2\%$, respectively. Although this degradation is less severe than ablating the single-turn warmup or MirrorPop, BDR's impact is highly concentrated on the long tail of complex tasks: it surgically transfigures terminal all-failed groups—which natively lack any positive baseline—into feedback-conditioned rectification tasks, thereby resurrecting critical gradient signals.

\begin{figure}[htbp]
    \centering
    \includegraphics[width=0.55\textwidth]{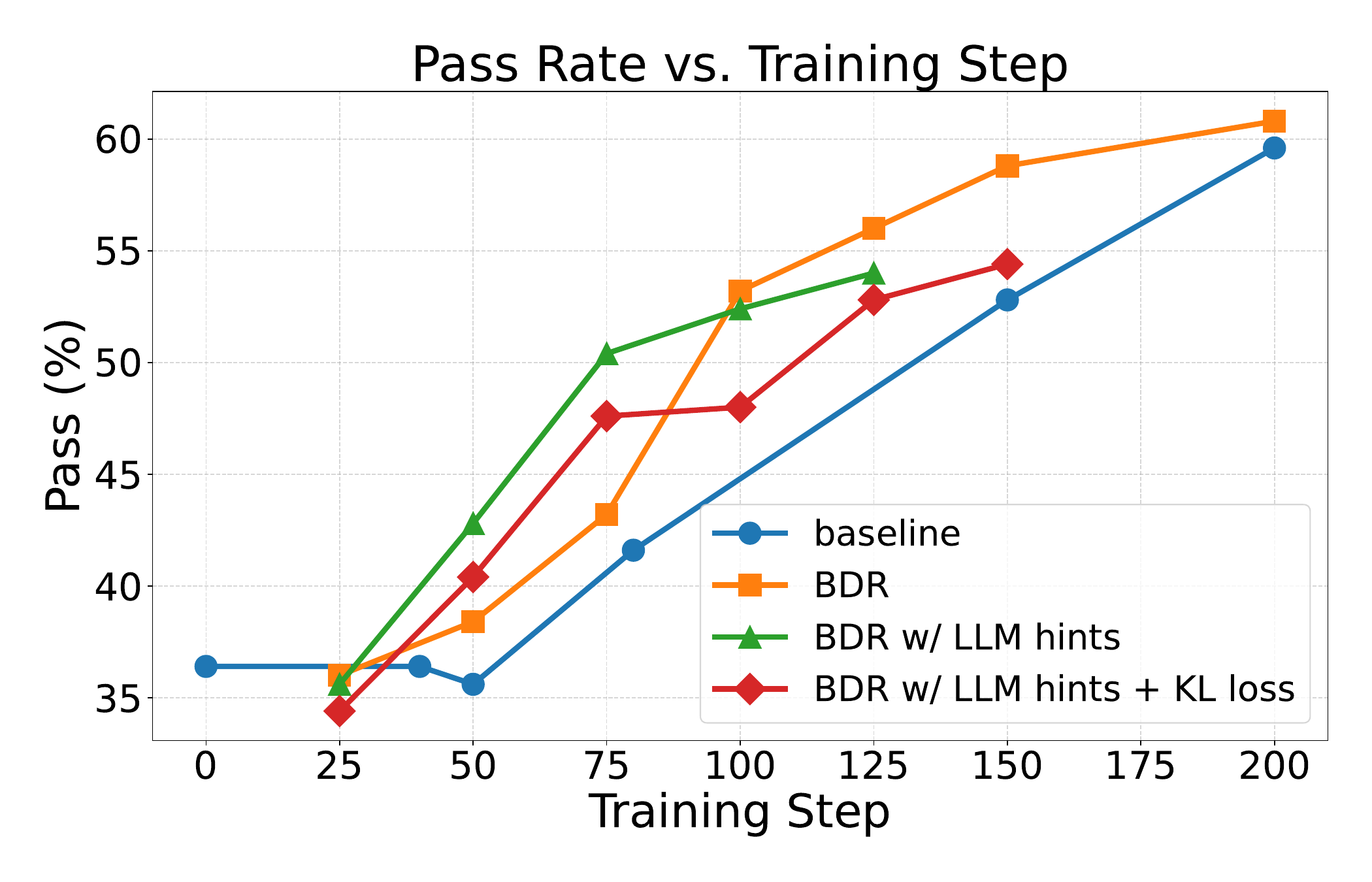}
    \caption{Ablation of Buffered Dynamic Retry under different feedback settings.}
    \label{fig:bdr_compare}
\end{figure}

Figure~\ref{fig:bdr_compare} juxtaposes the training trajectories across disparate BDR configurations against the baseline using Qwen3-8B. We empirically evaluated three paradigms: utilizing native MooreEval execution feedback directly, employing structured feedback synthesized by an external frontier model, and supplementing the external feedback with KL regularization. Empirical evidence indicates that while external frontier feedback accelerates convergence during early training phases, such prompts never manifested during the SFT phase. Consequently, they introduce severe out-of-distribution (OOD) contexts, inducing destabilizing fluctuations in the policy's entropy. Imposing KL regularization fails to entirely suppress these anomalies and inadvertently retards convergence. Therefore, the optimal terminal configuration exclusively deploys the native sandbox feedback from MooreEval as the retry directive.

Table~\ref{tab:bdr_results} details the outcomes of a short-horizon BDR training regimen initialized from the single-turn RL checkpoint. For Qwen3-8B, BDR elevates the Pass@8 from $59.6\%$ to $62.4\%$, successfully resurrecting approximately $16\%$ of initially completely failed samples. For Qwen3-9B, BDR advances the Pass@8 from $73.2\%$ to $74.4\%$, yielding a recovery rate of approximately $28\%$. This underscores that BDR does not merely naively augment the dataset volume; it selectively resurrects actionable learning signals exclusively on instances where the current policy genuinely fails. A higher recovery rate directly proxies the policy's enhanced capacity to autonomously assimilate execution feedback for self-correction.

\begin{table}[htbp]
\centering
\small
\caption{Effect of Buffered Dynamic Retry. The results are obtained by continuing training from the best single-turn RL checkpoints.}
\label{tab:bdr_results}
\begin{tabular}{llcc}
\toprule
\textbf{Model} & \textbf{Method} & \textbf{Pass@8} & \textbf{Recovery Rate} \\
\midrule
\multirow{2}{*}{Qwen3-8B}
& Baseline & 59.6\% & -- \\
& BDR      & 62.4\% & $\sim$16\% \\
\midrule
\multirow{2}{*}{Qwen3.5-9B}
& Baseline & 73.2\% & -- \\
& BDR      & 74.4\% & $\sim$28\% \\
\bottomrule
\end{tabular}
\end{table}

It is crucial to note that BDR is fundamentally ill-suited as a high-frequency global training regimen throughout the entire RL pipeline. Because it structurally transfigures a subset of zero-shot generation tasks into feedback-conditioned repair tasks, excessive or premature integration risks task interference, potentially eroding the primary objective of first-turn synthesis. Consequently, we strategically deploy BDR strictly as a late-stage stabilization technique. Once the model has attained a robust baseline generation proficiency, feedback-augmented samples are introduced at a marginal sampling ratio, serving specifically to conquer the long-tail boundary of hard tasks.

\subsubsection{Effect of MirrorPop}
\label{sec:ablation_mirrorpop}

The ablation results show that MirrorPop plays an important role in both final performance and RL training stability. 
As reported in Table \ref{tab:ablation}, removing MirrorPop reduces the Overall Pass@8 from $93.2\%$ to $86.0\%$ and the Avg.@8 from $88.60\%$ to $80.75\%$, corresponding to absolute drops of $7.2$ and $7.85$ percentage points, respectively. 
The performance-oriented metrics also degrade: the Faster Rate vs. Eager decreases from $15.0\%$ to $13.1\%$, while the Faster Rate vs. Compile drops from $9.2\%$ to $7.8\%$. 
These results indicate that MirrorPop is not merely a training-curve smoothing heuristic; it directly contributes to final kernel correctness and empirical speedup.

\begin{figure}[htb]
    \centering

    \begin{subfigure}{0.32\linewidth}
        \centering
        \includegraphics[width=\linewidth]{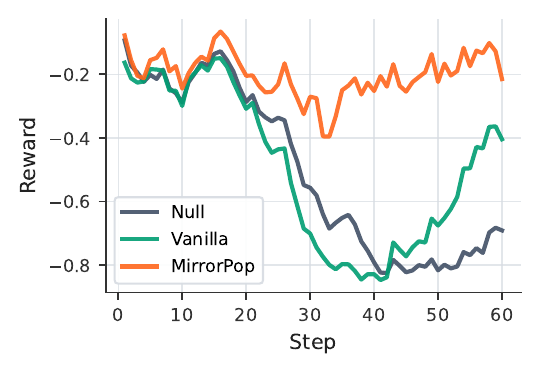}
        \caption{Training reward}
    \end{subfigure}
    \hfill
    \begin{subfigure}{0.32\linewidth}
        \centering
        \includegraphics[width=\linewidth]{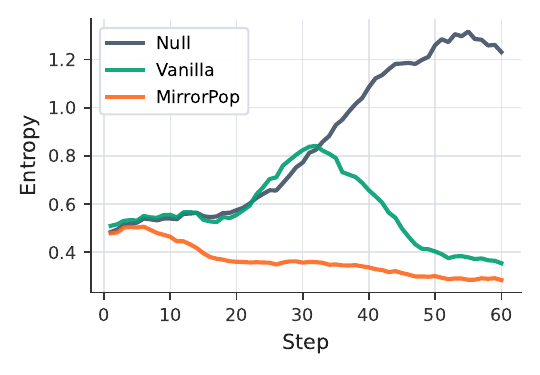}
        \caption{Entropy}
    \end{subfigure}
    \hfill
    \begin{subfigure}{0.32\linewidth}
        \centering
        \includegraphics[width=\linewidth]{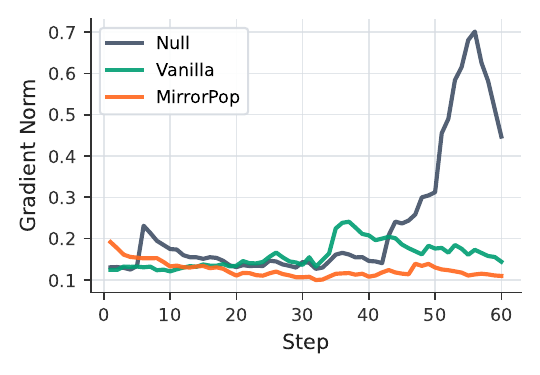}
        \caption{Gradient norm}
    \end{subfigure}

    \vspace{0.5em}

    \begin{subfigure}{0.32\linewidth}
        \centering
        \includegraphics[width=\linewidth]{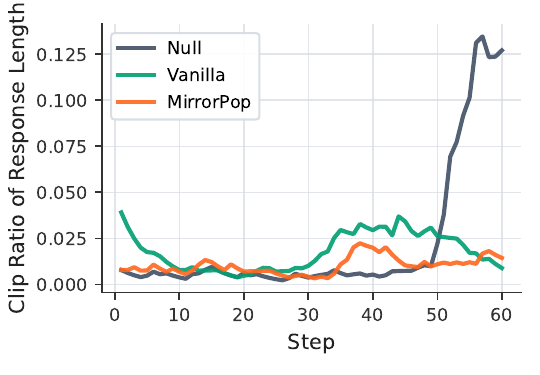}
        \caption{Response length clip ratio}
    \end{subfigure}
    \hfill
    \begin{subfigure}{0.32\linewidth}
        \centering
        \includegraphics[width=\linewidth]{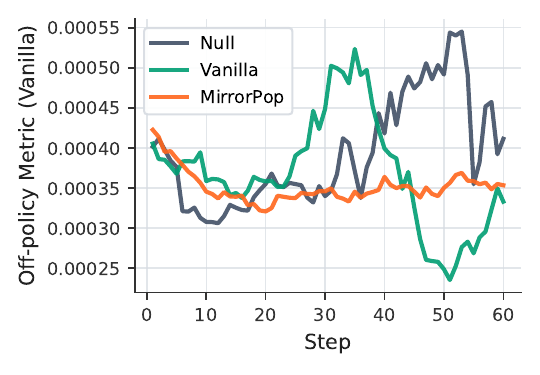}
        \caption{Off-policy metric in Vanilla form}
    \end{subfigure}
    \hfill
    \begin{subfigure}{0.32\linewidth}
        \centering
        \includegraphics[width=\linewidth]{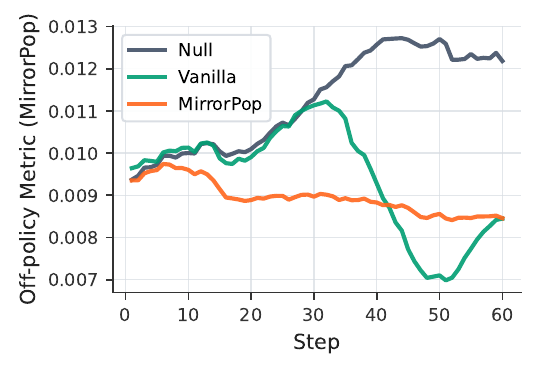}
        \caption{Off-policy metric in MirrorPop form}
    \end{subfigure}

    \caption{
    Training dynamics under different off-policy sequence masking strategies: Null, Vanilla, and MirrorPop. 
    All runs are continued from the same checkpoint selected before this ablation. 
    The last two panels report passive sequence-level off-policy diagnostics computed in the Vanilla signed-log-ratio form and the MirrorPop absolute-log-ratio form, respectively.
    }
    \label{fig:mirrorpop_ablation}
\end{figure}

Figure~\ref{fig:mirrorpop_ablation} further compares three sequence-level off-policy masking configurations: \textbf{Null}, which disables sequence-level filtering and updates the policy using all rollouts; \textbf{Vanilla}, which performs masking based on conventional signed log-ratio aggregation; and \textbf{MirrorPop}, which masks sequences according to absolute log-ratio deviations. 
All three runs are continued from the same strong checkpoint, selected according to the best validation performance before this ablation.

The training reward trajectories reveal a clear difference among the three settings. 
Both Null and Vanilla suffer from an initial reward drop after continued training begins, indicating that the policy can be destabilized when high-mismatch trajectories are not reliably filtered. 
Vanilla later recovers part of the lost reward, suggesting that signed log-ratio masking still removes some harmful off-policy samples. 
However, its recovered reward remains below MirrorPop, while Null shows little or only marginal recovery after the drop. 
Overall, Null and Vanilla both degrade to a substantially lower reward level than MirrorPop, with the main difference being that Vanilla partially rebounds whereas Null largely fails to recover. 
In contrast, MirrorPop maintains a consistently higher and more stable reward trajectory, indicating more reliable identification and suppression of high-risk off-policy samples.

The entropy, gradient norm, and response length clipping curves (Figure~\ref{fig:mirrorpop_ablation}, b-d) provide consistent evidence. 
Under Null, the model shows pronounced fluctuations in gradient norm and entropy, together with a higher response length clipping ratio, indicating unstable optimization and a tendency toward pathological long responses. 
Vanilla partially alleviates these issues, but still exhibits noticeable oscillations. 
MirrorPop produces smoother gradient dynamics, more controlled entropy, and a lower response length clipping ratio, suggesting that its sequence-level filtering improves both optimization stability and output-structure control.

To further analyze off-policy behavior, we report two passive sequence-level diagnostics corresponding to the Vanilla and MirrorPop formulations (Figure~\ref{fig:mirrorpop_ablation}, e-f). 
The Vanilla-form diagnostic computes the signed mean log-ratio:
$
\frac{1}{L_i}
\sum_{t=1}^{L_i}
\log \rho_{i,t}.
$
This statistic measures the signed average shift of a response, but positive and negative token-level deviations can cancel each other. 
The MirrorPop-form diagnostic instead computes the mean absolute log-ratio:
$
\frac{1}{L_i}
\sum_{t=1}^{L_i}
\left|\log \rho_{i,t}\right|,
$
which measures the magnitude of sequence-level policy mismatch regardless of direction.

The two diagnostics reveal different behaviors. 
Under the Vanilla-form metric, both Null and Vanilla strategies first increase and then decrease, without showing a clear trend of off-policy control. 
However, under the MirrorPop-form metric, Null strategy continues to increase over training, while Vanilla strategy still shows a rise-and-fall pattern. 
This discrepancy suggests that signed log-ratio aggregation can underestimate the true severity of off-policy drift due to cancellation, and therefore may provide a misleading picture of whether high-mismatch trajectories are being effectively controlled. 
In contrast, MirrorPop strategy remains stable and gradually decreases under both diagnostics, indicating more consistent control over sequence-level policy drift.

Overall, these results support the design rationale of MirrorPop. 
Rather than measuring only the directional bias of a response relative to the rollout policy, MirrorPop estimates the magnitude of sequence-level mismatch. 
This allows it to more reliably mask severely off-policy samples, thereby reducing unstable policy updates and improving the stability of RL training.

\subsection{MUSA KernelBench}
\label{sec:musa_kernelbench_results}

To further evaluate MusaCoder's cross-backend kernel generation capability, we conduct experiments on a MUSA-ported variant of KernelBench. This benchmark is adapted from the original CUDA version while preserving the same task structure and difficulty splits. We replace CUDA-specific task descriptions, backend constraints, and device-placement operations with their MUSA counterparts, e.g., mapping \texttt{.cuda()} to \texttt{.musa()}. For prompt exemplars that contain CUDA kernels, we use Musify to convert CUDA APIs, launch interfaces, and compilation commands into MUSA-compatible implementations. Converted exemplars are further checked through compilation and correctness validation, and only verified examples are retained to ensure consistency with the MUSA runtime.

Since \texttt{torch.compile} has not yet been fully benchmarked on the current MUSA platform, we report Faster Rate only against PyTorch eager execution. A candidate is counted as faster only if it passes correctness and legality checks and achieves a speedup greater than $1.1\times$, filtering out marginal timing noise. We compare MusaCoder with strong general code models, including DeepSeek-V4-Pro and GLM-5.1, to assess whether a model trained with CUDA/MUSA kernel data and execution feedback can generate native MUSA kernels directly from PyTorch references.

\begin{table*}[t]
\centering
\scriptsize
\setlength{\tabcolsep}{3.2pt}
\caption{
MUSA KernelBench evaluation results. We report correctness-oriented Pass Rate metrics and performance-oriented Faster Rate metrics across Overall, Level 1, Level 2, and Level 3. Pass@8 denotes whether at least one of 8 sampled candidates passes verification, while Avg.@8 measures the average correctness rate among 8 samples. Faster denotes Faster Rate against PyTorch eager execution; \texttt{torch.compile} was not evaluated on the MUSA platform. A candidate is counted as faster only when its speedup is larger than $1.1\times$.
}
\label{tab:musa_kernelbench_results}

\resizebox{\textwidth}{!}{
\begin{tabular}{lcccccccccccc}
\toprule
\multirow{2}{*}{Model}
& \multicolumn{3}{c}{Overall}
& \multicolumn{3}{c}{Level 1}
& \multicolumn{3}{c}{Level 2}
& \multicolumn{3}{c}{Level 3} \\
\cmidrule(lr){2-4}
\cmidrule(lr){5-7}
\cmidrule(lr){8-10}
\cmidrule(lr){11-13}
& Pass@8 & Avg.@8 & Faster
& Pass@8 & Avg.@8 & Faster
& Pass@8 & Avg.@8 & Faster
& Pass@8 & Avg.@8 & Faster \\
\midrule
DeepSeek-V4-Pro      & 92.0 & 56.9 & 5.7 & 96 & 63.9 & 6.6 & 96 & 61.5 & 7.5 & 76 & 33.8 & 0.0 \\
GLM-5.1              & 88.0 & 66.4 & 6.9 & 92 & 75.5 & 8.3 & 93 & 68.1 & 8.9 & 70 & 45.0 & 0.0 \\
\midrule
MusaCoder-9B-SFT      & 64.8 & 56.2 & 1.9 & 84 & 74.0 & 3.1 & 68 & 59.0 & 1.7 & 20 & 15.0 & 0.0 \\
MusaCoder-9B-RL       & 84.4 & 72.3 & 6.4 & 94 & 82.5 & 9.0 & 90 & 77.0 & 5.7 & 54 & 42.5 & 2.6 \\
MusaCoder-27B-SFT     & 79.6 & 63.5 & 5.2 & 92 & 80.1 & 7.6 & 84 & 64.5 & 4.9 & 46 & 27.9 & 1.0 \\
\textbf{MusaCoder-27B-RL}   & \textbf{92.4} & \textbf{81.7} & \textbf{12.5} & \textbf{98} & \textbf{92.7} & \textbf{19.6} & \textbf{96} & \textbf{84.2} & \textbf{10.0} & \textbf{74} & \textbf{54.1} & \textbf{3.3} \\
\bottomrule
\end{tabular}
}
\end{table*}



Table~\ref{tab:musa_kernelbench_results} shows that execution-feedback training consistently improves MusaCoder on the MUSA backend. For the 9B model, RL increases Overall Pass@8 from $64.8\%$ to $84.4\%$, Avg.@8 from $56.2\%$ to $72.3\%$, and Faster Rate from $1.9\%$ to $6.4\%$. For the 27B model, RL further improves Overall Pass@8 from $79.6\%$ to $92.4\%$, Avg.@8 from $63.5\%$ to $81.7\%$, and Faster Rate from $5.2\%$ to $12.5\%$. These gains indicate that RL does not merely improve functional correctness; it also helps the model discover implementation patterns that are more likely to accelerate on the MUSA backend.

Compared with general-purpose code models, MusaCoder-27B-RL achieves the strongest overall stability and performance. Although DeepSeek-V4-Pro already reaches a high Overall Pass@8 of $92.0\%$, its Avg.@8 is only $56.9\%$, suggesting that correct solutions appear less consistently across samples. MusaCoder-27B-RL achieves a similar Pass@8 of $92.4\%$ but substantially higher Avg.@8 of $81.7\%$, improving single-sample correctness stability by $24.8$ points. It also improves Faster Rate from DeepSeek's $5.7\%$ and GLM's $6.9\%$ to $12.5\%$. On Level 3, MusaCoder-27B-RL reaches $54.1\%$ Avg.@8 and $3.3\%$ Faster Rate, while both baselines obtain $0.0\%$ Faster Rate, showing that MusaCoder is more effective at generating usable and accelerated kernels for difficult MUSA workloads.

These results strengthen the central claim of this technical report: MusaCoder is not only a CUDA-oriented kernel generation model, but also a cross-backend native kernel synthesis system. The same data construction, verification, and execution-feedback training pipeline can transfer to the MUSA ecosystem and produce competitive native-kernel generation capability. This provides an important foundation for building an open MUSA kernel generation community, where LLMs can assist developers in writing, validating, repairing, and optimizing backend-specific GPU kernels.

\newpage

\bibliographystyle{plainnat}
\bibliography{musacode}

\newpage
\appendix

\section{Revised prompt for KernelBench}
\label{app:prompt_kernelbench}
\begin{tcblisting}{
    colback=gray!5, 
    colframe=gray!40, 
    title=Revised prompt for KernelBench,
    fonttitle=\bfseries,
    breakable,
    listing only,
    listing engine=listings,
    listing options={
        language={},           % <-- Forces it to treat everything as plain text
        basicstyle=\ttfamily\small, 
        breaklines=true,        
        breakatwhitespace=true,
        columns=fullflexible
    }
}
You are an expert Performance Engineer specializing in CUDA and PyTorch internals.

### STRICT MANDATE: TOTAL CONVERSION
You must optimize the provided architecture, named `Model`, by replacing standard PyTorch operator with a custom CUDA kernel.
1. **No Standard Ops:** You are STRICTLY FORBIDDEN from using standard PyTorch compute functions (e.g., `torch.matmul`, `torch.relu`) in the forward pass.
2. **Fusion:** You may fuse consecutive operators (e.g., merging MatMul + Bias + Activation) to minimize memory access.
3. **Scope of Work:** Focus ONLY on the nn.Module class definition. Ignore any other codes, for example codes that are related to testing.

### CRITICAL: WEIGHTS VS COMPUTATION
The optimized `ModelNew` will load pre-trained weights from the original `Model` when testing, but you MUST NOT use the original PyTorch implementations for compute.

**1. Rules for `__init__` (Preserve Structure):**
You MUST preserve the original `nn.Module` definitions (e.g., `self.conv = nn.Conv2d(...)`). This is required solely to ensure variable names and shapes match the original `state_dict` for loading weights.

**2. Rules for `forward` (Custom Compute):**
You are **STRICTLY FORBIDDEN** from calling the module objects directly. You must access their underlying parameters and pass them to your custom kernels.
E.g.,
* **FORBIDDEN:** `out = self.conv(x)` or `out = self.fc(x)`
* **REQUIRED:** `out = custom_kernel(x, self.conv.weight, self.conv.bias)`

### Reference Examples (This is only for one-shot example guideline. Not the real model you need to optimize.)
Here are examples of how to replace PyTorch operators with {{backend_display}} while maintaining state dict compatibility:
{{one_shot_example}}

### Input Architecture (This is the actual model you need to optimize.)
Optimize the following PyTorch model definition:
```python
{{reference_pytorch_code}}

```

### Output Requirements

* **Target:** Name the optimized output architecture `ModelNew`.
* **Quality:** Generate REAL, compilable code. NOT pseudocode. Include all necessary dependencies and helper functions.
* **Formatting:** Output ONLY the code block. Do not include testing code, training loops, or conversational text.
* **load_inline() arguments**: Leave the parameters extra_cflags, extra_ldflags args empty. Only include necessary args like name, cpp_sources, cuda_sources, functions and set verbose=False.
\end{tcblisting}

\section{Example of Six-Step SFT Data}
\label{app:six_step_sft}
This example shows how a training sample decomposes kernel generation into semantic analysis, scalar math, shape calculus, indexing, boundary handling, and error pre-analysis before presenting the final CUDA implementation.

\begin{tcblisting}{
    colback=gray!5, 
    colframe=gray!40, 
    title=Example of Six-Step SFT Data,
    fonttitle=\bfseries,
    breakable,
    listing only,
    listing engine=listings,
    listing options={
        language={},           % <-- Forces it to treat everything as plain text
        basicstyle=\ttfamily\small, 
        breaklines=true,        
        breakatwhitespace=true,
        columns=fullflexible
    }
}

sft_data_enhanced = {
    ''instruction'': ''Please perform an in-depth operator translation from PyTorch to a CUDA Kernel. You must include an exhaustive derivation of the mathematics, shape transformations, and index mappings.'',
    ''input_pytorch_code'': ''''''
import torch
import torch.nn as nn

class Model(nn.Module):
    def __init__(self, num_groups, num_channels, eps=1e-5):
        super(Model, self).__init__()
        self.num_groups = num_groups
        self.eps = eps
        self.weight = nn.Parameter(torch.randn(num_channels))
        self.bias = nn.Parameter(torch.randn(num_channels))

    def forward(self, x):
        return torch.nn.functional.group_norm(x, self.num_groups, self.weight, self.bias, self.eps)
'''''',
    ''analysis_and_derivation'': ''''''
### 1. Global Analysis and Operator Decomposition
This operator is GroupNorm. Its core logic involves partitioning the channel dimension $C$ into $G$ groups, and performing independent normalization within each group (encompassing $C/G$ channels and all spatial pixels).
* **Computational Flow**: Input $x \rightarrow$ Grouping $\rightarrow$ Intra-group Mean $\mu$ $\rightarrow$ Intra-group Variance $\sigma^2$ $\rightarrow$ Standardization $\rightarrow$ Channel-wise Affine Transformation ($\gamma, \beta$).
* **Parameter Binding**:
    * `x`: `(N, C, H, W)`.
    * `weight` ($\gamma$): `(C,)`.
    * `bias` ($\beta$): `(C,)`.
    * `num_groups` ($G$): Scalar.

### 2. Atomic Mathematical Derivation
To facilitate parallelization, we decompose the algorithm into accumulable forms:
1.  **Statistic Accumulation**:
    * Total elements per group: $L = (C/G) \times H \times W$.
    * First moment (Sum): $S_1 = \sum_{i \in \text{Group}} x_i$.
    * Second moment (Square Sum): $S_2 = \sum_{i \in \text{Group}} x_i^2$.
2.  **Moment-derived Statistics**:
    * Mean: $\mu = S_1 / L$.
    * Variance: $\sigma^2 = (S_2 / L) - \mu^2$.
3.  **Standardization Transformation**:
    * $\hat{x}_i = (x_i - \mu) \cdot \frac{1}{\sqrt{\sigma^2 + \epsilon}}$.
4.  **Affine Transformation**:
    * $y_i = \hat{x}_i \cdot \gamma_c + \beta_c$.

### 3. Shape Calculus and Transformation Logic
* **Input Tensor Logical Reshape**: $(N, C, H, W) \xrightarrow{\text{Reshape}} (N, G, C//G, H, W)$.
* **Reduction Domain**:
    * For each instance $n \in [0, N-1]$ and group $g \in [0, G-1]$, summation is performed across three contiguous logical dimensions $(C//G, H, W)$.
* **Output Alignment**: The final output is restored to $(N, C, H, W)$, fully aligning its spatial structure with the input.

### 4. Deep Index Relational Mapping (Indexing Matrix)
Assuming the output coordinates are $(n, c, h, w)$, we must locate its corresponding group and associated statistics:
1.  **Global Linear Index Parsing**:
    * `idx = n * (C*H*W) + c * (H*W) + h * W + w`.
2.  **Logical Group Mapping**:
    * Group index $g = c // (C/G)$.
    * Intra-batch group offset: `group_idx_within_batch = n * G + g`.
3.  **Memory Access Layout (NCHW)**:
    * `input[idx]` points to the element currently being processed.
    * `weight[c]` and `bias[c]` point to the scaling weights corresponding to the current channel.
4.  **Intra-group Iterative Addressing (for Mean/Variance Computation)**:
    * Iterate over $ic \in [0, (C/G)-1], ih \in [0, H-1], iw \in [0, W-1]$.
    * `offset = ((n * C) + (g * (C/G) + ic)) * (H * W) + ih * W + iw`.

### 5. Boundary Handling and Type Safety
* **Grid Out-of-Bounds**: `if (idx >= N * C * H * W) return;` Prevents trailing threads from accessing illegal memory addresses.
* **Spatial Overflow**: When processing $H, W$, ensure proper multiplication order to prevent `int32` overflow (using `long long` or `size_t` for total calculations is recommended).
* **Numerical Stability**: $\epsilon$ must be added inside the `sqrt`, and if $\sigma^2$ results in a minuscule negative value due to floating-point inaccuracies, it must be clamped to 0 using `max`.

### 6. Error Anticipation and Pitfall Avoidance
* **Parameter Fallacies**: Channel dimensions must never be hardcoded. For instance, fixing `out_channels` to 16 in Case 0 is a critical error.
* **Logical Omissions**: The multiply-add operation for weight $\gamma$ and bias $\beta$ must be implemented; Dummy implementations (like the 0.01f padding in Case 2) are strictly prohibited.
* **Performance Bottlenecks**: Iterating over group elements using triple-nested loops within a single thread is highly inefficient. It is recommended to introduce Shared Memory reduction in advanced SFT stages.
'''''',
    ''output_cuda_kernel'': r''''''
#include <torch/extension.h>
#include <cuda_runtime.h>
#include <cmath>

__global__ void group_norm_kernel_detailed(
    const float* __restrict__ input,
    const float* __restrict__ weight,
    const float* __restrict__ bias,
    float* output,
    int N, int C, int H, int W, int G, float eps) {

    int idx = blockIdx.x * blockDim.x + threadIdx.x;
    int total_elements = N * C * H * W;
    if (idx >= total_elements) return;

    // 1. Reverse Coordinate Parsing
    int w = idx % W;
    int h = (idx / W) % H;
    int c = (idx / (W * H)) % C;
    int n = idx / (W * H * C);

    // 2. Locate Group Space
    int channels_per_group = C / G;
    int g = c / channels_per_group;
    int group_start_c = g * channels_per_group;
    int num_elements_in_group = channels_per_group * H * W;

    // 3. Compute First and Second Moments (Parallel optimized versions typically precompute these via independent kernels; full derivation shown here)
    float sum_val = 0.0f;
    float sum_sq_val = 0.0f;

    for (int ic = 0; ic < channels_per_group; ++ic) {
        int cur_c = group_start_c + ic;
        for (int ih = 0; ih < H; ++ih) {
            for (int iw = 0; iw < W; ++iw) {
                int cur_idx = ((n * C + cur_c) * H + ih) * W + iw;
                float val = input[cur_idx];
                sum_val += val;
                sum_sq_val += val * val;
            }
        }
    }

    // 4. Statistics Derivation
    float mean = sum_val / num_elements_in_group;
    float variance = (sum_sq_val / num_elements_in_group) - (mean * mean);
    if (variance < 0) variance = 0; // Floating-point precision safeguard
    float inv_std = 1.0f / sqrtf(variance + eps);

    // 5. Output Transformation
    // y = ( (x - mean) * inv_std ) * weight[c] + bias[c]
    float x_hat = (input[idx] - mean) * inv_std;
    output[idx] = x_hat * weight[c] + bias[c];
}

torch::Tensor group_norm_forward_cuda(torch::Tensor input, int num_groups, torch::Tensor weight, torch::Tensor bias, float eps) {
    const int N = input.size(0);
    const int C = input.size(1);
    const int H = input.size(2);
    const int W = input.size(3);

    auto output = torch::empty_like(input);
    int total_elements = N * C * H * W;
    int block_size = 256;
    int num_blocks = (total_elements + block_size - 1) / block_size;

    group_norm_kernel_detailed<<<num_blocks, block_size>>>(
        input.data_ptr<float>(),
        weight.data_ptr<float>(),
        bias.data_ptr<float>(),
        output.data_ptr<float>(),
        N, C, H, W, num_groups, eps
    );
    return output;
}
''''''
}
\end{tcblisting}

\section{Incorporating Shape Information}
\label{app:shape}


This appendix illustrates the concrete prompt format used for tensor-shape augmentation. The goal is not to expose an offline debugging log, but to show the exact information inserted into the generation prompt after the PyTorch reference code and before the model is asked to write a CUDA/MUSA kernel. We use a compact exclusive-cumsum example because it contains a common failure mode: the output length is $L$, while the intermediate cumulative-sum tensor has length $L-1$.

\begin{tcblisting}{
    colback=gray!5,
    colframe=gray!40,
    title=PyTorch reference used in the prompt,
    fonttitle=\bfseries,
    breakable,
    listing only,
    listing engine=listings,
    listing options={
        language=Python,           % <-- Forces it to treat everything as plain text
        basicstyle=\ttfamily\small,
        breaklines=true,
        breakatwhitespace=true,
        columns=fullflexible
    }
}

import torch
import torch.nn as nn

class Model(nn.Module):
  ''''''
  A model that performs an exclusive cumulative sum (does not include the current element).

  Parameters:
  dim (int): The dimension along which to perform the exclusive cumulative sum.
  ''''''

  def __init__(self, dim):
    super(Model, self).__init__()
    self.dim = dim

  def forward(self, x):
    cumsum = torch.cumsum(x.narrow(dim=self.dim, start=0, length=x.size(self.dim)-1), dim=self.dim)
    return torch.cat((torch.zeros_like(x.select(self.dim, 0).unsqueeze(self.dim)), cumsum), dim=self.dim)

batch_size = 128
input_shape = (4000,)
dim = 1

def get_inputs():
  return [torch.rand(batch_size, *input_shape)]

def get_init_inputs():
  return [dim]
\end{tcblisting}


Without explicit tensor metadata, the model must infer the shape transitions of \texttt{narrow}, \texttt{select}, \texttt{unsqueeze}, and \texttt{cat} purely from the code. This often leads to off-by-one reasoning errors or incorrect assumptions about contiguity. We therefore append a structured shape block immediately after the reference implementation:

\begin{tcblisting}{
    colback=gray!5,
    colframe=gray!40,
    title=Shape-enhanced prompt block,
    fonttitle=\bfseries,
    breakable,
    listing only,
    listing engine=listings,
    listing options={
        language={},
        basicstyle=\ttfamily\small,
        breaklines=true,
        breakatwhitespace=true,
        columns=fullflexible
    }
}
[Tensor Shape Information]
Assume input x has shape [B, L]. For this test case, B=128 and L=4000.
- input x: shape=[B, L], stride=[L, 1], contiguous=True
- x.narrow(dim=1, start=0, length=L-1): shape=[B, L-1], stride=[L, 1], contiguous=False
- torch.cumsum(..., dim=1): shape=[B, L-1], stride=[L-1, 1], contiguous=True
- x.select(dim=1, index=0): shape=[B], stride=[L], contiguous=False
- unsqueeze(select, dim=1): shape=[B, 1], stride=[L, 1], contiguous=False
- zeros_like(unsqueeze): shape=[B, 1], stride=[1, 1], contiguous=True
- output cat(..., dim=1): shape=[B, L], stride=[L, 1], contiguous=True
[Kernel Generation Hint]
Use the listed shape and stride information when deriving memory offsets.
\end{tcblisting}


The tensor metadata can be obtained either statically with \texttt{torch.fx} and \texttt{ShapeProp}, or symbolically with \texttt{torch.export} when dynamic dimensions must be preserved. In the paper we keep only the essential extraction logic, since the implementation is straightforward and the important modeling signal is the final prompt block.

\begin{tcblisting}{
    colback=gray!5,
    colframe=gray!40,
    title=Minimal metadata extraction sketch,
    fonttitle=\bfseries,
    breakable,
    listing only,
    listing engine=listings,
    listing options={
        language=Python,
        basicstyle=\ttfamily\small,
        breaklines=true,
        breakatwhitespace=true,
        columns=fullflexible
    }
}
# Static shapes with torch.fx
from torch.fx.passes.shape_prop import ShapeProp

gm = torch.fx.symbolic_trace(model)
ShapeProp(gm).propagate(example_inputs)
for node in gm.graph.nodes:
  meta = node.meta.get(''tensor_meta'')
  if meta is not None:
    record(node.name, shape=meta.shape, stride=meta.stride, dtype=meta.dtype)

# Symbolic shapes with torch.export
from torch.export import export

dynamic_shapes = {''x'': {0: torch.export.Dim(''B'', min=2),
                        1: torch.export.Dim(''L'', min=3)}}
ep = export(model, args=(torch.randn(128, 4000),), dynamic_shapes=dynamic_shapes)
for node in ep.graph.nodes:
  val = node.meta.get(''val'')
  if isinstance(val, torch.Tensor):
    record(node.name, shape=val.shape, stride=val.stride(), contiguous=val.is_contiguous())
\end{tcblisting}


For this example, the crucial correction is to distinguish the final output length $L$ from the intermediate cumulative-sum length $L-1$. The generated kernel should special-case the first column and use the shifted input range for all subsequent output positions. In our manual inspections, this kind of shape-enhanced prompt fixed several shape-mismatch and indexing errors, but we do not report it as an ablation result because a systematic controlled experiment has not yet been conducted.

\section{Basic Operator Data}
\label{app:basic_operator_data}
This appendix gives the prompt used to synthesize diverse implementation mechanisms for basic operators. The prompt asks the teacher model to enumerate naive, single-feature, algorithmic, and composite variants before concrete code generation.

\begin{tcblisting}{
    colback=gray!5, 
    colframe=gray!40, 
    % title=Example of Six-Step SFT Data,
    fonttitle=\bfseries,
    breakable,
    listing only,
    listing engine=listings,
    listing options={
        language={},           % <-- Forces it to treat everything as plain text
        basicstyle=\ttfamily\small, 
        breaklines=true,        
        breakatwhitespace=true,
        columns=fullflexible
    }
}
**System / Role:**
You are a Principal CUDA Algorithm Engineer. You specialize in designing high-performance, portable CUDA kernels. Your expertise lies in **general parallel algorithms** (tiling, reduction trees, scan, sort), **memory hierarchy tuning**, and **kernel selection heuristics**.

**Task:**
I will provide a design requirement in the form of JSON data. Your goal is to generate an **exhaustive** list of implementation variants.

**Required Variant Categories:**
You must provide candidates across these 4 categories. Provide candidates as many as you can.

1.  **Baseline:**
    * The Naive/Reference implementation (for correctness checking).
2.  **Single-Feature Optimizations (Isolation):**
    * Standard, isolated techniques applied one at a time.
    * *Examples:* ''Global Memory Coalescing only'', ''Shared Memory Tiling only'', ''Vectorized Loads (float4) only'', ''Register Caching only''.
3.  **Algorithmic & Input-Specialized:**
    * Mathematical transformations (e.g., Winograd vs. GEMM, Online Softmax).
    * Shape-specific optimizations (e.g., ''Tall-and-skinny'' matrix handling, specialized batch sizes).
4.  **Composite/Hybrid Variants (Max Performance):**
    * **Combine** the methods from Category 2 and 3.
    * *Example:* ''Tiled Shared Memory + Vectorized Loads + Double Buffering.''
    * *Example:* ''Persistent Threads + Warp-Shuffle Reduction + Atomics.''

**Input Design Requirement:**
{{JSON_DATA}}

**Guidelines for Output:**

1.  **Strict JSON:** The output must be valid JSON. No markdown formatting, no intro text.
2.  **JSON Safety:** Escape all LaTeX backslashes (e.g., `\\frac` instead of `\frac`).
3.  **Technical Depth:** In the `mechanism` field, you must specify:
    * **Thread Mapping:** 1D/2D Block dims, Grid dims, and Z-curve/Swizzle patterns if relevant.
    * **Memory Hierarchy:** Exact flow (Global -> Shared (no banks?) -> Reg). Mention `LDG`, `LDS`, `STS`.
    * **Sync Strategy:** Specifics like `__syncthreads()`, `__shfl_down_sync`, or `atomicAdd`.
    * **Resource Trade-offs:** Mention Register Pressure or Shared Memory constraints if the variant is resource-heavy.

**Output Schema:**
Strictly follow this JSON structure:

```json
{
  ''method_name_snake_case'': {
    ''category'': ''One of [Baseline, Single-Feature, Algorithmic, Composite]'',
    ''target_scenario'': ''Specific conditions (e.g., 'Large Batch', 'Debug', 'Max Throughput')'',
    ''mechanism'': ''Detailed technical explanation including math logic, memory banking, and sync primitives.''
  },
  ''next_method_name'': {
    ''category'': ''One of [Baseline, Single-Feature, Algorithmic, Composite]'',
    ''target_scenario'': ''...'',
    ''mechanism'': ''...''
  }
}
\end{tcblisting}

\section{NCU Data}
\label{app:ncu_data}

This appendix documents how profiling feedback is converted into supervised training examples. The full Nsight Compute CSV can be extremely long, so the training prompt contains only the kernel code, compact PyTorch/Nsys summaries, and a selected subset of NCU metrics that are most useful for diagnosing memory, occupancy, and instruction-throughput bottlenecks.

\subsection{NCU Data Sample}
\label{app:ncu_sample}

The following template shows the structure of one profiling-analysis sample. It is intentionally schematic: the placeholders are populated by the profiling pipeline after kernel execution and report filtering.

\begin{tcblisting}{
    colback=gray!5,
    colframe=gray!40,
    title=Profiling data sample template,
    fonttitle=\bfseries,
    breakable,
    listing only,
    listing engine=listings,
    listing options={
        language=Python,
        basicstyle=\ttfamily\small,
        breaklines=true,
        breakatwhitespace=true,
        columns=fullflexible,
        showstringspaces=false,  % <-- 禁止在字符串中显示可见空格
        showspaces=false         % <-- 禁止在普通代码中显示可见空格
    }
}
data_sample = f'''
This is a pytorch module with custom CUDA kernel:
```python
{kernel_code}
```
Here is the torch profile key averages result:
{torch_profile_res}
Here is the nsys reports:
cuda_api_sum:
{nsys_cuda_api_sum}
cuda_gpu_kern_sum:
{nsys_cuda_gpu_kern_sum}
cuda_gpu_mem_time_sum:
{nsys_cuda_gpu_mem_time_sum}
cuda_gpu_mem_size_sum:
{nsys_cuda_gpu_mem_size_sum}
Here are some columns selected from ncu reports:
{ncu_report_selected_columns}
'''
\end{tcblisting}

\subsection{Prompt Template for Querying the Teacher Model}
\label{app:teacherprompt}

The teacher model is asked to interpret the profiling data rather than merely restate timing numbers. The prompt emphasizes root-cause analysis of the custom kernel itself, avoiding recommendations that only reduce launch overhead or change unrelated host-side code.

\begin{tcblisting}{
    colback=gray!5,
    colframe=gray!40,
    title=Teacher-analysis prompt,
    fonttitle=\bfseries,
    breakable,
    listing only,
    listing engine=listings,
    listing options={
        language=Python,
        basicstyle=\ttfamily\small,
        breaklines=true,
        breakatwhitespace=true,
        columns=fullflexible,
        showstringspaces=false,
        showspaces=false
    }
}
question = f'''{data_sample}
Please analyze the profile reports and answer:
- What is the bottleneck of the custom kernel?
- Which code-level changes can improve kernel performance?
Focus on the fused/custom kernel itself, not on launch overhead or generic GPU underutilization.
'''
\end{tcblisting}


\subsection{Prompt Template for Kernel Code Optimization}
\label{app:kernelprompt}

After obtaining a bottleneck analysis, we construct an optimization prompt that preserves the original KernelBench task constraints while injecting the teacher model's diagnosis as targeted feedback.

\begin{tcblisting}{
    colback=gray!5,
    colframe=gray!40,
    title=Profile-guided optimization prompt,
    fonttitle=\bfseries,
    breakable,
    listing only,
    listing engine=listings,
    listing options={
        language=Python,
        basicstyle=\ttfamily\small,
        breaklines=true,
        breakatwhitespace=true,
        columns=fullflexible,
        showstringspaces=false,
        showspaces=false
    }
}
prompt = kernel_bench_query_template
prompt += f'''This is the CUDA kernel code generated last time:
```python
{v1_cuda_code}
```

Here is the profiling-based bottleneck analysis of the previous kernel:

{v1_analysis}

Optimize the CUDA kernel while preserving the original PyTorch semantics.
Return only the new code.
'''

\end{tcblisting}

\subsection{Reference Methodology for Extracting Nsys/NCU Text Reports}
\label{app:sysncu}

We collect Nsys and NCU reports after correctness has been verified. The raw reports are then filtered by kernel name and exported to text or CSV fields that can be inserted into the training prompt.

\begin{tcblisting}{
    colback=gray!5,
    colframe=gray!40,
    title=Profiling command sketch,
    fonttitle=\bfseries,
    breakable,
    listing only,
    listing engine=listings,
    listing options={
        language=bash,
        basicstyle=\ttfamily\small,
        breaklines=true,
        breakatwhitespace=true,
        columns=fullflexible,
        showstringspaces=false,
        showspaces=false
    }
}
# Nsys profiling command:
nsys profile --stats=true -o {save_root}/{sample_id}/nsys --force-overwrite true python {test_script} \
--sample_id {sample_id} --device_id {device_id} 
# Note: Nsys reports are subsequently filtered from the redirected log telemetry via regular expressions.

# NCU profiling command:
ncu --target-processes all -f -o {save_root}/{sample_id}/ncu python {test_script} \
--sample_id {sample_id} --device_id {device_id} 

# Export selected kernels to CSV:
ncu --import {profile_path}/{sample}/ncu.ncu-rep --csv --page raw --kernel-name regex:\''{'|'.join(kernel_names)}\'' --log-file {profile_path}/{sample}/ncu-filter.csv
\end{tcblisting}

\subsection{Extracting Key NCU Metrics}
\label{app:keyncu}


For training, we retain a compact set of metrics instead of the full CSV. Memory-bound kernels prioritize DRAM/L1/L2 throughput and memory-traffic indicators; compute-bound kernels prioritize SM utilization, issue activity, instruction-pipe activity, occupancy, register pressure, and shared-memory usage.

\begin{tcblisting}{
    colback=gray!5,
    colframe=gray!40,
    title=Representative selected NCU fields,
    fonttitle=\bfseries,
    breakable,
    listing only,
    listing engine=listings,
    listing options={
        language={},
        basicstyle=\ttfamily\small,
        breaklines=true,
        breakatwhitespace=true,
        columns=fullflexible
    }
}
Block Size,Grid Size,dram__cycles_active.avg.pct_of_peak_sustained_elapsed (%),launch__occupancy_per_block_size, sm__inst_executed.avg.pct_of_peak_sustained_elapsed (%),device__attribute_max_gpu_frequency_khz, device__attribute_max_mem_frequency_khz, gpc__cycles_elapsed.max (cycle),sm__cycles_active.avg (cycle),gpu__time_duration.sum (msecond),gpu__dram_throughput.avg.pct_of_peak_sustained_elapsed (%),l1tex__throughput.avg.pct_of_peak_sustained_active (%),lts__throughput.avg.pct_of_peak_sustained_elapsed (%),gpu__compute_memory_throughput.avg.pct_of_peak_sustained_elapsed (%),sm__throughput.avg.pct_of_peak_sustained_elapsed (%),sm__issue_active.avg.pct_of_peak_sustained_elapsed (%),sm__pipe_fma_cycles_active.avg.pct_of_peak_sustained_elapsed (%),sm__pipe_fp64_cycles_active.avg.pct_of_peak_sustained_elapsed (%),sm__warps_active.avg.pct_of_peak_sustained_active (%),launch__registers_per_thread (register/thread),launch__shared_mem_per_block (Kbyte/block)
\end{tcblisting}

\subsection{Concrete Profiling Example}
\label{app:ncu_concrete_example}

Here are some essential diagnostic patterns below: the kernel redundantly recomputes the same channel-wise norm for every output channel, and the profiling metrics reveal this as a memory-traffic bottleneck despite high occupancy.

\begin{tcblisting}{
    colback=gray!5,
    colframe=gray!40,
    title=Condensed NCU example,
    fonttitle=\bfseries,
    breakable,
    listing only,
    listing engine=listings,
    listing options={
        language={},
        basicstyle=\ttfamily\small,
        breaklines=true,
        breakatwhitespace=true,
        columns=fullflexible
    }
}
Kernel pattern:
  input shape = [B=32, C=64, H=64, W=64]
  one output thread computes output[b, c, h, w]
  inside each thread, the kernel loops over all C channels to compute sum_sq(b, h, w)

Selected profiling signals:
  gpu__time_duration.sum                         ~= 0.55 ms
  gpu__dram_throughput.avg.pct_of_peak           ~= 5.2%
  sm__inst_executed.avg.pct_of_peak              ~= 48%
  sm__warps_active.avg.pct_of_peak_sustained     ~= 95%
  launch__registers_per_thread                   = 29
  launch__shared_mem_per_block                   ~= 1 KB

Teacher-model diagnosis:
  - The kernel is memory-traffic inefficient because sum_sq(b, h, w) is recomputed C times.
  - The redundant reads inflate the channel-wise reduction from B*C*H*W to B*C*C*H*W accesses.
  - High occupancy does not remove the bottleneck because each active warp repeatedly fetches the same data.

Optimization direction:
  - Map one block or warp group to each (b, h, w) location.
  - Compute the channel reduction once using shared memory or warp-level reduction.
  - Broadcast the norm factor and write all channel outputs for that pixel location.
  - Preserve output shape and numerical semantics while reducing redundant global-memory reads.
\end{tcblisting}

\section{Operator Examples}
\label{app:usedop}
This appendix summarizes the operator taxonomy used to build and audit the task distribution. The categories are intended to expose the model to both common neural-network layers and shape-sensitive tensor transformations. We categorize the supported evaluation operators into 7 functional domains, encompassing a total of 77 canonical operator archetypes and their multi-dimensional variants:

\begin{description}
    \item[\textbf{Activation Functions (6 Operators)}] \hfill \\
    \textbf{Core Operators:} \texttt{ReLU}, \texttt{GELU}, \texttt{LeakyReLU}, \texttt{PReLU}, \texttt{Sigmoid}, \texttt{Softmax} \\
    \textbf{Functionality:} Introduce non-linear transformations to enhance model expressivity. \\
    \textbf{PyTorch Mapping:} Directly map to the layers of the same name under \texttt{torch.nn} or functions under \texttt{torch} (e.g., \texttt{torch.relu}).

    \item[\textbf{Mathematical \& Arithmetic Operations (19 Operators)}] \hfill \\
    \textbf{Core Operators:} \texttt{Sin}, \texttt{Cos}, \texttt{Asin}, \texttt{Acos}, \texttt{Tan}, \texttt{Atan}, \texttt{Abs}, \texttt{Neg}, \texttt{Pow}, \texttt{Floor}, \texttt{Ceil}, \texttt{Clip}, \texttt{Round}, \texttt{Sqrt}, \texttt{Log2}, \texttt{Add}, \texttt{Sub}, \texttt{Mul}, \texttt{Div} \\
    \textbf{Functionality:} Foundational numerical computation, trigonometric functions, power/root operations, rounding, and clipping boundaries. \\
    \textbf{PyTorch Mapping:} Map to functions of the same name under \texttt{torch} (e.g., \texttt{torch.sin}, \texttt{torch.add}), or Tensor instance methods (e.g., \texttt{x.pow()}).

    \item[\textbf{Comparison \& Logical Operations (9 Operators)}] \hfill \\
    \textbf{Core Operators:} \texttt{Max}, \texttt{Min}, \texttt{Equal}, \texttt{Greater}, \texttt{Less}, \texttt{And}, \texttt{Or}, \texttt{Xor}, \texttt{Where} \\
    \textbf{Functionality:} Element-wise tensor comparisons, logical evaluation, and conditional selection. \\
    \textbf{PyTorch Mapping:} \texttt{Max}/\texttt{Min} correspond to element-wise twin-tensor comparison via \texttt{torch.max}/\texttt{torch.min}; \texttt{Where} maps to \texttt{torch.where}; others map to \texttt{torch.logical\_*} or \texttt{Tensor.eq}/\texttt{gt}/\texttt{lt} methods.

    \item[\textbf{Tensor Manipulation \& Layout Transformation (20 Operators)}] \hfill \\
    \textbf{Core Operators:} \texttt{Slice}, \texttt{Reshape}, \texttt{Flatten}, \texttt{Transpose}, \texttt{Squeeze}, \texttt{ExpandLast1~4}, \texttt{ConstPad}, \texttt{ReflectPad}, \texttt{ReplicatePad}, \texttt{Concat1~5} \\
    \textbf{Functionality:} Dimensional reshaping, slicing, spatial padding, concatenation, and axis expansion. \\
    \textbf{PyTorch Mapping:} \texttt{Reshape}/\texttt{Flatten}/\texttt{Transpose} correspond to APIs under \texttt{torch}; padding operators map to padding layers under \texttt{torch.nn} (e.g., \texttt{ConstantPad2d}); \texttt{Concat} variations map to \texttt{torch.cat} across different input sequence lengths.

    \item[\textbf{Dimensional Reduction (6 Operators)}] \hfill \\
    \textbf{Core Operators:} \texttt{TorchReduceSum}, \texttt{ReduceMin}, \texttt{ReduceMax}, \texttt{ReduceMean}, \texttt{ArgMin}, \texttt{ArgMax} \\
    \textbf{Functionality:} Compute sum, extrema, mean, or indices of extrema along designated dimensions. \\
    \textbf{PyTorch Mapping:} Correspond to \texttt{torch.sum}, \texttt{torch.amin}, \texttt{torch.argmax}, etc., necessitating an explicit \texttt{dim} argument.

    \item[\textbf{Canonical Neural Network Layers (5 Operators)}] \hfill \\
    \textbf{Core Operators:} \texttt{Conv1d}, \texttt{NCHWConv2d}, \texttt{BatchNorm2d}, \texttt{MaxPool2d}, \texttt{AvgPool2d}, \texttt{Linear} \\
    \textbf{Functionality:} Core architectural layers in deep learning responsible for feature extraction and affine mapping. \\
    \textbf{PyTorch Mapping:} Directly map to layers of the same name under \texttt{torch.nn} (e.g., \texttt{Conv2d}, \texttt{BatchNorm2d}, \texttt{Linear}), where \texttt{NCHWConv2d} represents the explicit layout-adapted variant of \texttt{Conv2d}.

    \item[\textbf{Interpolation \& Type Casting (12 Operators)}] \hfill \\
    \textbf{Core Operators:} \texttt{NearestInterp}, \texttt{LinearInterp}, \texttt{BilinearInterp}, \texttt{BicubicInterp}, \texttt{TrilinearInterp}, \texttt{CastBool}, \texttt{CastF32}, \texttt{CastF64}, \texttt{CastI32}, \texttt{CastI64}, \texttt{Tril}, \texttt{Triu} \\
    \textbf{Functionality:} Tensor spatial interpolation scaling, primitive data type casting, and triangular matrix extraction. \\
    \textbf{PyTorch Mapping:} Interpolation variants correspond to \texttt{torch.nn.functional.interpolate} configured with divergent \texttt{mode} flags; \texttt{Cast} operators map to \texttt{Tensor.to(dtype=...)}; \texttt{Tril}/\texttt{Triu} map to \texttt{torch.tril}/\texttt{torch.triu}.
\end{description}

\section{Multi-turn Dialogue Data}
\label{app:multidialogue}

This appendix gives the feedback template used to convert MooreEval telemetry into the next user message during multi-turn training. Correct candidates receive optimization-oriented feedback, while failed candidates receive repair-oriented feedback that prioritizes compilation and correctness before performance.

\begin{tcblisting}{
    colback=gray!5,
    colframe=gray!40,
    title=Multi-turn pipeline prompt template,
    fonttitle=\bfseries,
    breakable,
    listing only,
    listing engine=listings,
    listing options={
        language={},
        basicstyle=\ttfamily\small,
        breaklines=true,
        breakatwhitespace=true,
        columns=fullflexible
    }
}
=== System Feedback for Round {round_num} ===
```json
{feedback_json},
```

### Correct
Task for Round {round_num+1} (Optimization):
Correctness already passed. Keep outputs and state_dict compatibility unchanged.
Optimize kernel/runtime performance only; avoid risky rewrites.'',
Use the full feedback JSON above as the source of truth.

Potential directions (Optional):
- Current kernel is slower than torch; reduce launch overhead and fuse adjacent ops.
- Speedup is small; improve memory access pattern and occupancy before adding complexity.
- Keep current correctness path and attempt safe micro-optimizations for extra throughput.
- Prefer coalesced memory access, fewer temporary tensors, and minimal host-side logic in forward().
- Keep state_dict compatibility and preserve exact output semantics while optimizing.
####################################################

### Wrong ###
Task for Round {round_num+1} (Optimization):
Compilation/correctness/runtime is not fully passed. Fix correctness first.
Use the full feedback JSON above as the source of truth.
Do not optimize performance before correctness is stable.

Potential directions (Optional):
- Fix Python syntax first: close quotes/code fences/triple strings and keep one complete ModelNew file.
- Avoid process-level CUDA init in generated code; do not spawn/fork subprocesses in model code.
- Keep imports and load_inline declarations deterministic; avoid dynamic side effects in module scope.
- Audit CUDA indexing/bounds checks carefully (idx < numel), and verify shape/stride assumptions.
- Write outputs to valid addresses only; avoid using input stride for output indexing unless layout matches.
- Match reference output shape exactly, including batch/axis order and broadcasting behavior.
- Preserve reference math semantics and dtype/precision path; check clamp/relu boundaries and reductions.
- Remove disallowed aten ops from forward(); move heavy ops into custom CUDA kernels.
####################################################

Return only Python code.
\end{tcblisting}

\section{Reviewer Data}
\label{app:reviewerexampple}

The reviewer prompt trains the model to audit a candidate CUDA kernel against a PyTorch reference. It requires an explicit correctness verdict, root-cause analysis for incorrect kernels, and performance analysis for correct kernels.

\begin{tcblisting}{
    colback=gray!5,
    colframe=gray!40,
    title=CUDA Kernel Review Instructions,
    fonttitle=\bfseries,
    breakable,
    listing only,
    listing engine=listings,
    listing options={
        language={},
        basicstyle=\ttfamily\small,
        breaklines=true,
        breakatwhitespace=true,
        columns=fullflexible
    }
}
You act as a professional CUDA kernel reviewer with deep expertise in PyTorch, CUDA programming model, GPU architecture, and high-performance computing (HPC) best practices. Your core task is to review a candidate CUDA kernel implementation against a reference PyTorch code snippet.

## Review Materials
[Pytorch Code]
```python
{pytorch_code}
```

[Candidate CUDA Kernel]
```cpp
{cuda_kernel_code}
```

## Core Task
Evaluate the correctness of the candidate CUDA kernel relative to the provided PyTorch code, and (if correct) analyze its performance characteristics; (if incorrect) identify and fix the bugs.

### Step 1: Correctness Verification
 - First, validate functional equivalence between the candidate CUDA kernel and the PyTorch code:
   - Confirm that the kernel implements the exact mathematical/logical operations of the PyTorch code (e.g., tensor computations, data transformations, boundary handling).
   - Check for critical correctness issues such as:
     - Mismatched tensor shapes/dimensions in kernel launch parameters (grid/block size).
     - Incorrect memory access (out-of-bounds, uncoalesced access, wrong indexing).
     - Invalid thread synchronization (missing/wrong __syncthreads()).
     - Incorrect handling of edge cases (e.g., tensors with non-multiple-of-block-size dimensions).
     - Type mismatches (e.g., float vs. half precision, integer overflow).
     - Incorrect CUDA API usage (e.g., cudaMalloc, cudaMemcpy errors).
     
### Step 2: Action Based on Correctness
 - If the kernel is INCORRECT:
   1. Clearly identify all bugs (list each bug with:
     - Exact location (line number/range in the kernel code).
     - Root cause (why the bug breaks functionality, e.g., ''Thread indexing uses x + y * width instead of y + x * width, leading to transposed memory access'').
     - Impact (what incorrect behavior/output this bug causes, e.g., ''Results in wrong tensor values for non-square matrices'').
   2. Provide a complete, fixed version of the CUDA kernel code (with comments highlighting the fixes).
   3. Explain the rationale for each fix (how the change resolves the specific bug and aligns the kernel with the PyTorch code).
 - If the kernel is CORRECT:
   1. Conduct a comprehensive performance analysis of the kernel, covering (at minimum) the following dimensions:
     - Memory access pattern: Coalescing (whether global memory access is coalesced), bank conflicts (shared memory), cache utilization (L1/L2 hit rate potential).
     - Parallelism: Thread block/grid configuration (optimal vs. suboptimal), warp utilization, occupancy (active warps per SM, limiting factors like register/shared memory usage).
     - Divergence: Warp divergence (e.g., conditional branches leading to inactive threads), control flow efficiency.
     - Arithmetic intensity: Ratio of compute operations to memory operations (GPU utilization potential).
     - Optimization opportunities: Specific, actionable suggestions to improve performance (e.g., ''Use shared memory to reduce global memory access for repeated tensor elements'' or ''Adjust block size from 128 to 256 to increase occupancy'').

### Step 3: Final Verdict
 - End your review with an explicit verdict in the exact format:
   `VERDICT: CORRECT` or `VERDICT: INCORRECT`

### Additional Requirements
 - Your review must be structured, easy to follow, and use technical terminology accurately (with brief explanations if needed for clarity).
 - For code fixes/optimizations, prioritize CUDA best practices (e.g., avoiding global memory bottlenecks, maximizing SM utilization).
 - Focus only on the kernel's alignment with the provided PyTorch code-do not add unrelated functionality or optimizations.
\end{tcblisting}

\section{MooreEval System Implementation Details}
\label{app:moore_eval_system}

\subsection{Design Objectives}

MooreEval is engineered as a highly scalable, executable evaluation environment tailored specifically for GPU kernel generation tasks. In scenarios where Large Language Models (LLMs) are trained via reinforcement learning to synthesize CUDA/MUSA kernels, the evaluation framework ceases to be a passive offline benchmarking utility; instead, it transmutes into the foundational closed-loop \textit{reward environment}. Massive volumes of candidate codes emitted by the policy model at each rollout iteration must be rapidly compiled, executed, and validated for semantic correctness, while concurrently benchmarking empirical performance gains relative to the PyTorch baseline. These multidimensional telemetry data are subsequently synthesized into scalar reward signals that directly orchestrate the optimization trajectory of the policy model.

Conventional single-node synchronous evaluation scripts are fundamentally inadequate for the stringent demands of this paradigm. First, the high-concurrency rollouts within each training batch generate an immense influx of candidate implementations, demanding an execution backend capable of sustained high throughput and long-term multi-node stability. Second, model-generated CUDA/MUSA code is inherently untrusted. It can precipitate an array of critical failure modes, including Python syntax anomalies, NVCC compilation or linkage failures, runtime illegal memory accesses, GPU kernel hangs, Out-of-Memory (OOM) exceptions, numerical discrepancies, and adversarial attempts to exploit high-level PyTorch/\texttt{aten::*} operators to bypass native kernel authoring. If the evaluation framework merely returns a primitive binary pass/fail verdict, the training pipeline suffers from severe reward opacity, crippling fault analysis and making it impossible to differentiate between model algorithmic errors and infrastructure anomalies.

To eliminate these bottlenecks, MooreEval seamlessly transcribes raw model-generated kernel outputs into deterministic, scalable, and highly interpretable reinforcement learning rewards. Architecturally, its implementation is anchored by the following core system requirements:

\begin{itemize}
    \item \textbf{High Throughput.} During RL training, a single batch encapsulates numerous model responses, each necessitating independent compilation and GPU-accelerated execution. MooreEval explicitly decouples CPU-bound compilation from GPU-bound execution, enabling both resource pools to scale horizontally and independently. This asymmetric scaling effectively prevents GPU starvation during heavy compilation workloads and eliminates CPU processing bottlenecks during execution, maximizing cluster-wide hardware utilization.
    
    \item \textbf{Robustness and Fault Isolation.} Because candidate kernel code is fundamentally untrusted, it can induce compiler deadlocks, critical runtime crashes, GPU execution timeouts, or catastrophic parent-process crashes. MooreEval orchestrates task execution via distributed task queues, explicit inflight state management, visibility timeouts, bounded retry policies, and rigid sandbox subprocess isolation. This multi-layered containment strategy guarantees that anomalous or malicious samples are securely isolated, proactively shielding the global training loop from disruption.
    
    \item \textbf{Reward Interpretability.} Beyond emitting a coarse scalar reward, MooreEval captures and structures granular execution metadata, including compilation states, precise correctness metrics, historical speedup ratios, specific error taxonomies, condensed log diagnostics, execution latencies, and disallowed operator profiles. These highly structured fields provide immediate actionability for reward function shaping, cluster telemetry monitoring, long-tail failure debugging, and multi-turn conversational feedback generation.
    
    \item \textbf{Anti-Hacking Resilience.} In native kernel generation tasks, the model frequently attempts to game the reward function by invoking optimized black-box routines like \texttt{torch.matmul} or APIs within \texttt{torch.nn.functional} to yield correct numerical outputs while circumventing actual custom kernel authoring. MooreEval combats this via a hybrid verification stack combining static analysis with a runtime profiler to trap forbidden \texttt{aten::*} operations, classifying them under an isolated, non-rewardable infraction taxonomy.
    
    \item \textbf{Seamless RL Framework Abstraction.} Via a unified Reward Adapter interface, MooreEval abstracts the underlying distributed infrastructure into a standard, deterministic reward function. It natively provisions large-batch rewards, single-sample telemetry, and iterative multi-turn conversational feedback. This modular abstraction ensures that the core training framework remains entirely agnostic to low-level implementation complexities, such as Redis queue operations, worker scheduling, sandbox orchestration, and cross-node telemetry aggregation.
\end{itemize}

\subsection{Distributed Overall Architecture}

\begin{figure}[t]
    \centering
    \includegraphics[width=0.95\linewidth]{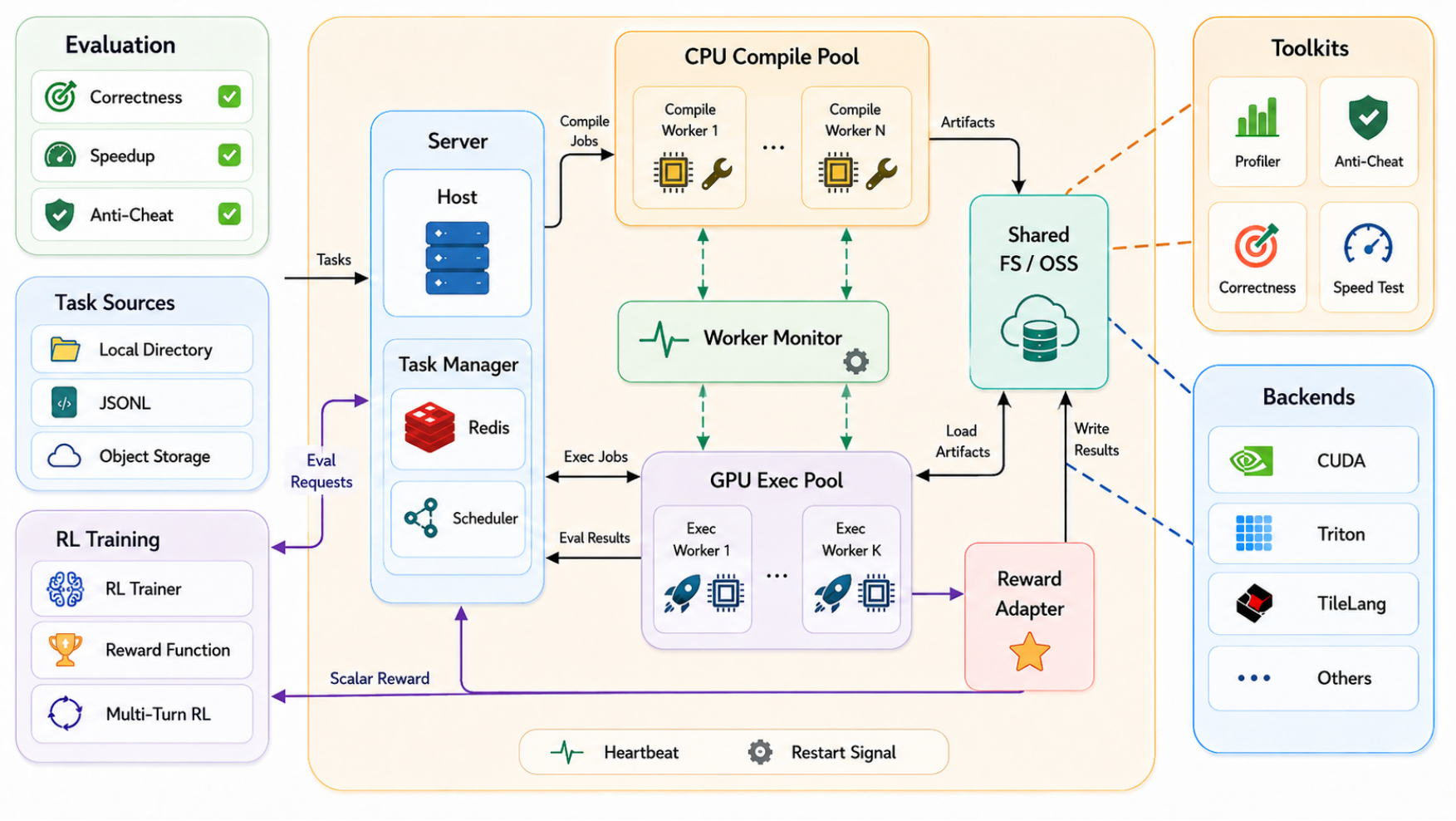}
    \caption{Architecture of \textbf{MooreEval}, a scalable execution-based evaluation environment for compiling, verifying, profiling, and rewarding generated native GPU kernels.
}
    \label{fig:moore_eval_arch}
\end{figure}

As illustrated in Figure~\ref{fig:moore_eval_arch}, MooreEval adopts a distributed asynchronous pipeline architecture, primarily conceptualized into six core components: the Host, Compile Workers, Exec Workers, a Redis Queue infrastructure, a Shared Filesystem/OSS layer, and the Reward Adapter.

\paragraph{Host.}
The Host acts as the orchestrator that ingests pending telemetry samples from various task sources, encapsulates them into standardized tracking payloads, and enqueues them into the compilation pipeline. Ingest sources natively span local directories, jsonl data streams, or raw payloads in object storage. Concurrently, the Host runs an asynchronous aggregator that polls finalized evaluations from the result queue, subsequently committing them back to persistent storage or delivering them to the Reward Adapter.

\paragraph{Compile Worker.}
Compile Workers are dedicated horizontally scalable nodes tasked with CPU-bound pre-compilation. A worker dequeues a task from the compilation pipeline, mirrors the associated PyTorch reference and candidate implementation into a deterministic workspace directory, and invokes the host compiler. Upon a successful build, the worker instantiates a \texttt{CompileArtifact}—cataloging the binary paths, compilation latency, return codes, and condensed log diagnostics—and forwards the task into the execution queue. Conversely, if compilation fails, the worker immediately synthesizes a terminal failure artifact, short-circuiting the pipeline to preempt the allocation of downstream GPU execution resources.

\paragraph{Exec Worker.}
Exec Workers manage the GPU-bound execution phase. A worker pulls a compiled artifact from the execution queue, allocates a specific target GPU device, and spawns an isolated sandbox subprocess. Inside this sandbox, it dynamically loads both the PyTorch reference baseline and the model-generated \texttt{ModelNew} kernel, sequentially orchestrating randomized inputs verification, anti-hacking detection, and micro-benchmarking speed tests. The consolidated execution telemetry is subsequently pushed back into the result queue for the Host to aggregate.

\paragraph{Redis Queue and Inflight State Management.}
The Redis infrastructure serves a dual purpose as both the distributed task backbone and the ledger for inflight state management. MooreEval utilizes Redis lists to maintain separate task queues across different lifecycle stages, while maintaining concurrent inflight sorted sets (ZSET) to track active tasks alongside their respective visibility deadlines. In the event of a worker crash, node eviction, or unacknowledged task timeout, an active sweeper daemon automatically reinstates the task into the primary queue and increments its retry attempt counter, providing robust at-least-once execution guarantees.

\paragraph{Shared Filesystem / OSS.}
The shared storage layer provisions unified persistence for raw source code, volatile build workspaces, compiled object binaries, and runtime validation artifacts. Because Compile Workers and Exec Workers are designed to scale independently across distinct physical topologies, a shared filesystem (e.g., NFS, Lustre) or object storage service (OSS) serves as the centralized medium to pass immutable build artifacts seamlessly between CPU compilation clusters and GPU execution nodes.

\paragraph{Reward Adapter.}
The Reward Adapter functions as the definitive API abstraction layer integrating MooreEval with the reinforcement learning framework. Upon kernel token generation, the adapter stamps each candidate response with a unique tracking UUID, dispatches the assessment payloads to the distributed cluster, and awaits aggregation. Ultimately, the Reward Adapter maps the low-level \texttt{FinalResult} telemetry into clean, training-ready fields, exposing metrics such as scalar scores, functional accuracy, architectural legality, and empirical speedups.

\subsection{Tiered Heterogeneous Evaluation Pipeline}

MooreEval explicitly bifurcates the evaluation of a single candidate kernel into a CPU-bound compilation phase and a GPU-bound execution phase. This architectural decision stems from the inherent \textit{resource heterogeneity} of the kernel benchmarking lifecycle: compiling candidate source code primarily saturates CPU cores, host memory, compiler processes, and filesystem IOPS, whereas validating semantic correctness and profiling execution efficiency exclusively monopolize GPU compute and device VRAM resources. Monolytically binding these asymmetric operations within a synchronous execution thread inevitably precipitates severe resource throttling—leaving the GPU idling or stalled during heavy compilation bottlenecks, or forcing the CPU into resource-starvation states during prolonged asynchronous GPU execution sweeps. Consequently, MooreEval fully decouples compilation from execution, allowing both resource domains to scale and undergo scheduling independently.

During the compilation phase, the Host encapsulates the model-generated implementation into a standardized \texttt{TaskPayload} and dispatches it into the Redis-backed compilation queue. Upon popping a task payload, a Compile Worker clones the PyTorch reference and the candidate kernel source into a deterministic sandbox build directory hosted on the shared filesystem. The pathing schema for this workspace is uniquely mapped via the combination of the tracking task UUID and the candidate code's cryptographic hash, neutralizing the risk of path-collision race conditions during high-concurrency compilation batches. If the host compiler returns a successful exit code, the Compile Worker instantiates a persistent \texttt{CompileArtifact} and enqueues the metadata into the downstream execution pipeline. Conversely, if compilation fails, the worker immediately traps the error taxonomy and exits, terminating the task prior to downstream GPU allocation.

During the execution phase, an Exec Worker pulls a verified compiled artifact from the execution queue and spawns an isolated sandbox subprocess bound to a dedicated target GPU device. The sandbox orchestrates the dynamic library loading of the PyTorch reference baseline and the model-generated \texttt{ModelNew} extension, subsequently executing randomized input validation, anti-hacking detection, and micro-benchmarking profiling sweeps. MooreEval enforces a strict \textit{correctness-first} dependency graph: the pipeline proceeds to forbidden operator inspection and speedup measurement if and only if the candidate kernel successfully satisfies all tensor shape constraints, dtype alignments, and numerical precision tolerance thresholds. This sequential gating ensures that performance bonuses are exclusively predicated on functional completeness, preventing mathematically broken implementations from acquiring illusory positive rewards via speculative optimizations or fake acceleration behavior.

Inter-stage state orchestration and metadata handshakes between the dual phases are facilitated via the Redis queue fabric and the shared filesystem medium. Redis acts as the centralized transaction ledger governing task state transitions, while the shared storage fabric provisions hot persistence for raw source trees, compiled binary objects, and volatile verification inputs. This architectural decoupling allows compute nodes with dense CPU profiles to scale out Compile Workers, while dedicated GPU clusters remain narrowly focused on code execution and fine-grained performance profiling. For reinforcement learning pipelines, this heterogeneous pipeline architecture optimally services the massive concurrency spikes characteristic of batch reward evaluations.

Furthermore, MooreEval integrates deterministic failure isolation and fault recovery heuristics within the execution loop. Tasks popped from active queues are immediately transitioned into an active \textit{inflight} ledger configured with a dynamic visibility timeout. In the event of an unacknowledged worker crash, cluster node eviction, or heartbeat timeout, an active sweeper daemon automatically reinstates the compromised payload back into the primary staging queue. For transient compilation or execution anomalies driven by underlying \textit{resource jitter}, the system tolerates a bounded retry allocation. Conversely, for \textit{deterministic software anomalies} inherent to the candidate code itself—such as syntactic violations, illegal memory access traps, or mathematical mismatches—the evaluation engine immediately truncates the task and records the specific error taxonomy. This dual-path fault handling reduces infrastructural noise during long-run training while aggressively eliminating redundant evaluations that waste cluster compute cycles.

\subsection{Structured Telemetry and Reward Modeling}

Rather than crudely truncating the evaluation pipeline into a binary pass/fail verdict, MooreEval decomposes the entire bench-running lifespan into granular, structured data components. This rich telemetry equips the reinforcement learning framework with high-fidelity, interpretable, and mathematically stable optimization signals. Each inbound evaluation target is encapsulated within a standardized \texttt{TaskPayload}, comprising the source code definitions of the PyTorch reference baseline, the model-generated candidate implementation, lineage metadata tracing the task origin, an incremental retry index, and an embedded \texttt{EvalConfig} ledger. The \texttt{EvalConfig} explicitly parameterizes execution behaviors, defining randomized correctness permutation test counts, performance micro-benchmarking sweeps, numerical precision tolerance thresholds, hardware timing and synchronization modalities, and gating semantics for forbidden operator capture and speculative excessive speedup clipping. This precise task representation matrix enables MooreEval to fluidly adapt its runtime evaluation heuristics across disparate post-training regimes while enforcing immutable data serialization contracts across distributed cluster workers.

Throughout the execution topology, MooreEval independently intercepts and structures intermediate telemetry from both the compilation and execution loops. Compile Workers instantiate a persistent \texttt{CompileArtifact}, archiving binary success states, compilation destination targets, process return codes, condensed stderr diagnostics, compilation latency, and explicit syntactic error taxonomies. Downstream, Exec Workers populate an \texttt{ExecResult} object, logging functional alignment rates, candidate runtime execution latencies, reference runtime execution latencies, empirical speedup ratios, forbidden \texttt{aten::*} hook exceptions, benchmarking profiling metrics, and definitive runtime fault taxonomies. Ultimately, the Host synchronizes and merges these asynchronous state objects into a unified \texttt{FinalResult} payload. This structured data schema not only feeds the upstream Reward Adapter for score calculation, but also acts as the primary data engine driving failure distribution statistics, automated performance regression analysis, and cluster-wide training curve monitoring.

In terms of algorithmic optimization, MooreEval enforces a \textit{hierarchical reward modeling strategy}. First, in the event of extraction failures, compilation aborts, runtime execution panics, catastrophic numerical deviations, or the explicit detection of a forbidden PyTorch/\texttt{aten::*} fallback operator, the candidate implementation is immediately assigned a hard failure penalty. This negative anchor aggressively suppresses policy distribution drift toward unexecutable or adversarial trajectories. Second, for candidate codes that compile successfully but fail to perfectly fulfill all randomized correctness assertions, MooreEval provisions a bounded, fractional reward proportional to the exact functional accuracy or failure severity. This maintains a weak but continuous \textit{reward shaping signal} that guides the model through the critical sparse-to-dense exploratory transition from completely broken variants to partially functioning code blocks. Finally, the performance metric enters the reward optimization loop if and only if the candidate implementation cleanly satisfies all functional correctness and architectural compliance constraints. Validated implementations receive a positive base scalar reward, augmented by a clipped speedup bonus if they deliver empirical runtime gains relative to the PyTorch baseline. This cascading reward architecture mathematically guarantees that the gradient updates strictly prioritize execution safety and functional integrity over performance tuning, systematically neutralizing the risk of the model gaming the system via computational truncation or black-box operator fallbacks.

\subsection{Error Taxonomies, Anti-Hacking Safeguards, and Feedback Diagnostics}

To guarantee that evaluation telemetries can be robustly consumed by the reinforcement learning pipeline, MooreEval formalizes a unified and immutable error taxonomy across all computational tasks. Each evaluation instance is deterministically cataloged into a granular \texttt{error\_category}, structurally defined as follows:
\begin{itemize}
    \item \texttt{\_ok}: The candidate implementation successfully satisfies all host compilation, functional correctness, and architectural compliance checks.
    \item \texttt{compile\_error:*}: Captures Python syntactic anomalies, NVCC compilation aborts, downstream linkage errors, missing header file directives, or compiler execution timeouts.
    \item \texttt{runtime\_error:*}: Flags runtime illegal memory accesses, CUDA driver launch failures, device-side assertion traps (\texttt{TORCH\_USE\_CUDA\_DSA}), host/device Out-of-Memory (OOM) faults, or execution wall-clock timeouts.
    \item \texttt{correctness\_error:*}: Identifies tensor shape mismatches, primitive dtype alignment failures, or numerical precision tolerance deviations.
    \item \texttt{cheating:*}: Marks forbidden PyTorch/\texttt{aten::*} high-level computational fallbacks or statistically anomalous speculative speedup metrics.
    \item \texttt{infra:*}: Denotes asynchronous worker panics, sandbox initialization aborts, task retry exhaustion, or transient distributed infrastructure exceptions.
\end{itemize}

This formal error taxonomy establishes a critical dichotomy between deterministic execution anomalies triggered by model-authored source code and non-deterministic structural faults induced by underlying infrastructure states. For software bugs inherent to the candidate code itself—such as syntax syntax violations, out-of-bounds pointer dereferences, or numerical deviations—MooreEval immediately records the specific sub-category, outputting it as a direct gradient optimization signal reflecting inadequate generation quality. Conversely, for anomalies precipitated by the evaluation runtime environment—such as node worker crashes, sandbox orchestration aborts, or scheduling timeout exhaustion—the system enqueues them into the isolated \texttt{infra:*} domain. This rigorous decoupling prevents transient environmental jitter from being misattributed to the policy model as fake negative algorithmic feedback.

The anti-hacking safeguard represents a cornerstone design within MooreEval to enforce absolute reward integrity. In primitive kernel synthesis tasks, the policy model frequently attempts to adversarial exploit high-level black-box operations—such as directly invoking \texttt{torch.matmul}, \texttt{torch.nn.functional}, or other hyper-optimized \texttt{aten::*} symbols—to effortlessly yield valid numerical outputs while circumventing actual hardware-level custom kernel engineering. To secure this perimeter, MooreEval deploys a defensive stack blending static syntax-tree inspection with a localized runtime profiler hook to trap prohibited execution traces, permanently branding such infractions as \texttt{cheating:disallowed\_aten}. Furthermore, the framework monitors for statistically improbable execution performance metrics; if a candidate code exhibits an empirical speedup surpassing a physically realistic optimization threshold, it is automatically flagged as \texttt{cheating:excessive\_speedup}. This dual-gate architecture mathematically guarantees that the policy model can harvest positive performance bonuses if and only if its native implementation is holistically sound, safe, and compliant with architectural constraints.

Leveraging this structured telemetry ledger, MooreEval further synthesizes highly contextual, actionable natural language diagnostic signals tailored for multi-turn reinforcement learning. For instance, when a candidate kernel breaks during compilation, the diagnostic payload aggregates a condensed stderr diagnostic block localized to the failing lines. When a tensor payload triggers shape, type, or precision alignment violations, the feedback pinpoints the exact multi-dimensional offset discrepancy. Similarly, upon trapping a forbidden operator signature, the feedback emits an explicit compliance warning, compelling the model to author a custom native block rather than leaning on preexisting PyTorch library abstractions. Compared to a primitive, uninformative binary failure penalty, this error taxonomy and automated diagnostic engine empower the model to execute target-oriented, multi-turn bug rectifications while rendering the global failure distribution profile entirely transparent and interpretable throughout the training lifespan.

\subsection{Reinforcement Learning Training Integration}

MooreEval integrates with the reinforcement learning training pipeline via a dedicated Reward Adapter, encapsulating the underlying distributed evaluation infrastructure into a standard, deterministic reward function interface. In a \texttt{verl}-style learning framework, upon the policy model emitting candidate kernel tokens, the distributed trainer invokes the standardized \texttt{compute\_score} API. The Reward Adapter immediately stamps each candidate within the batch with a unique orchestration UUID, dispatching the PyTorch reference code, the model-generated output string, and runtime evaluation configurations into the primary pending task ledger. Subsequently, MooreEval's Host, Compile Workers, and Exec Workers asynchronously execute host compilation, randomized input verification, anti-hacking checks, and micro-benchmarking sweeps, ultimately consolidating the execution lifecycle into a unified \texttt{FinalResult}. The Reward Adapter then transcribes this low-level results payload into standard, algorithm-facing metrics such as scalar scores, functional accuracy, compliance legality, and empirical speedups.

Through this monolithic abstraction layer, the RL trainer remains entirely agnostic to low-level engineering complexities—such as Redis transaction buffers, distributed worker orchestration, sandbox execution containerization, and cross-node telemetry aggregation—requiring only a single, uniform reward interface invocation to establish the complete optimization closed loop. MooreEval natively provisions dual support for high-throughput batch evaluations, single-sample runtime tracking, and iterative multi-turn conversational feedback loops. Under the batched reward regime, the Reward Adapter concurrently dispatches bulk evaluation payloads and polls the transaction ledger, optimally satisfying the massive, non-blocking concurrency requirements characteristic of RL rollout phases. In the single-sample modality, the environment is tailored for diagnostic isolation or sequential score computations. In multi-turn feedback mode, MooreEval archives the execution metrics across each conversational turn, enabling the synthesis of complex trajectory-level rewards predicated on the earliest successful turn, the absolute peak performance epoch, or first-turn-anchored trajectory formulations.

On the data engineering front, MooreEval supplies a comprehensive, KernelBench-compatible data parsing blueprint, converting raw, idiomatic PyTorch model configurations into high-fidelity ingestion tokens digestible by the post-training framework. Each decoupled training sample maps a static system instruction prompt, a descriptive user task schema, and the foundational baseline code acting as the computational ground truth, alongside lineage keys and split boundaries embedded within a structured \texttt{extra\_info} metadata slot. The policy model is mathematically compelled to author a custom \texttt{ModelNew} extension adhering to the exact interface signatures of the reference architecture, completely swapping out black-box PyTorch operations for custom hardware-native kernels. Governed by this design, the curated data distribution, the executable evaluation sandbox, and the programmatic reward function interface holistically unify to construct a highly robust, scalable reinforcement learning training closed loop specifically optimized for GPU kernel synthesis tasks.

\section{Allowed ATen Operators}
\label{app:op_whitelist}
\begin{oplist}[3]
\op{aten::_assert_async}
\op{aten::_assert_async.msg}
\op{aten::_assert_scalar}
\op{aten::_assert_tensor_metadata}
\op{aten::_cast_Byte}
\op{aten::_cast_Char}
\op{aten::_cast_Double}
\op{aten::_cast_Float}
\op{aten::_cast_Half}
\op{aten::_cast_Int}
\op{aten::_cast_Long}
\op{aten::_cast_Short}
\op{aten::alias}
\op{aten::arange}
\op{aten::as_strided}
\op{aten::broadcast_to}
\op{aten::clone}
\op{aten::contiguous}
\op{aten::copy_}
\op{aten::empty}
\op{aten::empty_like}
\op{aten::expand}
\op{aten::flatten}
\op{aten::full}
\op{aten::ones}
\op{aten::permute}
\op{aten::rand}
\op{aten::randint}
\op{aten::randn}
\op{aten::reshape}
\op{aten::select}
\op{aten::slice}
\op{aten::squeeze}
\op{aten::stride}
\op{aten::transpose}
\op{aten::unsqueeze}
\op{aten::view}
\op{aten::zeros}
\op{aten::_local_scalar_dense}
\op{aten::allclose}
\op{aten::equal}
\op{aten::item}
\end{oplist}

\section{Off-policy Cancellation Example}
\label{app:Cancellation_case}

\begin{figure}[p]
    \centering
    \includegraphics[
        width=\textwidth,
        height=1.0\textheight,
        keepaspectratio
    ]{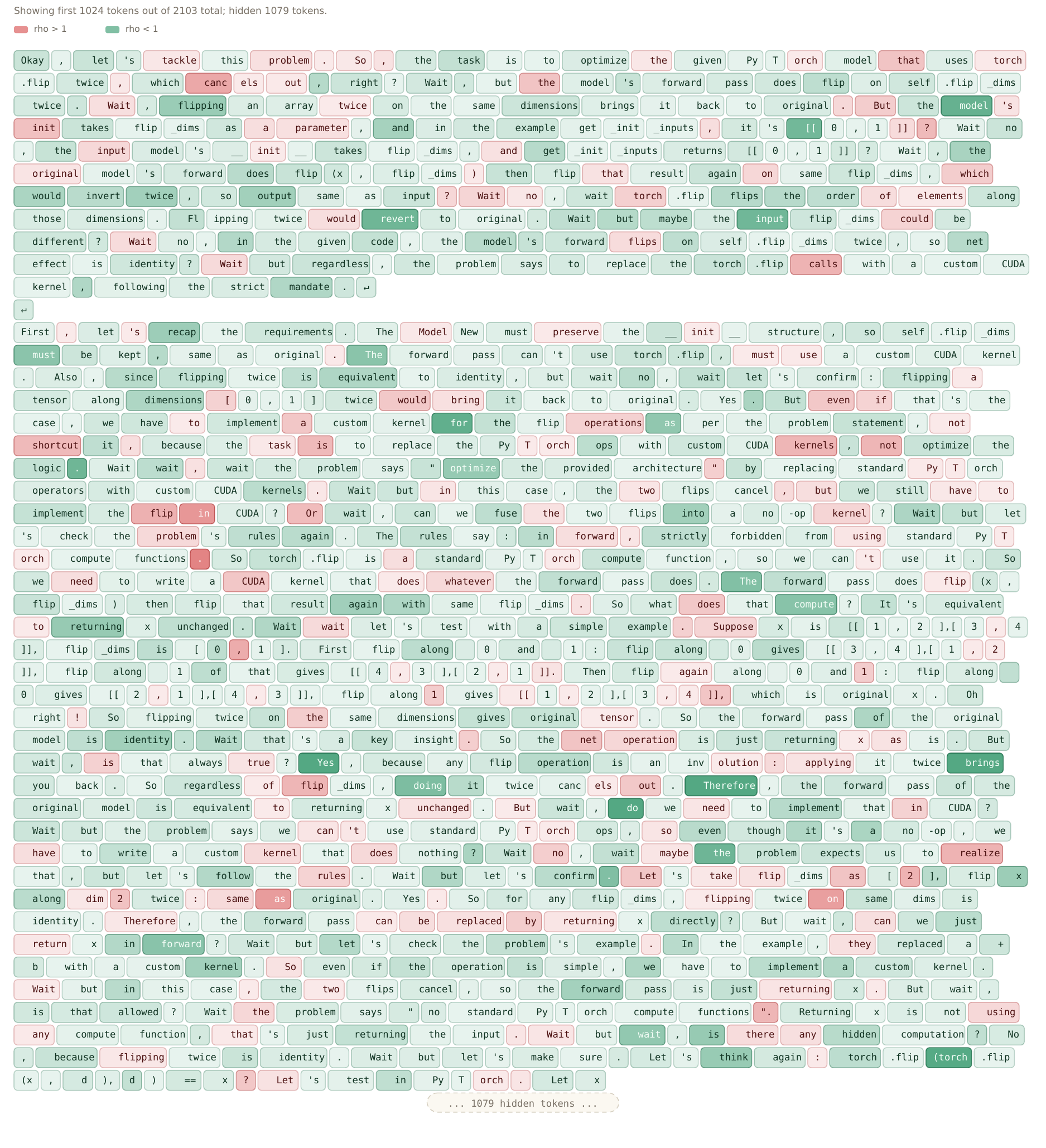}
    \caption{Cancellation case in vanilla off-policy sequence masking. Red tokens indicate $\rho_t>1$, and green tokens indicate $\rho_t<1$; darker colors denote larger deviations from 1. When signed log-ratios are averaged (or ratios are multiplied) across the sequence, these opposite deviations can cancel each other out, making a strongly off-policy sequence appear nearly on-policy in extreme cases.}
    \label{fig:cancellation_case}
\end{figure}

Figure~\ref{fig:cancellation_case} provides a concrete example of the off-policy cancellation effect discussed in the main text. Each colored token visualizes the token-level importance ratio $\rho_t=\pi_\theta/\pi_{\mathrm{rollout}}$. Red indicates $\rho_t>1$, meaning that the current training policy assigns a higher probability to the sampled token than the rollout policy; green indicates $\rho_t<1$, meaning that the current training policy assigns a lower probability. The color intensity encodes the magnitude of the deviation: darker red means a larger ratio above 1, darker green means a smaller ratio below 1, and in both cases a darker color indicates a stronger deviation from the rollout policy. The alternating red and green regions therefore directly reveal why cancellation occurs: positive and negative signed log-ratio deviations coexist in the same response and can offset each other when vanilla sequence-level masking averages signed log-ratios.

\end{document}